\documentclass[10pt]{article} % For LaTeX2e
\usepackage[preprint]{tmlr}

\usepackage{amsmath,amsfonts,bm}

\def\eqref#1{equation~\ref{#1}}
\def\1{\bm{1}}

\DeclareMathAlphabet{\mathsfit}{\encodingdefault}{\sfdefault}{m}{sl}
\SetMathAlphabet{\mathsfit}{bold}{\encodingdefault}{\sfdefault}{bx}{n}

\DeclareMathOperator*{\argmax}{arg\,max}
\DeclareMathOperator*{\argmin}{arg\,min}

\usepackage{hyperref}
\usepackage{url}

\usepackage[utf8]{inputenc} % allow utf-8 input
\usepackage[T1]{fontenc}    % use 8-bit T1 fonts
\usepackage{hyperref}       % hyperlinks
\usepackage{url}            % simple URL typesetting
\usepackage{booktabs}       % professional-quality tables
\usepackage{amsfonts}       % blackboard math symbols
\usepackage{nicefrac}       % compact symbols for 1/2, etc.
\usepackage{microtype}      % microtypography
\usepackage{xcolor}         % colors

\usepackage{amsmath} % declare math operator
\usepackage{amsthm} % proof environment
\usepackage{graphicx} % figure
\usepackage{multirow} % multi-row command
\usepackage[ruled,vlined]{algorithm2e} % algorithm environment
\usepackage{algorithmic} % algorithmic environment
\usepackage{enumitem} % remove left margin of itemize
\usepackage{caption}
\usepackage{subcaption} % subfigure

\colorlet{color1}{blue}
\colorlet{color2}{red!50!black}

\definecolor{ivory}{RGB}{218,215,203}

\definecolor{cuhkp}{RGB}{98,56,105} 	% purple dark
\definecolor{cuhkpl}{RGB}{152,24,147} 	% purple light
\definecolor{cuhkb}{RGB}{219,160,1} 	% ocher
\definecolor{cuhkbd}{RGB}{178,129,0} 	% ocher dark
\definecolor{cuhkr}{RGB}{88,35,155}  	% magenta-red

\usepackage{hyperref}[6.83]
\hypersetup{
	colorlinks=true,
	frenchlinks=false,
	pdfborder={0 0 0},
	naturalnames=false,
	hypertexnames=false,
	breaklinks,
	allcolors = cuhkbd,
	urlcolor = color2,
}

\title{Finite-Time Analysis of Decentralized Single-Timescale \newline Actor-Critic}

\author{\name Qijun Luo \email qijunluo@link.cuhk.edu.cn \\
      \addr School of Science and Engineering\\
      Shenzhen Research Institute of Big Data (SRIBD) \\
      The Chinese University of Hong Kong, Shenzhen \\
      Shenzhen, China \\[10pt]
      \name Xiao Li \email lixiao@cuhk.edu.cn \\
      \addr School of Data Science\\
      Shenzhen Institute of Artificial Intelligence and Robotics for Society (AIRS) \\
      The Chinese University of Hong Kong, Shenzhen \\
      Shenzhen, China
  }

\def\month{01}  % Insert correct month for camera-ready version
\def\year{2023} % Insert correct year for camera-ready version
\def\openreview{\url{https://openreview.net/forum?id=KQRv0O8iW4}} % Insert correct link to OpenReview for camera-ready version

\newcommand{\cA}{\mathcal{A}}
\newcommand{\cP}{\mathcal{P}}
\newcommand{\cS}{\mathcal{S}}
\newcommand{\cN}{\mathcal{N}}

\newcommand{\cO}{\mathcal{O}}
\newcommand{\cE}{\mathcal{E}}

\newcommand{\cG}{\mathcal{G}}

\newcommand{\RR}{\mathbb{R}}
\newcommand{\EE}{\mathbb{E}}
\newcommand{\PP}{\mathbb{P}}
\newcommand{\VV}{\mathbb{V}}
\newcommand{\NN}{\mathbb{N}}
\newcommand{\pit}{\pi_{\theta}}
\newcommand{\bone}{\textbf{1}}
\newcommand{\bomega}{\boldsymbol{\omega}}
\newcommand{\blambda}{\boldsymbol{\lambda}}
\newcommand\numberthis{\addtocounter{equation}{1}\tag{\theequation}}

\newtheorem{theorem}{Theorem}

\newtheorem{lemma}{Lemma}
\newtheorem{proposition}{Proposition}
\newtheorem{assumption}{Assumption}

\begin{document}

\maketitle

\begin{abstract}
	Decentralized Actor-Critic (AC) algorithms have been widely utilized for multi-agent reinforcement learning (MARL) and have achieved remarkable success. Apart from its empirical success, the theoretical convergence property of decentralized AC algorithms is largely unexplored. Most of the existing finite-time convergence results are derived based on either double-loop update or two-timescale step sizes rule, {and this is the case even for centralized AC algorithm under a single-agent setting}. In practice, the \emph{single-timescale} update is widely utilized, where actor and critic are updated in an alternating manner with step sizes being of the same order. In this work, we study a decentralized \emph{single-timescale} AC algorithm.
	Theoretically, using linear approximation for value and reward estimation, we show that the algorithm has sample complexity of $\tilde{\mathcal{O}}(\varepsilon^{-2})$ under Markovian sampling, which matches the optimal complexity with {a} double-loop implementation (here, $\tilde{\mathcal{O}}$ hides a logarithmic term).  {When we reduce to the single-agent setting, our result yields new sample complexity for centralized AC using a single-timescale update scheme.}
	The central to establishing our complexity results is \emph{the hidden smoothness of the optimal critic variable} we revealed. We also provide a local action privacy-preserving version of our algorithm and its analysis. Finally, we conduct experiments to show the superiority of our algorithm over the existing decentralized AC algorithms.
\end{abstract}

\section{Introduction}\label{section:introduction}

% RL -> MARL -> MAAC -> theory of decentralized MAAC, short review
Multi-agent reinforcement learning (MARL)~\citep{littman1994markov, vinyals2019grandmaster} has been successful in various models of multi-agent systems, such as robotics~\citep{lillicrap2015continuous}, autonomous driving~\citep{yu2019distributed}, Go~\citep{silver2017mastering}, etc.  MARL has been extensively explored in the past decades; see, e.g., \citep{lowe2017multi, omidshafiei2017deep, zhang2021multi, son2019qtran, espeholt2018impala, rashid2018qmix}.
These works either focus on the setting where a central controller is available, or assuming a common reward function for all agents. Among the many cooperative MARL settings, the work \citep{zhang2018fully} proposed the fully decentralized MARL with networked agents. In this setting, each agent maintains a private heterogeneous reward function, and agents can only access local/neighboring information through communicating with its neighboring agents on the network. Then, the objective of all agents is to jointly maximize the average long-term reward through interacting with environment modeled by multi-agent Markov decision process (MDP).
They proposed the decentralized Actor-Critic (AC) algorithm to solve this MARL problem, and showed its impressive performance.  
However, the theoretical convergence properties of such class of decentralized AC algorithms are largely unexplored; see \citep{zhang2021multi} for a comprehensive survey. In this work, our goal is to establish the finite-time convergence results under this fully decentralized MARL setting. 
% Compared with single-agent AC, the analysis on the decentralized setting is naturally more challenging due to incomplete information and an additional consensus error term. 
We first review some recent progresses on this line of research below.

\textbf{Related works and motivations.} The first fully decentralized AC algorithm with provable convergence guarantee was proposed by~\citep{zhang2018fully}, and they achieved  asymptotic convergence results under two-timescale step sizes, which requires actor's step sizes to diminish in a faster scale than the critic's step sizes.
The sample complexities of  decentralized AC were established recently. In particular,~\citep{chen2021sample} and~\citep{hairi2021finite} independently proposed two communication efficient decentralized AC algorithms with optimal sample complexity of $\cO(\varepsilon^{-2}\log(\varepsilon^{-1}))$ under Markovian sampling scheme. Nevertheless, their analysis are based on \emph{double-loop} implementation, where each policy optimization step follows a nearly accurate critic optimization step (a.k.a. policy evaluation), i.e., solving the critic optimization subproblem to $\varepsilon$-accuracy. 
Such a double-loop scheme requires careful tuning of two additional hyper-parameters, which are the batch size and inner loop size. In particular, the batch size and inner loop size need to be of order $\mathcal{O}(\varepsilon^{-1})$ and  $\mathcal{O}(\log(\varepsilon^{-1}))$ in order to achieve their sample complexity results, respectively.
% Such a double-loop scheme may have some practical issues as performing an accurate critic optimization is often time-consuming. In addition,  it is not clear how many benefits one can have through solving the critic optimization accurately since it is just one iteration of the overall alternating-type algorithm.
 In practice, single-loop algorithmic framework is often utilized, where one updates the actor and critic in an alternating manner by performing {any constant} algorithmic iterations for {both subproblems}; see, e.g., ~\citep{schulman2017proximal, lowe2017multi, lin2019communication, zhang2020bi}. The work ~\citep{zeng2022learning} proposed a new decentralized AC algorithm based on such a single-loop alternative update. However, they have to adopt \emph{two-timescale} step sizes rule to ensure convergence, which requires actor's step sizes to diminish in a faster scale than the critic's step sizes. Due to the separation of the step sizes, the critic optimization subproblem  is solved exactly  when the number of iterations tends to $\infty$. Such a restriction on the step size will slow down the convergence speed of the algorithm. As a consequence, they only obtain sub-optimal sample complexity of $\cO(\varepsilon^{-\frac{5}{2}})$. In practice, most algorithms are 
implemented with \emph{single-timescale} step sizes rule, where  the step sizes for the actor's and critic's updates are  of the same order. 
Though there are some theoretical achievements for single-timescale update in other areas such as TDC~\citep{wang2021tdc} and bi-level optimization~\citep{chen2021closing}, similar theoretical understanding under AC setting is largely unexplored.

Indeed, even when reducing to single-agent setting, the convergence property of single-timescale AC algorithm is not well established. The works \citep{fu2021singletimescale, guo2021decentralized} established the finite-time convergence result under a special single-timescale implementation, where they attained the sample complexity of $\cO(\varepsilon^{-2})$. Their analysis is based on an algorithm where the critic optimization step is formulated as a least-square temporal difference (LSTD) at each iteration, {which requires} to sample the transition tuples for $\tilde\cO(\varepsilon^{-1})$ times to form the data matrix in the LSTD {subproblem}. Then, they solve the LSTD {subproblem} in a closed-form fashion {by inverting} a matrix of large size. 
Later,~\citep{chen2021closing} obtained the same sample complexity using TD(0) update for critic variables
under i.i.d. sampling. Their analysis highly relies on the assumption that the Jacobian of the stationary distribution is Lipschitz continuous, which is not justified in their work.

% Thus, by summarizing the above recent advances, we conclude that the existing  finite-time convergence results for decentralized AC algorithms (even for single-agent AC algorithm) either use double-loop algorithmic framework or employ a two-timescale step sizes rule. This observation motivates us to ask the following question:

The above observations motivate us to ask the following question:

\emph{Can we establish finite-time convergence result for decentralized AC algorithm with single-timescale step sizes rule?\footnote{As convention in  \citep{fu2021singletimescale}, when we use "single-timescale", it means we utilize a single-loop algorithmic framework with single-timescale step sizes rule.}}

%  In this work, we will answer this question positively. We now list our main contributions below. 

\textbf{Main contributions.} By answering this question positively, we have the following contributions:

% \qijun{discuss the main challenge of single-timescale MAAC.}

% \textbf{Main contributions.} 
% In this work, we analyze a fully decentralized AC algorithm, which utilizes a single-loop algorithmic framework, employs a single timescale step sizes rule, and adopts Markovian sampling scheme. Our key results are summarized below:

\begin{enumerate}[label=$\bullet$,topsep=0pt,itemsep=0ex,partopsep=0ex]
	\item We design a decentralized AC algorithm, which employs a \emph{single-timescale} step sizes rule and adopts Markovian sampling scheme. The proposed algorithm allows communication between agents for every $K_c$ iterations with $K_c$ being any integer lies in $[1, \cO(\varepsilon^{-\frac{1}{2}})]$, rather than communicating at each iteration as adopted by previous single-loop decentralized AC algorithms ~\citep{zeng2022learning, zhang2018fully}.

	\item {Using linear approximation for value and reward estimation,} we establish the \emph{finite-time} convergence result for {the proposed} algorithm under standard assumptions. In particular, we show that the algorithm has a sample complexity of $\tilde{\cO}(\varepsilon^{-2})$, which matches the optimal complexity up to a logarithmic term. In addition, we show that the logarithmic term {hidden in the ``$\tilde{\cO}$''} can be removed under the i.i.d. sampling scheme. These convergence results are valid for all the above mentioned choices for $K_c$. 
	
	\item To preserve {privacy of local actions}, we propose a variant of our algorithm which utilizes noisy local rewards for estimating global rewards. We show that such an algorithm will maintain the optimal sample complexity at the expense of communicating at each iteration. 
\end{enumerate}

Our key technical result is to reveal \emph{the hidden smoothness of the optimal critic variable}, so that we can derive a sufficient descent on the averaged critic's optimal gap under the single-timescale update. Consequently, we can resort to the classic convergence analysis for alternating optimization algorithms to establish the approximate ascent property of the overall optimization process, which leads to the final sample complexity results. We also designed a Lyapunov function to analyze the descent of the objective function with a single-timescale update under the decentralized setting.

{When we reduce to the non-decentralized case, i.e., the single-agent setting, our results yield  \emph{new} sample complexity guarantees for the classic centralized AC algorithm using a \emph{single-timescale} update scheme.}

% We remark that our convergence results are even new for single-agent AC algorithms under the setting of single-timescale step sizes rule.

\textbf{Discussion on a concurrent work}. We note that there is a concurrent work~\citep{olshevsky2023small} which also analyzes the single-timescale AC algorithm and achieves similar complexity results. Their analysis is based on the small gain theorem, which is different from ours. These two analysis frameworks provide useful insights for the AC algorithm from different perspectives. \citep{olshevsky2023small} shows that the coupled expression on the errors of actor and critic can be fit into a non-linear small gain theorem framework, which bounds the actor's error by desired order. Our analysis reveals the hidden smoothness of the optimal critic variable so that approximate descent on the critic's objective can be achieved. 
In addition, \citep{olshevsky2023small} considers the single-agent setting while our analysis deals with {the more general decentralized setting}. 
Moreover, \citep{olshevsky2023small} analyzes the i.i.d. sampling scheme where the single agent is assumed to have access to the transition tuples from the stationary distribution and the discounted state-visitation distribution. By contrast, our setting considers the practical Markovian sampling scheme, where the transition tuples are from the trajectory generated during the update of the agents.

\section{Preliminary}
In this section, we introduce the problem formulation and the policy gradient theorem, which serves as the preliminary for the analyzed decentralzed AC algorithm.

Suppose there are multiple agents aiming to independently optimize a common global objective, and each agent can communicate with its neighbors through a network. To model the topology, we define the graph as $\cG=(\cN, \cE)$, where $\cN$ is the set of nodes with $|\cN|=N$ and $\cE$ is the set of edges with $|\cE|=E$. In the graph, each node represents an agent, and each edge represents a communication link. The interaction between agents follows the networked multi-agent MDP.

\subsection{Markov decision process}

A networked multi-agent MDP is defined by a tuple $(\cG,\cS,\{\cA^i\}_{i\in[N]},\cP,\{r^i\}_{i\in[N]},\gamma)$. $\cG$ denotes the communication topology (the graph),
$\cS$ is the finite state space observed by all agents, $\cA^i$ represents the finite action space of agent $i$. Let $\cA:=\cA^1\times\cdots\times\cA^N$ denote the joint action space and
$\cP(s'|s,a):\cS\times\cA\times\cS\to[0,1]$ denote the transition probability from any state $s\in\cS$ to any state $s'\in\cS$ for any joint action $a\in\cA$. $r^i:\cS\times\cA\to\mathbb{R}$ is the local reward function that determines the reward received by agent $i$ given transition $(s,a)$; $\gamma\in[0,1]$ is the discount factor.

% We will use $\theta^i \in \RR^{d_\theta}$ to denote the parameter of actor $i$, and $\theta:=[\theta^1, \cdots, \theta^N] \in \RR^{Nd_\theta}$ to denote the joint parameter of actor. We also denote the joint action as $a:=[a^1, \cdots, a^N]$.
For simplicity, we will use $a:=[a^1, \cdots, a^N]$ to denote the joint action, and $\theta \in \RR^{N d_\theta}$ to denote concatenation of all actor's joint parameters of all actors, with $\theta^i \in \RR^{d_\theta}$. Here, without loss of generality, we assume that every agent has the same number of parameters for notation brevity. The MDP goes as follows: For a given state $s$, each agent make its decision $a^i$ based on its policy $a^i \sim \pi_{\theta^i}(\cdot|s)$. The state transits to the next state $s'$ based on the joint action of all the agents: $s' \sim \cP(\cdot|s,a)$. Then, each agent will receive its own reward $r^i(s, a)$. For the notation brevity, we assume that the reward function mapping is deterministic and does not depend on the next state without loss of generality. The stationary distribution induced by the policy $\pi_{\theta}$ and the transition kernel is denoted by $\mu_{\pit}(s)$.

Our objective is to find a set of policies that maximize the accumulated discounted mean reward received by agents
\begin{align}
\theta^* = \argmax_{\theta} J(\theta):= \EE\left[\sum_{k=0}^{\infty}\gamma^k \bar{r}(s_k,a_k)\right].
\label{eq:objective}
\end{align}
Here, $k$ represents the time step. $\bar{r}(s_k, a_k):=\frac{1}{N}\sum_{i=1}^{N} r^i(s_k, a_k)$ is the mean reward among agents at time step $k$. The randomness of the expectation comes from the initial state distribution $\mu_{0}(s)$, the transition kernel $\cP$, and the stochastic policy $\pi_{\theta^i}(\cdot|s)$.

\subsection{Policy gradient Theorem}
Under the discounted reward setting, the global state-value function, action-value function, and advantage function for policy set $\theta$, state $s$, and action $a$, are defined as 
\begin{align*}
V_{\pi_\theta}(s)&:= \EE\left[\sum_{k=0}^{\infty} \gamma^k \bar{r}(s_k,a_k) | s_0=s\right]\numberthis
\label{eq:state-value} \\
Q_{\pi_\theta}(s,a) &:= \EE\left[\sum_{k=0}^{\infty} \gamma^k \bar{r}(s_k,a_k) | s_0=s, a_0=a\right] \\
A_{\pi_\theta}(s,a) &:= Q_{\pi_\theta}(s,a) - V_{\pi_\theta}(s).
\end{align*}
%The state-action value function for a given set of policies $\theta$, state $s$, and action $a$, is defined as 
%\begin{align*}
%Q_{\pi_\theta}(s,a) := \EE[\sum_{k=0}^{\infty} \gamma^k \bar{r}(s_k,a_k) | s_0=s, a_0=a].\numberthis
%\label{eq:state-action-value}
%\end{align*}
%Based on the above definitions, the advantage function is defined as 
%\begin{align*}
%A_{\pi_\theta}(s,a) := Q_{\pi_\theta}(s,a) - V_{\pi_\theta}(s). \numberthis
%\label{eq:advantage}
%\end{align*}
To maximize the objective function defined in (\ref{eq:objective}), the policy gradient~\citep{sutton2000policy} can be computed as follow 
% {\color{blue}[references?]}
\begin{align*}
\nabla_{\theta} J(\theta) = \EE_{s \sim d_{\pi_\theta}, a \sim \pi_\theta} \left[\frac{1}{1-\gamma}A_{\pi_\theta}(s,a)\psi_{\pi_\theta}(s,a)\right], 
\end{align*}
where $d_{\pi_\theta}(s):=(1-\gamma)\sum_{k=0}^{\infty}\gamma^{k}\PP(s_k=s)$ is the discounted state visitation distribution under policy $\pit$, and  $\psi_{\pi_\theta}(s,a):=\nabla \log\pi_{\theta}(s,a)$ is the score function. 

Following the derivation of~\citep{zhang2018fully}, the policy gradient for each agent under discounted reward setting can be expressed as
\begin{align*}
\nabla_{\theta^i}J(\theta) = \EE_{s \sim d_{\pit}, a\sim \pit} \left[\frac{1}{1-\gamma} A_{\pit}(s,a) \psi_{\pi_{\theta^i}}(s,a^i) \right]. \numberthis \label{eq:policy-gradient-MARL}
\end{align*}

\section{Algorithms}
\label{section:decentralized-ac}

\subsection{Decentralized single-timescale Actor-Critic}
\begin{algorithm}[h]
	\null
	\caption{Decentralized single-timescale AC (reward estimator version)}
	\label{algorithm:dec-ac-re}
	\small
	\begin{algorithmic}[1]
		\STATE {\bfseries Initialize:} Actor parameter $\theta_0$, critic parameter $\omega_0$, reward estimator parameter $\lambda_0$, initial state $s_0$.
		\FOR{$k=0,\cdots,K-1$}
		
		\STATE \textbf{{Option 1: i.i.d. sampling:}}
		\STATE $s_k \sim \mu_{\theta_k}(s), a_k \sim \pi_{\theta_k}(\cdot|s_k), s_{k+1} \sim \cP(\cdot | s_k, a_k)$.
		
		\STATE \textbf{{Option 2: Markovian sampling:}}
		\STATE $a_k \sim \pi_{\theta_k}(\cdot|s_k), s_{k+1} \sim \cP(\cdot | s_k, a_k)$.
		
		\STATE
		
		\STATE \textbf{{Periodical consensus:}} Compute $\tilde{\omega}_k^i$ and $\tilde{\lambda}_k^i$ by (\ref{eq:critic-consensus}) and (\ref{eq:reward-estimator-consensus}).
		
		\STATE
		
		\FOR{$i=0, \cdots, N$ \textbf{in parallel}}
		
		\STATE {\textbf{Reward estimator update:}} update $\lambda_{k+1}^i$ by (\ref{eq:reward-estimator-update}).
		
		\STATE {\textbf{Critic update:}} Update $\omega_{k+1}^{i}$ by (\ref{eq:critic-update}).

		\STATE {\textbf{Actor update:}} Update $\theta_{k+1}^{i}$ by (\ref{eq:actor-update}).

		\ENDFOR
		\ENDFOR
	\end{algorithmic}
\end{algorithm}

We introduce the decentralized single-timescale AC algorithm; see Algorithm \ref{algorithm:dec-ac-re}. In the remaining parts of this section, we will explain the updates in the algorithm in details.

In fully-decentralized MARL, each agent can only observe its local reward and action, while trying to maximize the global reward (mean reward) defined in (\ref{eq:objective}). 
%We consider the  Actor-Critic algorithm to solve the problem. For the iteration, each local critic tries to estimate the global value function, and the actor updates its parameter according to the policy gradient theorem in~\ref{eq:policy-gradient}, based on the estimated value function. The detailed update is as follow:
The decentralized AC algorithm solves the problem by updating actor and critic variables alternatively on an online trajectory. Specifically, we have $N$ pairs of actor and critic. In order to maximize $J(\theta)$, each critic tries to estimate the \emph{global} state-value function $V_{\pi_\theta}(s)$ defined in (\ref{eq:state-value}). Then, each actor updates its policy parameter based on approximated policy gradient. We now provide more details about the algorithm.

{\bf Critics' update.} We will use $\omega^i \in \RR^{d_\omega}$ to denote the $i_{th}$ critic's parameter and $\bar{\omega}:=\frac{1}{N}\sum_{i=1}^{N}\omega^i$ to represent the averaged parameter of critic. Each critic approximates the {global} value function as  $V_{\pit}(s) \approx \hat{V}_{\omega^i}(s)$.

The critic's approximation error can be categorized into two parts, namely, the consensus error 
$\frac{1}{N}\sum_{i=1}^{N}\|\omega^i - \bar{\omega}\|,$
which measures how close the critics' parameters are; and the approximation error
$\|\bar{\omega} - \omega^*(\theta)\|,$
which measures the approximation quality of averaged critic.

In order for critics to reach consensus, { each critic exchanges its parameters with neighbors and perform the following update
\begin{align*}
\tilde{\omega}_{k}^{i} = 
\begin{cases}
\sum_{j=1}^{N} W^{ij}\omega_k^j \quad &\text{if}~ k\hspace{-0.2cm}\mod K_c=0 \\
\omega_k^i \quad &\text{otherwise}.
\end{cases}
\numberthis \label{eq:critic-consensus}
\end{align*}
Here, $K_c$ denotes the consensus frequency. The communication matrix $W \in \RR^{n\times n}$ is usually determined artificially in practice and can be sparse, which means that the number of neighbors for each agent is much fewer than the total number of agents. Thus, the cost for each consensus step is usually much lower than a full synchronization over the network.} The detailed requirements of matrix $W$ will be discussed in Assumption~\ref{assump:doub_stoch}.

% is a sparse communication matrix which is usually determined artificially in practice.

% for communication among agents, whose property will be specified in Assumption~\ref{assump:doub_stoch}; $K_c$ denotes the consensus frequency.

To reduce the approximation error, we will perform the local TD(0) update~\citep{tsitsiklis1997analysis} as 
\begin{align*}
%\delta^i(\xi_k, \omega_k^i)&:= r^i(s_k, a_k) + \gamma V_{\omega_k^i}(s_k) - V_{\omega_k^i}(s_{k+1}).
% \delta^i(\xi, \omega)&:=r^i(s,a) + \gamma \hat{V}_\omega(s') - \hat{V}_\omega(s) \\
% g_c^i(\xi, \omega)&:=\delta^i(\xi, \omega)\nabla \hat{V}_\omega(s) \\
\omega_{k+1}^i &= \Pi_{R_\omega}(\tilde{\omega}_k^i + \beta_k g_c^i(\xi_k, \omega_k^i)),
\numberthis \label{eq:critic-update}
\end{align*} 
where $\xi := (s, a, s')$ represents a transition tuple, $g_c^i(\xi, \omega):=\delta^i(\xi, \omega)\nabla \hat{V}_\omega(s)$ is the update direction, $\delta^i(\xi, \omega):=r^i(s,a) + \gamma \hat{V}_\omega(s') - \hat{V}_\omega(s)$ is the local temporal difference error (TD-error). $\beta_k$ is the step size for critic at iteration $k$. $\Pi_{R_\omega}$ projects the parameter into a ball of radius of $R_\omega$ containing the optimal solution, which will be explained when discussing Assumptions~\ref{assump:boundedness} and~\ref{assump:negative-definiteness}. 

%It is directly seen that $\omega^*(\theta)$ is a fixed point of~\ref{eq:critic-update} when the expectation for $s_k$ is taken over stationary distribution $\mu_\theta$. However, in practice, states are sampled from a Markov decision process, and thus introduce biases in the update. Such an error needs to be bounded when analyzing the algorithm's convergence.

{\bf Actors' update.} We will use stochastic gradient ascent to update the policy's parameter, which is calculated based on policy gradient theorem in (\ref{eq:policy-gradient-MARL}). The advantage function $A_{\pi_\theta}(s,a)$ can be estimated by
\begin{align*}
\delta(\xi, \theta) := \bar{r}(s,a) + \gamma V_{\pi_\theta}(s') - V_{\pi_\theta}(s),
\end{align*}
with $a$ sampled from $\pit(\cdot|s)$. However, to preserve the privacy of each agents, the local reward cannot be shared to other agents under the fully decentralized setting. Thus, the averaged reward $\bar{r}(s_k, a_k)$ is not directly attainable. To this end, we adopt the strategy proposed in~\citep{zhang2018fully} to approximate the averaged reward. In particular, each agent $i$ will have a local reward estimator with parameter $\lambda^i \in \RR^{d_\lambda}$, which estimates the global averaged reward as $\bar{r}(s_k, a_k) \approx \hat{r}_{\lambda^i}(s_k, a_k)$. 

Thus, the update of the $i_{th}$ actor is given by
\begin{align*}
\theta_{k+1}^i = \theta_{k}^{i} + \alpha_k \hat{\delta}(\xi_k, \omega_{k+1}^{i}, \lambda_{k+1}^{i})\psi_{\pi_{\theta_k^i}}(s_k, a_k^i), \numberthis \label{eq:actor-update}
\end{align*}
where $\hat{\delta}(\xi, \omega, \lambda) := \hat{r}_\lambda(s,a) + \gamma\hat{V}_\omega(s') - \hat{V}_\omega(s)$ is the approximated advantage function. $\alpha_k$ is the step size for actor's update at iteration $k$.

%When using expression in ~\ref{eq:policy-gradient} for policy gradient, the expectation of $s_k$ is taken over the discounted state visitation distribution $d_{\pit}(s)$. Nevertheless, in practical implementation, the states are sampled from an MDP. The bias induced by distribution mismatch needs to be bounded in the analysis of the algorithm.

{\bf Reward estimators' update.} Similar to critic, each reward estimator's approximation error can be decomposed into consensus error and the approximation error. 

For each local reward estimator, we perform the consensus step to minimize the consensus error as
\begin{align*}
\tilde{\lambda}_{k}^{i} = 
\begin{cases}
\sum_{j=1}^{N} W^{ij}\lambda_k^j \quad &\text{if}~ k\hspace{-0.2cm}\mod K_c=0 \\
\lambda_k^i \quad &\text{otherwise}.
\end{cases}
\numberthis \label{eq:reward-estimator-consensus}
\end{align*}
To reduce the approximation error, we perform a local update of stochastic gradient descent.
\begin{align*}
% g_r^i(\xi, \lambda) &= (r^i(s, a) - \hat{r}_\lambda(s,a)) \nabla \hat{r}_\lambda(s,a) \\
\lambda_{k+1}^{i} &= \Pi_{R_\lambda}(\tilde{\lambda}_k^i + \eta_k g_r^i(\xi_k, \lambda_k^i)), \numberthis
\label{eq:reward-estimator-update}
\end{align*}
where $g_r^i(\xi, \lambda) := (r^i(s, a) - \hat{r}_\lambda(s,a)) \nabla \hat{r}_\lambda(s,a)$ is the update direction. 
$\eta_k$ is the step size for reward estimator at iteration $k$. Note the calculation of $g_r^i(\xi, \lambda)$ does not depend on the next state $s'$; we use $\xi$ in (\ref{eq:reward-estimator-update}) just for notation brevity. Similar to critic's update, $\Pi_{R_\lambda}$ projects the parameter into a ball of radius of $R_\lambda$ containing the optimal solution.

In our Algorithm \ref{algorithm:dec-ac-re}, we will use the same order for $\alpha_k$, $\beta_k$, and $\eta_k$ and hence, our algorithm is in \emph{single-timescale}.

\textbf{Linear approximation for analysis.} In our analysis, we will use linear approximation for both critic and reward estimator variables, i.e. $\hat{V}_{\omega}(s):=\phi(s)^T\omega$; $ \hat{r}_{\lambda}(s,a):=\varphi(s,a)^T\lambda$, where $\phi(s):\cS\to \RR^{d_\omega}$ and $\varphi(s,a):\cS \times \cA \to \RR^{d_\lambda}$ are two feature mappings, whose property will be specified in the discussion of Assumption~\ref{assump:boundedness}.

% Such an strategy is inspired by~\citep{chen2021sample}'s noisy reward communication approach under double-loop implementation. Due to space limit, we defer 

% Sampling remark.
% Compare with zkq.
\textbf{Remarks on sampling scheme.} {Acquiring unbiased stochastic gradients} for critic and actor variables requires sampling from $\mu_{\pi_\theta}$ and $d_{\pi_{\theta}}$, respectively. However, in practical implementations, states are usually collected from an online trajectory (Markovian sampling), whose distribution is generally different from $\mu_{\pit}$ and $d_{\pit}$. Such a distribution mismatch will inevitably cause biases during the update of critic and actor variables. One has to bound the corresponding error terms when analyzing the algorithm.

\begin{algorithm}[t]
	\null
	\caption{Decentralized single-timescale AC (noisy reward version)}
	\label{algorithm:dec-ac-noisy}
	\small
	\begin{algorithmic}[1]
		\STATE {\bfseries Initialize:} Actor parameter $\theta_0$, critic parameter $\omega_0$, initial state $s_0$.
		\FOR{$k=0,\cdots,K-1$}
		
		\STATE \textbf{{Option 1: i.i.d. sampling:}}
		\STATE $s_k \sim \mu_{\theta_k}(s), a_k \sim \pi_{\theta_k}(\cdot|s_k), s_{k+1} \sim \cP(\cdot | s_k, a_k)$.
		
		\STATE \textbf{{Option 2: Markovian sampling:}}
		\STATE $a_k \sim \pi_{\theta_k}(\cdot|s_k), s_{k+1} \sim \cP(\cdot | s_k, a_k)$.
		
		\STATE
		
		\STATE \textbf{{Periodical consensus:}} Compute $\tilde{\omega}_k^i$ by (\ref{eq:critic-consensus}).
		
		\STATE
		
		\FOR{$i=0, \cdots, N$ \textbf{in parallel}}
		
		\STATE {\textbf{Global reward estimation:}} estimate $\bar{r}_k(s_k, a_k)$ by (\ref{eq:re-noise-consensus}).
		
		\STATE {\textbf{Critic update:}} Update $\omega_{k+1}^{i}$ by (\ref{eq:critic-update}).
		
		\STATE {\textbf{Actor update:}} Update $\theta_{k+1}^{i}$ by (\ref{eq:actor-update-re}).
		\ENDFOR
		\ENDFOR
	\end{algorithmic}
\end{algorithm}

\subsection{Variant for preserving local action}
Note that in Algorithm~\ref{algorithm:dec-ac-re}, the reward estimators need the knowledge of joint actions in order to estimate the global rewards. Inspired by~\citep{chen2021sample}, we further propose a variant of Algorithm~\ref{algorithm:dec-ac-re} to preserve the privacy of local actions. It estimates the global rewards by communicating noisy local rewards. As a trade-off, the approach requires $\cO(\log(\varepsilon^{-1}))$ communication rounds for each iteration; see Algorithm~\ref{algorithm:dec-ac-noisy}.

Let $r_k^i$ represents $r_k^i(s_k,a_k)$ for brevity. The reward estimation process goes as follow:
for each agent $i$, we first produce a noisy local reward $\tilde{r}_k^i = r_k^i(1+z)$, with $z \sim \cN(0, \sigma^2)$. Thus, the noise level is controlled by the variance $\sigma^2$, which is chosen artificially. When the noise level $\sigma^2$ increases, the local reward's privacy will be strengthen. In the meantime, the variance of the estimated global reward will increase. To estimate the global reward, each agent $i$ first initialize the estimation as $\tilde{r}_{k, 0}^i = \tilde{r}_{k}^{i}$. Then, each agent $i$ perform the following consensus step for $K_r$ times, i.e.
\begin{align*}
\tilde{r}_{k, l+1}^i = \sum_{j=1}^{N}W^{ij}\tilde{r}_{k, l}^i, \quad l=0,1,\cdots,K_r-1.
\numberthis \label{eq:re-noise-consensus}
\end{align*}
The reward $\tilde{r}_{k, K_r}^i$ will be used for estimating the global reward for agent $i$ at $k_{th}$ iteration. As we will see, the error $|\tilde{r}_{t, l+1}^{i} - \frac{1}{N}\sum_{i=1}^{N}\tilde{r}_{k}^{i}|$ will converge to 0 linearly. Hence, to reduce the error to $\varepsilon$, we need $K_r=\cO(\log(\varepsilon^{-1}))$ rounds of communications for each iteration. Based on the estimated global reward, the $i_{th}$ actor's update is given by
\begin{align*}
    \theta_{k+1}^{i} = \theta_k^i + \alpha_k (\tilde{r}_{k,K_r}^{i} + \gamma \hat{V}_{\omega^i}(s') - \hat{V}_{\omega^i}(s))\psi_{\pi_{\theta_k^i}}(s_k, a_k^i). \numberthis \label{eq:actor-update-re}
\end{align*}

% \vspace{-0.5cm}
\section{Main results}\label{section:main-result}
In this section, we first introduce the technical assumptions used for our analysis, which are standard in the literature. Then, we present the convergence results for both actor and critic variables.

% \vspace{-0.2cm}
\subsection{Assumptions}\label{subsection:assumptions}
\begin{assumption}[boundedness of rewards and feature vectors]\label{assump:boundedness}
The local rewards are uniformly bounded, i.e., there exists a positive constant $r_{\max}$ such that for all feasible $(s, a)$ and $i \in [N]$, we have $|\,r^{i}(s,a)\,|\leq r_{\max}$. The norm of feature vectors are bounded such that for all $s \in \cS, \ a \in \cA$, $\|\phi(s)\|\leq 1, \| \varphi(s,a) \| \leq 1$.\footnote{Through out the paper, we will use $\|\cdot\|$ to represent the Euclidean norm for vectors and Frobenius norm for matrices.}
\end{assumption}
Assumption~\ref{assump:boundedness} is standard and commonly adopted; see, e.g., ~\citep{bhandari2018finite, xu2020improving, zeng2022learning, shen2020asynchronous, qiu2019finite}. This assumption can be achieved via normalizing the feature vectors.

\begin{assumption}[sufficient exploration]\label{assump:negative-definiteness}
There exists two positive constants $\lambda_{\phi}, \lambda_{\varphi}$ such that for all policy $\pi_\theta$, the following two matrices are negative definite
	\begin{align*}
	A_{\theta, \phi} &:= \EE_{s \sim \mu_{\theta}(s)}[\phi(s)(\gamma\phi(s')^T-\phi(s)^T)] \\
	A_{\theta, \varphi} &:= \EE_{s\sim \mu_{\theta}(s), a \sim \pi_{\theta}(\cdot|s)}[-\varphi(s,a)\varphi(s,a)^T],
	\end{align*}
	with $\lambda_{\max}(A_{\theta, \phi}) \leq \lambda_{\phi}, \lambda_{\max}(A_{\theta, \varphi}) \leq \lambda_{\varphi}$,  where $\lambda_{\max}(\cdot)$ represents the largest eigenvalue.
	% with expectation is taken over $s \sim \mu_\theta(s), a \sim \pi_{\theta}(\cdot|s)$.
\end{assumption}
The Assumption~\ref{assump:negative-definiteness} characterizes a strong convexity-like property of critic and reward estimator's objective function, and thereby
ensures sufficient decrease of the estimation error for each update. It will be satisfied when $\inf_{\theta, s, a} \pi_\theta(a|s) \geq c$ for all policy $\pi_\theta, s \in \cS, a \in \cA$ with $c$ being positive. Thus, it can be understood as an exploration assumption on policy $\pi_\theta$. (see Proposition 3.1 of \citep{olshevsky2023small} for more detail). This assumption is widely seen in analysis of AC algorithms; see, e.g. \citep{shen2020asynchronous, xu2021sample, zeng2022learning}. 
%Assumption~\ref{assump:negative-definiteness}
%can be achieved when the matrices $\Phi_{\phi}:=[\phi(s_1), \cdots, \phi(s_{|\cS|})]$ and $\Phi_{\varphi}:=[\varphi(s_1,a_1), \cdots, \varphi(s_{|\cS|}, a_{|\cA|})]$ have full row rank, which ensures that the optimal critic and reward estimator are unique; see also~\citep{shen2020asynchronous, xu2021sample}.
Together with Assumption~\ref{assump:boundedness}, we can show that $\|\omega^*(\theta)\| \leq R_{\omega}:=\frac{r_{\max}}{\lambda_\phi}$, $\|\lambda^*(\theta)\| \leq R_{\lambda}:=\frac{r_{\max}}{\lambda_{\varphi}}$, which justifies the projection step. 
In practice, one can estimate $R_\omega$ and $R_\lambda$ online; see Section 8.2 of \citep{bhandari2018finite} for one approach. 
We provide more details for the projection in Appendix~\ref{appendix:supporting-lemma}.

% the norm of $\omega^*(\theta)$ and $\lambda^*(\theta)$ are bounded by some positive constant, which justifies the projection steps.

% {see Appendix~\ref{appendix:supporting-lemma} for the choice of the projection radius.}

\begin{assumption}[Lipschitz properties of policy]\label{assump:lipschitz-policy}
There exists constants $C_{\psi}, L_{\psi}, L_{\pi}$ such that for all policy parameter $\theta, \theta'$, $s \in \cS$ and $a \in \cA$, we have $(1).~ |\pi_{\theta}(a|s) - \pi_{\theta'}(a|s) | \leq L_{\pi} \| \theta - \theta' \|; (2).~ \| \psi_{\theta}(s,a) - \psi_{\theta'}(s,a) \| \leq L_{\psi} \| \theta - \theta' \|; (3).~ \| \psi_{\theta}(s,a) \| \leq C_{\psi}$.
	
	% For all $\theta, \theta', s \in \cS$ and $a \in \cA$, there exist constants $C_{\psi}, L_{\psi}, L_{\pi}$ such that $i) \| \psi_{\theta}(s,a) \| \leq C_{\psi}; \| \psi_{\theta}(s,a) - \psi_{\theta'}(s,a) \| \leq L_{\psi} \| \theta - \theta' \|; iii) |\pi_{\theta}(a|s) - \pi_{\theta'} | \leq L_{\pi} \| \theta - \theta' \|$.
\end{assumption}
Assumption~\ref{assump:lipschitz-policy} is common for analyzing policy-based algorithms; see, e.g., \citep{xu2019improved, wu2020finite, hairi2021finite}. The assumption implies the smoothness of objective function $J(\theta)$. It holds for policy classes such as tabular softmax policy~\citep{agarwal2020optimality}, Gaussian policy~\citep{doya2000reinforcement}, and Boltzmann policy~\citep{konda1999actor}. 

%\begin{assumption}[Lipschitz properties of stationary distribution]\label{assump:lipschitz-stationary}
%For any $\theta, \theta' \in \RR^d$, there exists a constant $L_{\mu,1}$  such that $ \| \nabla \mu_{\theta}(s) - \nabla \mu_{\theta'}(s)\|\leq L_{\mu, 1}\| \theta - \theta' \|$
%\end{assumption}
%The existence of $\nabla \mu_\theta(s)$ has been shown in~\citep{baxter2001infinite}. We note that this assumption is necessary for ensuring the smoothness of $\omega^*(\theta)$ and $\lambda^*(\theta)$.  Assumption~\ref{assump:lipschitz-stationary} is also adopted from~\citep{chen2021closing}. 
% which is used to ensure the descent of critic's error in the proof of Lemma~\ref{lemma:}. 

\begin{assumption}[mixing of Markov chain]\label{assump:markov-chain}
There exists constants $\kappa > 0$ and $\rho \in (0,1)$ such that 
\begin{align*}
\sup_{s\in \cS} d_{{TV}} \ (\PP(s_k \in \cdot | s_0=s, \pit), \mu_\theta) \leq \kappa \rho^{k}, \ \forall k.
\end{align*}
\end{assumption}
Assumption~\ref{assump:markov-chain} is a standard assumption; see, e.g. \citep{bhandari2018finite, wu2020finite, xu2019improved}. The assumption always holds for irreducible and aperiodic Markov chain. It ensures the geometric convergence of state to the stationary distribution.

\begin{assumption}[doubly stochastic weight matrix]\label{assump:doub_stoch} The communication matrix $W$ is doubly stochastic, i.e. each column/row sum up to 1. Moreover, the second largest singular value $\nu$ is smaller than 1.
	% there exists a positive constant $\cW$ such that $W^{ii} \geq \cW, \ \forall i \in \cN$, and $W^{ij} \in [\cW, 1]$ if $i$ and $j$ are connected, otherwise $W^{ij}=0$.
\end{assumption}
Assumption~\ref{assump:doub_stoch} is a common assumption in decentralized optimization and multi-agent reinforcement learning; see, e.g., ~\citep{sun2020finite, chen2021multi, chen2021sample}. It ensures the convergence of consensus error for critic and reward estimator variables.

\subsection{Sample complexity for Algorithm~\ref{algorithm:dec-ac-re}}

% The state distribution in a Markov chain is generally different from the stationary distribution $\mu_{\theta}$ for unbiased update of critic and reward estimator. It also differs from the discounted state visitation distribution $d_{\pi_\theta}$, which is used for the estimation of policy gradient. To tackle this issue, we must bound the biases caused by the mismatch between distributions. 

\begin{theorem}
	\label{thm:main-theorem-markovian}
	Suppose Assumptions~\ref{assump:boundedness}-\ref{assump:doub_stoch} hold. Consider the update of  Algorithm~\ref{algorithm:dec-ac-re} under Markovian sampling. Let $\alpha_k= \frac{\bar{\alpha}}{\sqrt{K}}$ for some positive constant $\bar{\alpha}$, $\beta_k=\frac{C_{9}}{2\lambda_{\phi}}\alpha_k$, and $\eta_k=\frac{C_{10}}{2\lambda_{\varphi}}\alpha_k$ and $K_c \leq \cO({K^{1/4}})$, where $K$ is the total number of iterations. Then, we have
	%	Let $\alpha_k = \alpha = \frac{\bar{\alpha}}{\sqrt{K}}$, $\beta_k = \beta =\frac{L_{\omega}^2 + C_1}{4\lambda} \alpha$, under the assumptions specified in (\ref{subsection:assumptions}), the algorithm~\ref{algorithm:dec-ac-re} will yield the following sample complexity
	\begin{align*}
	&\frac{1}{K} \sum_{k=1}^{K} \sum_{i=1}^{N}\EE\left[\left\|\omega_{k}^{i}-\omega^*(\theta_k)\right\|^2\right] \leq \cO\left(\frac{\log^2K}{\sqrt{K}}\right) \\
	&\frac{1}{K} \sum_{k=1}^{K} \sum_{i=1}^{N} \EE \left[\left\| \nabla_{\theta^i} {J}(\theta_k) \right\|^2\right] \leq \mathcal{O}\left(\frac{\log^2 K}{\sqrt{K}}\right)+ \cO\left(\varepsilon_{app}+\varepsilon_{sp}\right), \numberthis \label{eq:main-convergence-markovian}
	\end{align*}
	where $C_9, C_{10}$ are positive constants defined in proof.
\end{theorem}

The proof of Theorem~\ref{thm:main-theorem-markovian} can be found in Appendix~\ref{appendix:proof-thm-markovian}. It establishes the iteration complexity of $\cO({\log^2 K}/{\sqrt{K}})$, or equivalently, sample complexity of $\tilde{\cO}(\varepsilon^{-2})$ for Algorithm~\ref{algorithm:dec-ac-re}. Note that actors, critics, and reward estimators use the step size of the same order. The rate matches the 
state-of-the-art sample complexity of decentralized AC algorithms up to a logarithmic term, which are implemented in double-loop fashion~\citep{hairi2021finite, chen2021sample}. The approximation error is defined as
\begin{align*}
\varepsilon_{app} := \max_{\theta, a} 	{\EE_{s \sim \mu_\theta}\left[\left|V_{\pit}(s)-\hat{V}_{\omega^*(\theta)}(s)\right|^2 + \left|\bar{r}(s,a) -  \hat{r}_{\lambda^*(\theta)}(s,a)\right|^2\right]}.
\numberthis \label{eq:approximation-error-def}
\end{align*}
The error $\varepsilon_{app}$ captures the approximation power of critic and reward estimator. {When using function approximation, such an error is inevitable.} Similar terms also appear in the literature (see, e.g.,~\citep{xu2020improving, agarwal2020optimality, qiu2019finite}). $\varepsilon_{app}$ becomes zero in tabular case.
{The error $\varepsilon_{sp}$ represents the mismatch between the discounted state visitation distribution $d_{\pi_\theta}$ and stationary distribution $\mu_{\pi_\theta}$. It is defined as 
\[\varepsilon_{sp}:=4C_{\theta}^2 \left(\log_\rho\kappa^{-1} + \frac{1}{\rho}\right)^2 \left(1-\gamma\right)^2. \label{eq:sp-error-def}\]
By policy gradient theorem (\ref{eq:policy-gradient-MARL}), the states should be sampled from discounted state visitation distribution in order to attain unbiased estimation of policy gradient. Nevertheless, the state distribution converges to stationary distribution $\mu_{\pi_\theta}$ due to Markov chain's mixing, which inevitably introduces the sampling error $\varepsilon_{sp}$. Similar terms also appear in \citep{ zeng2022learning, shen2020asynchronous}. When $\gamma$ is close to 1, the error becomes small. This is because $d_{\pi_\theta}$ approaches to $\mu_{\pi_\theta}$ when $\gamma$ goes to 1. In the literature, some works assume that sampling from $d_{\pi_\theta}$ is permitted, thus eliminate this error; see, e.g.,~\citep{ chen2021closing}.}

% The error $\varepsilon_{sp}$ is inevitably caused by the mismatch between discounted state visitation distribution $d_{\pi_\theta}$ and stationary distribution $\mu_{\pi_\theta}$; see, e.g.,~\citep{zeng2022learning, shen2020asynchronous}. It is defined as 
% \[\varepsilon_{sp}:=4C_{\theta}^2 \left(\log_\rho\kappa^{-1} + \frac{1}{\rho}\right)^2 \left(1-\gamma\right)^2. \label{eq:sp-error-def}\]
% When $\gamma$ is close to 1, the error becomes small. This is because $d_{\pi_\theta}$ approaches to $\mu_{\pi_\theta}$ when $\gamma$ goes to 1. In the literature, some works assume that sampling from $d_{\pi_\theta}$ is permitted, thus eliminate this error; see, e.g.,~\citep{ chen2021closing}.

\textbf{{Complexity result} under i.i.d. sampling.} Under the i.i.d. sampling scheme, state can be directly sampled from $\mu_{\pi_\theta}$ and $d_{\pi_\theta}$. In this case, the logarithmic term caused by the Markovian mixing time, and the error $\varepsilon_{sp}$ caused by the distribution mismatch, can be avoided. In this sense, one can attain the iteration complexity of $\cO(1/\sqrt{K})$, or equivalently, sample complexity of $\cO(\varepsilon^{-2})$.

% and the distribution mismatch error $\varepsilon_{sp}$ appeared in \ref{eq:main-convergence-markovian} can be avoided.
% We provide the formal theorem statement and the corresponding proof sketch in Appendix~\ref{} \qijun{TODO}.

% This is the first finite-time convergence result for decentralized single-timescale single-loop AC algorithm under practical Markovian sampling.

% However, to the best of our knowledge, all of the results with $\tilde{\cO}(\varepsilon^{-2})$ sample complexity are based on double-loop update, in which each update of actor follows a mini-batch update of critic variables. 
% which are essentially different from our algorithm. Furthermore, we note~\citep{zeng2022learning} analyzed the decentralized AC with alternative update. Nevertheless, they only attain the sample complexity of $\cO(\varepsilon^{-\frac{5}{2}})$. The essential reason for the sub-optimal rate is that their analysis requires the ratio between step sizes of actor and critic converge to zero asymptotically. In our proof, we keep all the step sizes in the same order and attain a better sample complexity.

\subsection{Sample complexity for Algorithm~\ref{algorithm:dec-ac-noisy}}

\begin{theorem}\label{thm:main-theorem-markov-noisy}
	Suppose Assumptions~\ref{assump:boundedness}-\ref{assump:doub_stoch} hold. Consider the update of Algorithm~\ref{algorithm:dec-ac-noisy} under Markovian sampling. Let $\alpha_k = \frac{\bar{\alpha}}{\sqrt{K}}$ for some positive constant $\bar{\alpha}$ and $\beta_k = \frac{C_9}{2\lambda_\phi}\alpha_k$, $K_r=\log(K^{1/2})$, $K_c \leq \cO({K^{1/4}})$, where $K$ is the total number of iterations. Then, we have
	\begin{align*}
	&\frac{1}{K} \sum_{k=1}^{K} \sum_{i=1}^{N}\EE\left[\|\omega_{k}^{i}-\omega^*(\theta_k)\|^2\right] \leq \cO\left(\frac{\log^2K}{\sqrt{K}}\right) \\
	&\frac{1}{K} \sum_{k=1}^{K} \sum_{i=1}^{N} \EE \left[\| \nabla_{\theta^i} {J}(\theta_k) \|^2\right] \leq \mathcal{O}\left(\frac{\log^2 K}{\sqrt{K}}\right)+ \cO(\varepsilon_{app}^c+\varepsilon_{sp}), \numberthis
	\end{align*}
	where the constants are defined in proof.
\end{theorem}

The proof of Theorem~\ref{thm:main-theorem-markov-noisy} can be found in Appendix~\ref{appendix:proof-thm-noisy}. It establishes the sample complexity of $\tilde{\cO}(\varepsilon^{-2})$ for Algorithm~\ref{algorithm:dec-ac-noisy}. The $\varepsilon_{app}^{c}$ captures the approximation error of the critic variables, which is defined as 
\[\varepsilon_{app}^{c} := \max_\theta \EE_{s\sim \mu_\theta}\left[\left|V_{\pi_\theta}(s) - \hat{V}_{\omega^*(\theta)}(s) \right|^2 \right].\]

The Algorithm~\ref{algorithm:dec-ac-noisy} preserves the privacy of local actions and requires less parameters than Algorithm~\ref{algorithm:dec-ac-re} since there is no reward estimator. The cost is that it needs to communicate $\cO(\log(\varepsilon^{-1}))$ times for each iteration.

\subsection{Proof sketch}
We present the main elements for the proof of Theorem~\ref{thm:main-theorem-markovian}, which helps in understanding the difference between classical two-timescale/double-loop analysis and our single-timescale analysis. The proof of Theorem \ref{thm:main-theorem-markov-noisy} follows the similar framework.

Under Markovian sampling, it is possible to show the following inequality, which characterizes the ascent of the objective.
%the ascent of the objective can be characterized by the following inequality
%Based on the smoothness property of the objective function implied by Assumption~\ref{assump:lipschitz-policy}, we have
\begin{align*}
\EE[J(\theta_{k+1})] - J(\theta_k) 
&\geq \sum_{i=1}^{N} \left[\frac{\alpha_k}{2}\EE\|\nabla_{\theta^i}J(\theta_k)\|^2 + \frac{\alpha_k}{2} \EE \|g_a^i(\xi_k, \omega_{k+1}^i, \lambda_{k+1}^i)\|^2 \right. \\
&\quad \left. - 8 C_\psi^2\alpha_k \EE\| \omega^*(\theta_k) - \omega_{k+1}^i\|^2 - 4 C_\psi^2\alpha_k\EE\|\lambda^*(\theta_k) - \lambda_{k+1}^{i}\|^2 \vphantom{\frac{\alpha_k}{2}} \right] \\	
&\quad - \cO(\log^2(K)\alpha_k^2) - \cO((\varepsilon_{app} + \varepsilon_{sp})\alpha_k).\label{eq:actor-error-main} \numberthis
\end{align*}

To analyze the errors of critic $\|\omega^*(\theta_k) - \omega_{k+1}^{i}\|^2$ and reward estimator $\|\lambda^*(\theta_k) - \lambda_{k+1}^{i}\|^2$, the two-timescale analysis requires $\cO(\alpha_k) < \min\{\cO(\beta_k), \cO(\eta_k)\}$ in order for these two errors to converge. The double-loop approach runs lower-level update for $\cO(\log(\varepsilon^{-1}))$ times with batch size $\cO(\varepsilon^{-1})$ to drive these errors below $\varepsilon$ and hence, they cannot allow inner loop size and bath size to be $\cO(1)$ simultaneously. To obtain the convergence result for \emph{single-timescale} update, the idea is to further upper bound these two lower-level errors by the quantity $\cO(\alpha_k\EE \|g_a^i(\xi_k, \omega_{k+1}^i, \lambda_{k+1}^i)\|^2)$ (through a series of derivations), and then eliminate these errors by the ascent term $\frac{\alpha_k}{2}\EE \|g_a^i(\xi_k, \omega_{k+1}^i, \lambda_{k+1}^i)\|^2$. 

We mainly focus on the analysis of critic's error through the proof sketch. The analysis for reward estimator's error follows similar procedure.
% and we will present the detail in Appendix~\ref{appendix:reward-estimator-error}. 
We start by decomposing the error of critic as 
\begin{align*}
\sum_{i=1}^{N}\|\omega_{k+1}^{i} - \omega^*(\theta_k)\|^2 = \sum_{i=1}^{N} (\|\omega_{k+1}^{i} - \bar{\omega}_{k+1}\|^2 + \|\bar{\omega}_{k+1} - \omega^*(\theta_k)\|^2). \label{eq:critic-error-decomp-main} \numberthis
\end{align*}

The first term represents the consensus error, which can be bounded by the next lemma.
\begin{lemma}\label{lemma:consensus-error-main}
	Suppose Assumptions \ref{assump:boundedness} and \ref{assump:doub_stoch} hold. Consider the sequence $\{\omega_k^i\}$ generated by Algorithm~\ref{algorithm:dec-ac-re}, then the following holds
	%	\begin{align*}
	%	\sum_{i=1}^{N} \| \omega_k^i - \bar{\omega}_k \|^2 &\leq \nu^{2k} \| \bomega_0 \|_F +    \frac{16NC_\delta^2}{1-\nu}\beta_k^2 \\
	%	&\quad + \frac{8\sqrt{N}C_\delta\|\bomega_0\|_F}{1-\nu} \nu^k \beta_k, \numberthis \label{eq:consensus-error-main}
	%	\end{align*}
	\begin{align*}
	\|Q\bomega_{k+1}\| \leq \nu^{\frac{k}{K_c}-1}\|\bomega_0\| + 4\sqrt{N}C_\delta \sum_{t=0}^{k}\nu^{\frac{k-t}{K_c}-1}\beta_t,
	\end{align*}
	where {\normalfont$\bomega_k:=[\omega_k^1, \cdots, \omega_k^N]^T, Q:=I-\frac{1}{N}\bone\bone^T$}, $\nu\in(0,1)$ is the second largest singular value of $W$.
\end{lemma}

Based on Lemma~\ref{lemma:consensus-error-main} and follow the step size rule of Theorem~\ref{thm:main-theorem-markovian}, it is possible to show 
$\|Q\bomega_{k+1}\|^2=\sum_{i=1}^{N}\|\omega_{k+1}^{i}-\bar{\omega}_{k+1} \|^2=\cO(K_c^2\beta_k^2)$. Let $K_c=\cO(\beta_k^{-\frac{1}{2}})$, we have $\|Q\bomega_{k+1}\|^2= \cO(\beta_k)$, which maintains the optimal rate.

To analyze the second term in (\ref{eq:critic-error-decomp-main}), 
we first construct the following Lyapunov function
\begin{align*}
\VV_k:= -J(\theta_k) + \|\bar{\omega}_k - \omega^*(\theta_k)\|^2 + \|\bar{\lambda}_k - \lambda^*(\theta_k)\|^2.
\numberthis \label{eq:Lyapunov-main}
\end{align*}

%Then, the difference between Lyapunov functions will be
%\begin{align*}
%\VV_{k+1} - \VV_{k} &= J(\theta_{k+1}) - J(\theta_k) \\
%&\quad + \|\bar{\omega}_{k+1} - \omega^*(\theta_{k+1})\|^2 - \|\bar{\omega}_k - \omega^*(\theta_k)\|^2 \\
%&\quad + \|\bar{\lambda}_{k+1} - \lambda^*(\theta_{k+1})\|^2 - \|\bar{\lambda}_k - \lambda^*(\theta_k)\|^2. \numberthis
%\label{eq:lyapunov-difference-main}
%\end{align*}

Then, it remains to derive an approximate descent property of the term $\|\bar{\omega}_k - \omega^*(\theta_k)\|^2$ in (\ref{eq:Lyapunov-main}). Towards that end, our key step lies in establishing the \emph{smoothness of the optimal critic variables} shown in the next lemma.

%The key idea is to utilize the approximate descent property of $\|\bar{\omega}_k - \omega^*(\theta_k)\|^2$ to compensate the second term in (\ref{eq:critic-error-decomp-main}). The approximate descent property is established based on the smoothness of optimal critic variables.

\begin{lemma}[Smoothness of optimal critic]\label{lemma:opt-critic-smoothness}
	Suppose Assumptions~\ref{assump:boundedness}-\ref{assump:lipschitz-policy} hold, under the update of Algorithm~\ref{algorithm:dec-ac-re}, there exists a positive constant $L_{\omega, 2}$ such that for any policy parameter $\theta_1, \theta_2$, it holds that
	\begin{align*}
	\|\nabla \omega^*(\theta_1) - \nabla \omega^*(\theta_2)\| \leq L_{\omega, 2} \|\theta_1 - \theta_2\|,
	\end{align*}
% 	where $\nabla \omega^*(\theta)$ denotes the Jacobian of $\omega^*(\theta)$ with respect to $\theta$. \qijun{to verify}
\end{lemma}
This smoothness property is essential for achieving our $\tilde\cO(1/\sqrt{K})$ convergence rate.

To the best of our knowledge, the smoothness of $\omega^*(\theta)$ has not been justified in the literature.  Equipped with Lemma \ref{lemma:opt-critic-smoothness}, we are able to establish the 
following lemma.

\begin{lemma}[Error of critic]\label{lemma:critic-error-main}
	Under Assumptions~\ref{assump:boundedness}-\ref{assump:doub_stoch}, consider the update of Algorithm~\ref{algorithm:dec-ac-re}. Then, it holds that
	\begin{align*}
	\EE[\|\bar{\omega}_{k+1} - \omega^*(\theta_{k+1})\|^2]
	&\leq (1 + C_{9}\alpha_k) \|\bar{\omega}_{k+1} - \omega^*(\theta_k)\|^2 \\
	&\quad + \frac{\alpha_k}{4}\sum_{i=1}^{N}\|\EE[g_a^i(\xi_k, \omega_{k+1}^i, \lambda_{k+1}^i)]\|^2 + \cO(\alpha_k^2). \numberthis
	\label{eq:critic-error-1-main} \\
	\EE[\|\bar{\omega}_{k+1} - \omega^*(\theta_k)\|^2] &\leq (1-2\lambda_{\phi} \beta_k) \| \bar{\omega}_k - \omega^*(\theta_k) \|^2 \\
	&\quad + C_{K_1} \beta_k\beta_{k-Z_K} + C_{K_2}\alpha_{k-Z_K} \beta_k. \numberthis \label{eq:critic-error-2-main}
	\end{align*}
	
	Here, $Z_K:= \min\{z \in \NN^+ | \kappa\rho^{z-1} \leq \min \{\alpha_K, \beta_K, \eta_K\}\}$, $C_9$, $\lambda_{\phi}$ are constants specified in appendix, and $C_{K_1}$ and $C_{K_2}$ are of order $\cO(\log(K))$ and $\cO(\log^2(K))$ respectively.
\end{lemma}

Plug (\ref{eq:critic-error-2-main}) into (\ref{eq:critic-error-1-main}), we can establish the approximate descent property of $\|\bar{\omega}_k - \omega^*(\theta_k)\|^2$ in (\ref{eq:Lyapunov-main}):
\begin{align*}
\EE[\|\bar{\omega}_{k+1} - \omega^*(\theta_{k+1})\|^2] 
&\leq (1 +  C_9 \alpha_k)(1-2\lambda_{\phi} \beta_k) \| \bar{\omega}_k - \omega^*(\theta_k) \|^2 \\
&\quad + \frac{\alpha_k}{4}\sum_{i=1}^{N}\|\EE[g_a^i(\xi_k, \omega_{k+1}^i, \lambda_{k+1}^i)]\|^2 + \cO(C_{K_1} \beta_k\beta_{k-Z_K} + C_{K_2}\alpha_{k-Z_K} \beta_k). \numberthis \label{eq:approx-descent-main}
\end{align*}
Finally, plugging (\ref{eq:actor-error-main}), (\ref{eq:critic-error-1-main}), and (\ref{eq:approx-descent-main}) into (\ref{eq:Lyapunov-main}) gives the ascent of the Lyapunov function, which leads to our convergence result through steps of standard arguments.

% {\qijun{I would suggest to delete the part below, since the reviewer's confusion has been addressed and the extension is natural}}

\textbf{Remarks on update step.} In Algorithms~\ref{algorithm:dec-ac-re} and \ref{algorithm:dec-ac-noisy}, the actor and critic update once for each iteration. This update scheme can be generalized to the case where actor and critic update arbitrary number of constant steps without affecting the order of the sample complexity. In particular, suppose that actor updates $C_a$ steps per iteration, and let $g_{a, k}^i$ be the actor's update direction at iteration $k$. The bounds (\ref{eq:actor-error-main}) and (\ref{eq:approx-descent-main})  become \begin{align*}
\EE[J(\theta_{k+1})] - J(\theta_k) 
&\geq \sum_{i=1}^{N} \left[\frac{\alpha_k}{2}\EE\|\nabla_{\theta^i}J(\theta_k)\|^2 + \frac{\alpha_k}{2} \EE \|g_{a,k}^i\|^2 - 8 C_\psi^2\alpha_k \EE\| \omega^*(\theta_k) - \omega_{k+1}^i\|^2 \right. \\
&\quad \left. {} - 4 C_\psi^2\alpha_k\EE\|\lambda^*(\theta_k) - \lambda_{k+1}^{i}\|^2 \right] - \cO(C_a^2\log^2(K)\alpha_k^2) - \cO((\varepsilon_{app} + \varepsilon_{sp})\alpha_k)
\end{align*}
\begin{align*}
\EE[\|\bar{\omega}_{k+1} - \omega^*(\theta_{k+1})\|^2] 
&\leq (1 +  C_9 \alpha_k)(1-2\lambda_{\phi} \beta_k) \| \bar{\omega}_k - \omega^*(\theta_k) \|^2 \\
&\quad + \frac{\alpha_k}{4}\sum_{i=1}^{N}\|\EE[g_{a,k}^i]\|^2 + \cO(C_{K_1} \beta_k\beta_{k-Z_K} + C_aC_{K_2}\alpha_{k-Z_K} \beta_k),
\end{align*}
where we replace the norm bound $\alpha_k\|g_a^i(\xi_k, \omega_{k+1}^i, \lambda_{k+1}^i)\|=\cO(\alpha_k)$ with $\|g_{a,k}^i\|$ and apply Cauchy-Schwartz inequality: $\|g_{a,k}^i\| \leq C_a \|g_a^i(\xi_k, \omega_{k+1}^i, \lambda_{k+1}^i)\|$.
When $C_a$ is a constant that is not related to $K$, these two bounds recovers (\ref{eq:actor-error-main}) and (\ref{eq:approx-descent-main}). Hence, we can follow exactly the same proof procedure and obtain the $\tilde\cO(\varepsilon^{-2})$ sample complexity result as before.
When critic update $C_c > 1$ steps per iteration, the expected temporal difference error will decrease for each step by controlling step size, so that the bound in (\ref{eq:approx-descent-main}) still holds. Thus, updating critic for multiple steps will not affect the sample complexity.

%Due to the space limit, we will omit the error analysis of reward estimator. We remark that it follows similar procedure as critic. Please refer to Appendix~\ref{appendix:reward-estimator-error} for detailed analysis.

% \vspace{-0.5cm}
% \begin{figure}[t]
% \centering
% \begin{subfigure}
%     \centering
%     \includegraphics[width=0.4\textwidth]{figures/sample_complexity.png}
%     \caption{Comparison between single-loop update and double-loop update. The result is averaged over 20 Monte Carlo runs.}
% \end{subfigure}
% \label{fig:comparison}
% \end{figure}
%\vspace{-0.3cm}

\subsection{Convergence of single-timescale decentralized NAC}
The natural Actor-Critic (NAC)~\citep{peters2008natural} is a popular variant of AC algorithm, which enjoys the convergence to a global optimum (with compatible function approximation error) instead of a local stationary point. While our main focus is the convergence of the single-timescale AC algorithm, we find that the proof technique can be directly extended to establish the global convergence of  single-timescale decentralized NAC. For reference, we design such an algorithm and provide its convergence result in Appendix~\ref{appendix:nac-convergence} as a by-product of our single-timescale AC's analysis. To the best of our knowledge, this is the first convergence result of single-timescale NAC. However, our analysis only establishes a $\cO(\varepsilon^{-6})$ rate for the algorithm. This result is sub-optimal compared with the existing best complexity of $\cO(\varepsilon^{-3})$~\citep{chen2021sample}, which is based on the double-loop implementation. The main reason for the sub-optimality is that in comparison with the double-loop update, the critic variables under the single-timescale update will inevitably converge slower due to the change of the actor's parameter in each iteration. Based on the classical NAC's analysis, the slower convergence of critic variables will result in a worse convergence rate of the optimality gap. Please refer to Appendix~\ref{appendix:nac-convergence} for more discussions on the sub-optimality.

% For reference, we also extend our analysis to decentralized single-timescale NAC as a by-product of our analysis on the single-timescale AC. The full convergence result is provided in Appendix~\ref{appendix:nac-convergence}. 

%\vspace{-0.3cm}
\begin{figure}[t]
	\centering
	\begin{subfigure}[b]{0.45\textwidth}
		\centering
	\includegraphics[width=\textwidth]{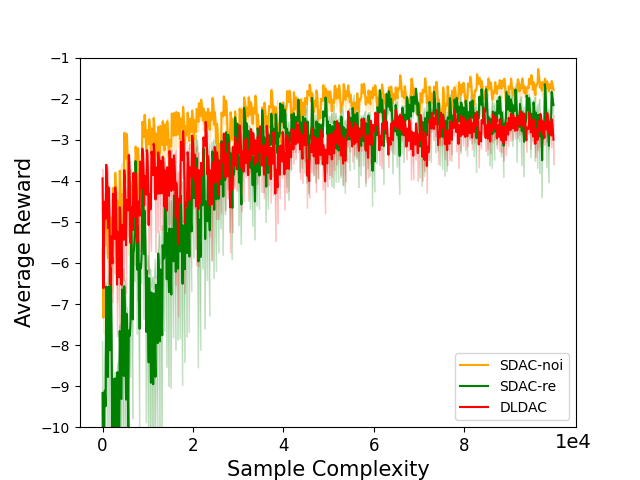}
		%  \caption{$y=x$}
	\end{subfigure}
	\hfill
	\begin{subfigure}[b]{0.45\textwidth}
		\centering
		\includegraphics[width=\textwidth]{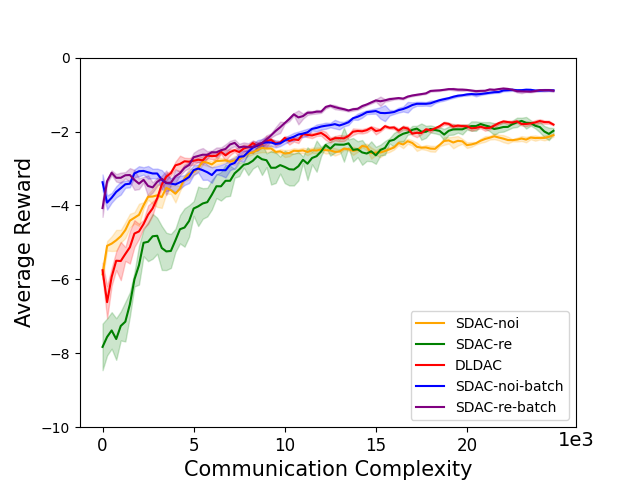}
		%  \caption{$y=3sinx$}
	\end{subfigure}
	\caption{Averaged reward versus sample complexity and communication complexity. The vertical axis is the averaged reward over all the agents. The result is averaged over 10 Monte Carlo runs.}
	\label{fig:simulation-main}
\end{figure}
% \vspace{-0.5cm}

% \vspace{-0.2cm}
\section{Numerical results}

\subsection{Experiment setting}
We adopt the grounded communication environment proposed in~\citep{mordatch2018emergence}. Our task consists of $N$ agents and the corresponding $N$ landmarks inhabited in a two-dimension world, where each agent can observe the relative position of other agents and landmarks. For every discrete time step, agents take actions to move along certain directions, and receive their rewards. Agents are rewarded based on the distance to their own landmark, and penalized if they collide with other agents. The objective is to maximize the long-term averaged reward over all agents. Since we focus on decentralized setting, each agent shall not know the target landmark of others, i.e., the reward function of others. To exchange information, each agent is allowed to send their local information via a fixed communication link. Through all the experiments, the agent number $N$ is set to be 5, and the discount factor $\gamma$ is set to be $0.95$.

% \vspace{-0.2cm}
\subsection{Comparison with existing decentralized AC algorithms}
In this section, we compare the proposed algorithm with existing decentralized AC algorithms under the cooperative MARL setting~\citep{chen2021sample, zeng2022learning} in terms of sample complexity and communication complexity. 
%Here, we present the comparison with the double-loop based implementation~\citep{chen2021sample}. The comparison with two-timescale based implemenation~\citep{zeng2022learning} is deferred to Appendix~\ref{appendix:experiment}. 
In the sequel, we refer  Algorithm~\ref{algorithm:dec-ac-re} as "SDAC-re" and  Algorithm~\ref{algorithm:dec-ac-noisy} as "SDAC-noi" (see Appendix~\ref{algorithm:dec-ac-noisy}). 
%We also provide the result which assumes full reward is available to serve as baseline, which we refer as "SDAC-full". 
The algorithm proposed in~\citep{chen2021sample} is referred as "DLDAC", which is based on double-loop implementation. 
The algorithm proposed in \citep{zeng2022learning} is denoted by "TDAC-re", which is based on two-timescale step size implementation. For comparison, we also implement a noisy reward version of "TDAC-re" and denote it by "TDAC-noi".

\textbf{Comparison to double-loop decentralized AC.} 
For "SDAC-re" and "SDAC-noi", we set $\alpha_k=0.01(k+1)^{-0.5}$, $\beta_k=0.1(k+1)^{-0.5}, \eta_k=0.1(k+1)^{-0.5}, K_c=5, \sigma=0.5, K_r=2$. For "DLDAC", we fix $T_c=50$, $T_c'=10$, $T'=5$, $N_c=10$, $N=100$, $\sigma=0.1$ \footnote{Note that we adopt the notations in \citep{chen2021sample}. Here, $T_c$ is the inner loop size, $T_c'$ is the communication number for each outer loop, $T'$ is the communication number for reward consensus, $N$ is the batch size for actor's update, and $N_c$ is the batch size for critic's update.}, which is adopted by their paper (see comparisons under different hyper-parameters in Appendix~\ref{appendix:experiment}). We set $\alpha=0.01, \beta=0.1$ for "DLDAC" since we observe that larger step sizes will result in divergence. We have to mention that such a inner loop size $T_c=50$ in "DLDAC" is not necessarily consistent with the theory of a {double-loop} algorithm, in which the loop size should be proportional to $\cO(\varepsilon^{-1})$.
The sample complexity and communication complexity results are shown in Figure~\ref{fig:simulation-main}. For the sample complexity, "SDAC-noi" enjoys a faster convergence compared with "DLDAC". In terms of communication complexity, "DLDAC" achieves better performance as it applies mini-batch technique and thereby requires less communication rounds when using the same amount of samples. Such a mini-batch approach can also be adopted to our proposed algorithms. Thus, we implement a mini-batch version of our proposed  algorithms, which we refer as "SDAC-noi-batch" and "SDAC-re-batch", respectively. We set 10 as the batch size for actor, critic, and reward estimator. We can see that by applying mini-batch update, these two variants achieve significantly better communication complexity compared with "DLDAC". This is because our algorithm updates actor for more times compared with "DLDAC" under the same communication rounds.

% \vspace{-0.5cm}

\begin{figure}[h!]
	\centering
	\includegraphics[width=0.45\textwidth]{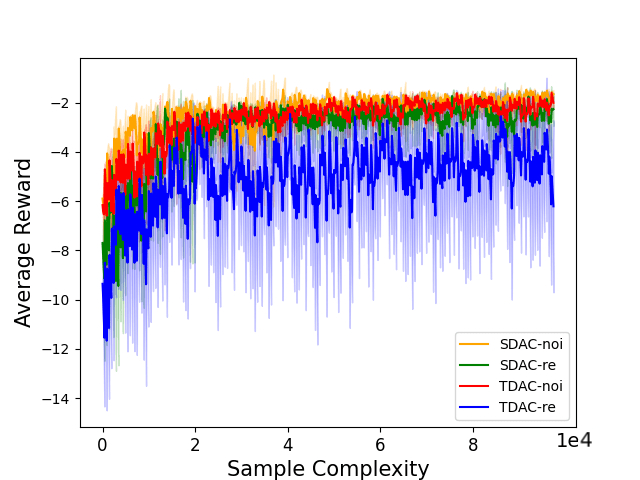}
	\caption{Comparison between the proposed algorithms and two-timescale decentralized AC algorithms~\citep{zeng2022learning}. The results are averaged over 10 Monte Carlo runs.}
	\label{fig:comparison-two-timescale}
\end{figure}
% \vspace{-0.5cm}
\textbf{Comparison with two-timescale decentralized AC.} We fix $K_c=1$, $K_r=5$ for this experiment. We set $\alpha_k=0.01(k+1)^{-0.5}$, $\beta_k=0.1(k+1)^{-0.5}$, and $\eta_k=0.1(k+1)^{-0.5}$ for "SDAC-re" and "SDAC-noi"; we set $\alpha_k=0.01(k+1)^{-0.6}$, $\beta_k=0.1(k+1)^{-0.4}$, and $\eta_k=0.1(k+1)^{-0.4}$ for "TDAC-re" and "TDAC-noi". The sample complexity is presented in Figure~\ref{fig:comparison-two-timescale}. We can observe that the convergence speed of "SDAC-noi" is slightly better than that the two-timescale counterpart "TDAC-noi". In addition, when using reward estimator for the global reward estimation, we see that "SDAC-re" has much more stable convergence behavior than "TDAC-re", and achieves significantly higher rewards.
\vspace{-0.3cm}

%\qijun{re-write the interpretation and add comparison with TDAC in appendix}.
%The sample complexity and communication complexity are shown in Figure~\ref{fig:simulation-main}. The results are averaged over 10 Monte Carlo runs. 
%As we can see, the proposed two algorithms achieve significantly higher reward than "DLDAC" in terms of both sample complexity and communication complexity. Moreover, their performances approach the baseline “SDAC-full", where the global reward is assumed to be available, indicating that the reward approximation is nearly accurate. Due to space limit, we will put additional experiments on the comparison with existing decentralized AC algorithms and the ablation study of hyper-parameters to Appendix~\ref{appendix:experiment}.
\begin{figure}[h]
	\centering
	\begin{subfigure}[b]{0.45\textwidth}
		\centering
		\includegraphics[width=\textwidth]{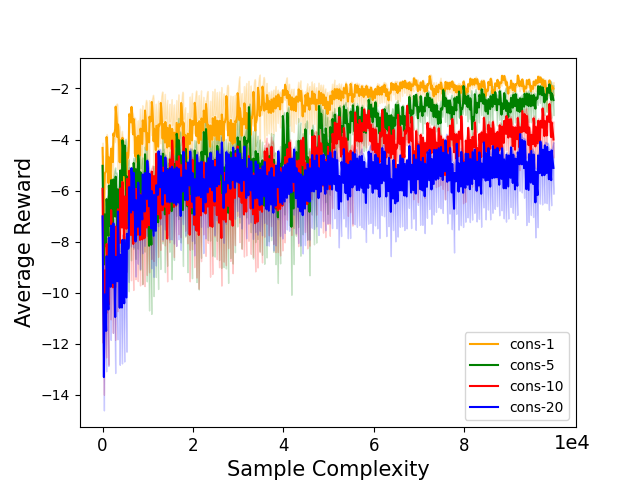}
		%		\caption{The sample complexity.}
		\label{fig:ablation-sample-main}
	\end{subfigure}
	\hfill
	\begin{subfigure}[b]{0.45\textwidth}
		\centering
		\includegraphics[width=\textwidth]{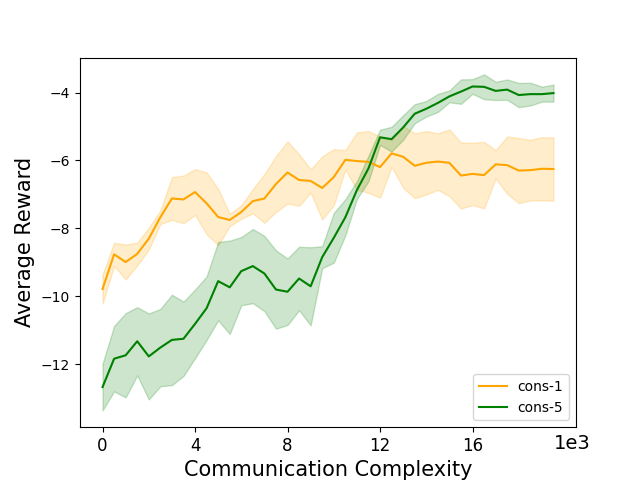}
		%		\caption{The communication complexity.}
		\label{fig:ablation-communication-main}
	\end{subfigure}
	\caption{Ablation study on the consensus periods. The results are averaged over 10 Monte Carlo runs.}\label{fig:ablation-main}
\end{figure}

% \vspace{-0.5cm}
\subsection{Ablation study on different choices of $K_c$}
\vspace{-0.1cm}
We compare the performance of "SDAC-noi" under different choices of consensus periods $K_c$. In particular, we set $\alpha_k=0.01(k+1)^{-0.5}$, $\beta_k=0.1(k+1)^{-0.5}$, $K_r=2$, $\sigma=0.5$ and examine the consensus periods $K_c$ of 1, 5, 10, and 20, respectively.
The corresponding sample complexity and communication complexity results are summarized in Figure~\ref{fig:ablation-main}.
Evidently, in terms of sample complexity,  the convergence becomes slower and relatively unstable as the consensus period $K_c$ increases. Therefore, when the communication cost is low, choosing a small $K_c$ will yield a better performance.  We also plot the communication complexity under the consensus periods of $1$ and $5$. We can see that the communication complexity of "cons-5" outperforms "cons-1" after $12\times 10^3$ communications. Thus, when the communication cost  is expensive and high averaged reward is required, one may use large $K_c$ and run the algorithm for a relatively large number of iterations. 

\vspace{-0.3cm}
\section{Conclusion and future direction}\label{section:conclusion}
\vspace{-0.2cm}
In this paper, we studied the convergence of fully decentralized AC algorithm under practical single-timescale update. We showed that the algorithm will maintain the optimal sample complexity of $\tilde \cO(\varepsilon^{-2})$ and is communication efficient. We also proposed a variant 
to preserve the privacy of local actions by communicating noisy rewards. Extensive simulation results demonstrate the superiority of our algorithms' empirical performance over existing decentralized AC algorithms. However, directly extending our single-timescale AC's analysis technique to single-timescale NAC will result in a sub-optimal sample complexity. We leave the study on improving the convergence rate and design a more efficient single-timescale NAC algorithm as promising future directions.

\section*{Acknowledgement}
The authors would like to thank the Action Editor and anonymous reviewers for their detailed and constructive comments, which have helped greatly to improve the quality and presentation of the manuscript.
	
X. Li was partially supported by the National Natural Science Foundation of China (NSFC) under Grant No. 12201534 and 72150002, by the Shenzhen Science and Technology Program under Grant No. RCBS20210609103708017, and by the Shenzhen Institute of Artificial Intelligence and Robotics for Society (AIRS) under Grant No. AC01202101108.
% We leave the study on improving the sample complexity of single-timescale NAC as a future work.

% One limitation of our work is that we only study the convergence to stationary point. Thus, we leave the research on the avoidance of saddle points and convergence to global optimums as promising future directions.

% Since our study is limited to the convergence to the stationary point, we leave the research on the global convergence as a future direction.

\clearpage
\bibliographystyle{tmlr}
\bibliography{ac_tmlr.bib}	

\clearpage
\appendix

\tableofcontents
\clearpage

\begin{figure}[t]
	\centering
	\begin{subfigure}[b]{0.48\textwidth}
		\centering
		\includegraphics[width=\textwidth]{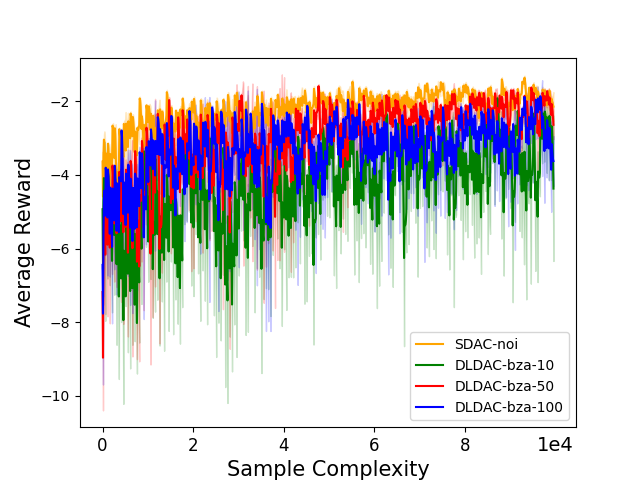}
		\caption{Different actor's batch sizes.}
		\label{fig:comparison-bza}
	\end{subfigure}
	\hfill
	\begin{subfigure}[b]{0.48\textwidth}
		\centering
		\includegraphics[width=\textwidth]{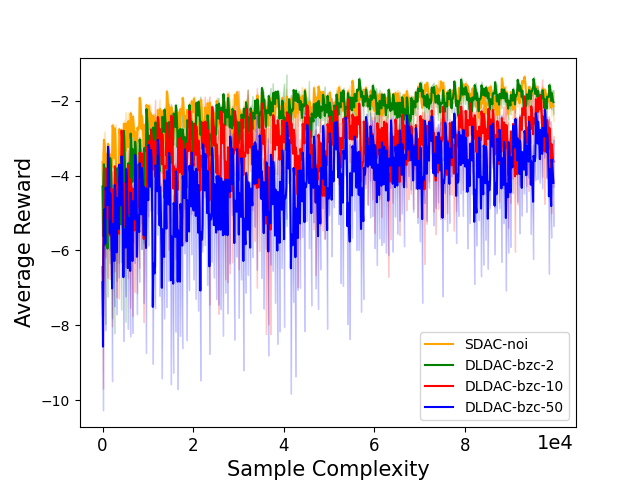}
		\caption{Different critic's batch sizes.}
		\label{fig:comparison-bzc}
	\end{subfigure}
	\hfill
	\begin{subfigure}[b]{0.48\textwidth}
		\centering
		\includegraphics[width=\textwidth]{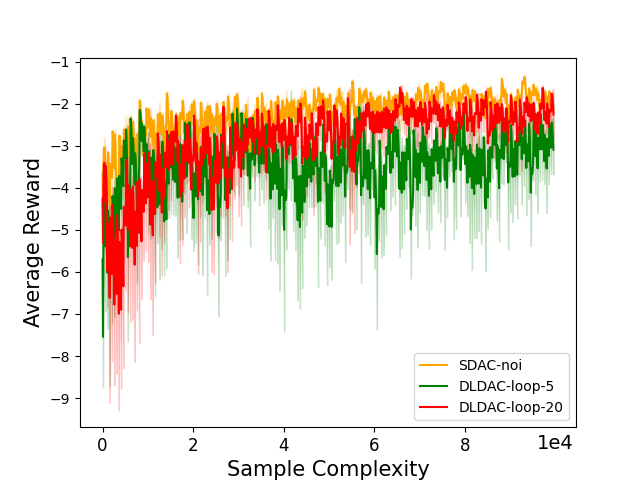}
		\caption{Different loop sizes.}
		\label{fig:comparison-loop-size}
	\end{subfigure}
	\caption{Comparison between the proposed algorithms and the double-loop decentralized AC algorithm that uses mini-batch update. The results are averaged over 10 Monte Carlo runs.}
	\label{fig:comparison-double-loop}
\end{figure}
\section{Additional simulation results}\label{appendix:experiment}
%In this section, we first introduce the experimental setting. Then, we present more experiments on the comparison between the proposed algorithms and existing decentralized AC algorithms. Additionally, we conduct ablation study on different consensus frequencies of the proposed algorithm.

% \begin{figure}[t]
% 	\centering
% 	\includegraphics[width=0.5\textwidth]{figures/ma_5agent_100bz.png}
% 	\caption{Simulation over large inner loop size. The performance is averaged over 20 Monte Carlo runs.}
% 	\label{fig:100-inner}
% \end{figure}

%\textbf{Experiment setting.} We adopt the grounded communication environment proposed in~\citep{mordatch2018emergence}. Our task consists of $N$ agents and the corresponding $N$ landmarks inhabited in a two-dimension world, where each agent can observe the relative position of other agents and landmarks. For every discrete time step, agents take actions to move along certain directions, and receive their rewards. Agents are rewarded based on the distance to their own landmark, and penalized if they collide with other agents. The objective is to maximize the long-term averaged reward over all agents. Since we focus on decentralized setting, each agent shall not know the target landmark of others, i.e., the reward function of others. To exchange information, each agent is allowed to send their local information via a fixed communication link. Through all the experiments, the agent number $N$ is set to be 5, and the discount factor $\gamma$ is set to be $0.95$.

In this section, we provide more experiments which compare the proposed algorithms with double-loop based decentralized AC algorithm under different batch sizes and inner loop sizes.

%\textbf{Comparison to double-loop decentralized AC under mini-batch update.} Since the algorithm in~\citep{chen2021sample} uses mini-batch update to reduce the variance during the update, we will compare the proposed algorithms with~\citep{chen2021sample} under different choices of actor's batch sizes, critic's batch sizes, and inner loop sizes, respectively. Since their algorithm communicates noisy reward to achieve consensus, we will use "SDAC-noi" to serve as baseline.
\begin{enumerate}
	\item \textbf{Actor's batch size.} We fix $T_c=50$, $T_c'=10$, $N_c=10$, \footnote{Note that we adopt the notations in \citep{chen2021sample}. Here, $T_c$ is the inner loop size, $T_c'$ is the communication number for each outer loop, $N$ is the batch size for actor's update, and $N_c$ is the batch size for critic's update.} which is adopted by \citep{chen2021sample}. We examine values of $N$ in $\{10, 50, 100\}$. The results are in Figure~\ref{fig:comparison-bza}. We observe that the best choice of actor's batch size $N$ is $50$, and the proposed "SDAC-noi" converges faster than it in terms of sample complexity.
	\item \textbf{Critic's batch size.} We fix $T_c=50$, $T_c'=10$, $N=100$, which is adopted by \citep{chen2021sample}. We examine values of $N_c$ in $\{2, 10, 50\}$. The results are shown in Figure~\ref{fig:comparison-bzc}. As we can see, "DLDAC" with smaller critic's batch sizes can achieve better sample complexity, indicating that the variance of critic's update is relatively small and the mini-batch update is not needed for this task. Our proposed "SDAC-noi" achieves better convergence compared with the double-loop decentralized AC under different choices of $N_c$.
	\item \textbf{Inner loop size.} We fix $T_c'=10$, $N=100$, $N_c=10$, which is adopted by~\citep{chen2021sample}. We examine values of $T_c$ in $\{5, 20\}$. The results are shown in Figure~\ref{fig:comparison-loop-size}. We can see that the proposed "SDAC-noi" enjoys a better convergence in terms of sample complexity.
\end{enumerate}

\section{Auxiliary lemmas}\label{appendix:auxiliary-lemma}
In this section, we provide some auxiliary lemmas, which serves as the preliminary for the proof of main theorems and lemmas.

The following two lemmas present the Lipschitz properties of the objective function and value function.

% show that the objective is smooth, and the value function is Lipschitz continuous.

\begin{lemma}[\citep{zhang2019global}, Lemma 3.2]\label{lemma:smooth-objective}
Suppose Assumption~\ref{assump:lipschitz-policy} holds, then there exists a positive constant $L$ such that for any policy parameter $\theta_1$ and $\theta_2$, we have $\|\nabla J(\theta_1) - \nabla J(\theta_2)\| \leq L\|\theta_1-\theta_2\|$. 
\end{lemma}

\begin{lemma}[\citep{shen2020asynchronous}, Lemma 4] \label{lemma:lipschitz-continuous-value}
	Suppose Assumption \ref{assump:lipschitz-policy} holds, for any policy parameter $\theta_1, \theta_2$ and $s \in \cS$, there exits a positive constant $L_V$ such that
	\begin{align*}
	\| \nabla V_{\pi_{\theta_1}}(s) \| &\leq L_V \\
	| V_{\pi_{\theta_1}}(s) - V_{\pi_{\theta_2}}(s) | &\leq L_V\|\theta_1 - \theta_2\|.
	\end{align*}
\end{lemma}

The next lemma shows that the stationary distribution is Lipschitz continuous with respect to policy.
\begin{lemma}[\citep{wu2020finite}, Lemma B.1] \label{lemma:error-total-variation}
	For any policy parameter $\theta_1$ and $\theta_2$, it holds that 
	\begin{align*}
	d_{TV}(\mu_{\theta_1}, \mu_{\theta_2}) &\leq |\cA| L_\pi(\log_{\rho}\kappa^{-1}+(1-\rho)^{-1})\|\theta_1 - \theta_2\| \\
	d_{TV}(\mu_{\theta_1}\otimes\pi_{\theta_1}, \mu_{\theta_2}\otimes \pi_{\theta_2}) &\leq |\cA| L_\pi(1+\log_{\rho}\kappa^{-1}+(1-\rho)^{-1})\|\theta_1 - \theta_2\| \\
	d_{TV}(\mu_{\theta_1}\otimes\pi_{\theta_1}\otimes\cP, \mu_{\theta_2}\otimes \pi_{\theta_2}\otimes\cP) &\leq |\cA| L_\pi(1+\log_{\rho}\kappa^{-1}+(1-\rho)^{-1})\|\theta_1 - \theta_2\|.
	\end{align*}
	We will define $L_\mu:=|\cA| L_\pi(\log_{\rho}\kappa^{-1}+(1-\rho)^{-1})$ for the proof of main theorems and lemmas.
\end{lemma}

The following lemma characterizes the geometric mixing of the Markov chain.
\begin{lemma}[\citep{shen2020asynchronous}, Lemma 1]\label{lemma:mix-markov-chain}
	Suppose Assumption \ref{assump:markov-chain} holds, then there exists $\kappa > 0, \rho \in [0,1]$ such that for any policy parameter $\theta$ we have
	\begin{align*}
	\sup_{s_0 \in \cS} d_{TV}(\PP((s_k, a_k, s_{k+1}) \in \cdot | s_0, \pi_\theta), \mu_\theta \otimes \pi_\theta \otimes \cP) \leq \kappa \rho^{k},
	\end{align*}
	where $\mu_\theta$ is the stationary distribution induced by $\pi_\theta$ and transition kernel $\cP(\cdot|s,a)$.
\end{lemma}

The next lemma bounds the error of the discounted state-visitation distribution and the stationary distribution.
\begin{lemma}[\citep{shen2020asynchronous}, Lemma 2] \label{lemma:mismatch-visitation-stationary}
	Suppose Assumption~\ref{assump:markov-chain} holds, then for any policy parameter $\theta$, there exists $\kappa > 0, \rho \in [0,1]$ such that
	\[d_{TV}(d_{\pi_\theta}, \mu_{\pi_\theta}) \leq 2(\log_{\rho}{\kappa^{-1}} + \frac{1}{1-\rho})(1-\gamma).\]
\end{lemma}

The following lemma bounds the total variation distance between state distribution under a fixed policy and that under an updating policy. The lemma is used for analyzing the sampling error.
\begin{lemma} \label{lemma:mismatch-markov-stationary}
	Consider the Markov chain:
	\[s_{k-z}\xrightarrow{\theta_{k-z}}a_{k-z}\xrightarrow{\cP}s_{k-z+1}
	\xrightarrow{\theta_{k-z+1}} a_{k-z+1} \cdots \xrightarrow{\theta_{k-1}}a_{k-1} \xrightarrow{\cP}s_k\xrightarrow{\theta_k}a_k\xrightarrow{\cP}s_{k+1}.\]
	Also consider the auxiliary Markov chain with fixed policy:
	\[s_{k-z}\xrightarrow{\theta_{k-z}}a_{k-z}\xrightarrow{\cP}s_{k-z+1}
	\xrightarrow{\theta_{k-z}} \tilde{a}_{k-z+1} \cdots \xrightarrow{\theta_{k-z}}\tilde{a}_{k-1} \xrightarrow{\cP}\tilde{s}_k\xrightarrow{\theta_{k-z}}\tilde{a}_k\xrightarrow{\cP}\tilde{s}_{k+1}.\]
	
	Let $\xi_k:=(s_k, a_k, s_{k+1})$ be sampled from chain 1, and $\tilde{\xi}_k:=(\tilde s_k, \tilde a_k, \tilde s_{k+1})$ be sampled from chain 2. Then we have
	\begin{align*}
	d_{TV}(\PP(\xi_k \in \cdot| \theta_{k-z}, s_{k-z+1}), \PP(\tilde{\xi}_k \in \cdot | \theta_{k-z}, s_{k-z+1})) \leq \frac{1}{2} \sum_{m=0}^{z-1} | \cA | L_\pi \|\theta_{k-m} - \theta_{k-z}\|.
	\end{align*}
	
	\begin{proof}
		\begingroup
		\allowdisplaybreaks
		\begin{align*}
		&\quad d_{TV}(\PP(\xi_k \in \cdot), \PP(\tilde{\xi}_k \in \cdot)) \\
		&= \frac{1}{2} \int_{s\in \cS}\int_{s' \in \cS}\sum_{a\in \cA} | \PP(s_k=ds, a_k=a, s_{k+1}=ds') - \PP(\tilde{s}_k = ds, \tilde{a}_k=a, \tilde{s}_{k+1}= ds') | \\
		&=\frac{1}{2} \int_{s\in \cS}\sum_{a\in \cA}| \PP(s_k=ds, a_k=a) - \PP(\tilde{s}_k = ds, \tilde{a}_k=a) | \int_{s' \in \cS} \PP(s_{k+1}=ds'|s_k=ds, a_k=a)\\
		&=\frac{1}{2} \int_{s\in \cS}\sum_{a\in \cA}| \PP(s_k=ds, a_k=a) - \PP(\tilde{s}_k = ds, \tilde{a}_k=a) | \\
		&=\frac{1}{2} \int_{s \in \cS} \sum_{a\in \cA} |\PP(s_k=ds)\pi_{\theta_k}(a|ds) - \PP(\tilde{s}_k=ds)\pi_{\theta_{k-z}}(a|ds)| \\
		&\leq \frac{1}{2} \int_{s \in \cS} \sum_{a\in \cA} |\PP(s_k=ds)\pi_{\theta_k}(a|ds) - \PP(s_k=ds)\pi_{\theta_{k-z}}(a|ds)| \\
		&\quad + \frac{1}{2} \int_{s \in \cS} \sum_{a\in \cA} |\PP(s_k=ds)\pi_{\theta_{k-z}}(a|ds) - \PP(\tilde{s}_k=ds)\pi_{\theta_{k-z}}(a|ds)| \\
		&\leq \frac{1}{2} \int_{s \in \cS} |\cA|L_\pi\|\theta_k - \theta_{k-z}\| \PP(s_k=ds) \\
		&\quad + \frac{1}{2} \int_{s\in \cS} |\PP(s_k=ds) - \PP(\tilde{s}_k=ds)| \sum_{a\in\cA}\pi_{\theta_{k-z}}(a|ds) \\
		&= \frac{1}{2} |\cA|L_\pi \|\theta_k - \theta_{k-z}\| + d_{TV}(\PP(s_k \in \cdot), \PP(\tilde{s}_k \in \cdot)). \numberthis \label{eq:markov-1}
		\end{align*}
		\endgroup
		
		The second term can be bounded as 
		\begin{align*}
		&d_{TV}(\PP(s_k\in \cdot), \PP(\tilde{s}_k \in \cdot)) \\
		&= \frac{1}{2}\int_{s' \in \cS} | \PP(s_k=ds) - \PP(\tilde{s}_k = ds)| \\
		&= \frac{1}{2} \int_{s' \in \cS} | \sum_{a \in \cA} \int_{s \in \cS} \PP(s_{k-1}=ds, a_{k-1}=a, s_k=ds') - \PP(\tilde{s}_{k-1}=ds, \tilde{a}_{k-1}=a, \tilde{s}_k=ds') | \\
		&\leq \frac{1}{2} \int_{s' \in \cS}  \sum_{a \in \cA} \int_{s \in \cS} |\PP(s_{k-1}=ds, a_{k-1}=a, s_k=ds') - \PP(\tilde{s}_{k-1}=ds, \tilde{a}_{k-1}=a, \tilde{s}_k=ds') | \\
		&= d_{TV}(\PP(\xi_{k-1}\in \cdot), \PP(\tilde{\xi}_{k-1}\in \cdot)). \numberthis \label{eq:markov-2}
		\end{align*}
		
		Combined (\ref{eq:markov-1}) and (\ref{eq:markov-2}), we obtain
		\begin{align*}
		d_{TV}(\PP(\xi_k\in \cdot), \PP(\tilde{\xi}_k \in \cdot)) \leq d_{TV}(\PP(\xi_{k-1}\in \cdot), \PP(\tilde{\xi}_{k-1} \in \cdot)) + \frac{1}{2}| \cA | L_\pi |\theta_k - \theta_{k-z}\|.
		\end{align*}
		
		Sum over $z-1$ steps, we obtain
		\begin{align*}
		d_{TV}(\PP(\xi_k\in \cdot), \PP(\tilde{\xi}_k \in \cdot)) & \leq d_{TV}(\PP(\xi_{k-z}\in \cdot), \PP(\tilde{\xi}_{k-z} \in \cdot)) + \frac{1}{2} \sum_{m=0}^{z-1} |\cA|L_\pi\|\theta_{k-m} - \theta_{k-z}\| \\
		&= \frac{1}{2} \sum_{m=0}^{z-1}|\cA|L_\pi\|\theta_{k-m} - \theta_{k-z}\|.
		\end{align*}
	\end{proof}
\end{lemma}

Next, we present some mathematical facts that are useful in our analysis.
\begin{lemma}[\citep{chen2021multi}, Lemma F.3] \label{lemma:consensus-matrix}
	For a doubly stochastic matrix $W\in\RR^{N\times N}$ and the difference matrix $\normalfont{Q:=I-\frac{1}{N}\bone\bone^T}$, it holds that for any matrix $H \in \RR^{N \times N}$, $\|W^kH\| \leq \nu^k \|QH\|$, where $\nu$ is the second largest singular value of $W$.
\end{lemma}

\begin{lemma}[descent lemma in high dimension]\label{lemma:descent-lemma-multi}
	Consider the mapping $F: \RR^{n} \rightarrow \RR^{m}$. If there exists a positive constant $L$ such that 
	\begin{align*}
	\| \nabla F(x) - \nabla F(y) \|_F \leq L \|x-y\|, \ \forall x, y \in \text{dom}(F), \numberthis \label{eq:descent-lemma-smooth}
	\end{align*}
	then the following holds
	\begin{align*}
	\| F(y) - F(x) - \nabla F(x)(y-x) \| \leq \frac{L}{2}\sqrt{m}\|y-x\|^2.
	\end{align*}
	\begin{proof}
		
		Observe that (\ref{eq:descent-lemma-smooth}) directly implies the smoothness of each entry $F_i$: 
		\[\|\nabla F_i(x) - \nabla F_i(y) \| \leq \|\nabla F(x) - \nabla F(y) \|_F \leq L \|x-y\|.\]
		Define 
		\[z_i(x,y):=F_i(y) - F_i(x) - \nabla F_i(x)^T(y-x).\]
		We have
		\begin{align*}
		\| F(y) - F(x) - \nabla F(x) (y-x) \| 
		&= \sqrt{\sum_{i=1}^{m}z_i(x,y)^2} \\
		&\leq \sqrt{m (\frac{L}{2} \| y-x\|^2)^2} \\
		&= \frac{L_1}{2}\sqrt{m}\|y-x\|^2,
		\end{align*}
		where the inequality follows the descent lemma.
	\end{proof}
\end{lemma}

\begin{lemma}[Lipschitz property of multiplication]\label{lemma:lipschitz-multi}
	Suppose $f(x)$ and $g(x)$ are two functions bounded by $C_f$ and $C_g$, and are $L_f$- and $L_g$-Lipschitz continuous, then $f(x)g(x)$ is $C_fL_g+C_gL_f$-Lipschitz continuous.
	\begin{proof}
		\begin{align*}
		\|f(x_1)g(x_1) - f(x_2)g(x_2)\| &= \|f(x_1)g(x_1) - f(x_1)g(x_2) + f(x_1)g(x_2) - f(x_2)g(x_2)\| \\
		&\leq \|f(x_1)\| \|g(x_1) - g(x_2)\| + \|f(x_1)-f(x_2)\| \|g(x_2)\| \\
		&\leq (C_fL_g + C_gL_f) \| x_1 - x_2 \|.
		\end{align*}
	\end{proof}
\end{lemma}
\begin{lemma}[invertible property of matrix]\label{lemma:invertible-matrix}
	If a square matrix $A$ satisfies $\lim_{t \rightarrow \infty} A^t = 0$, or equivalently, $|\lambda(A)| < 1$, then $I-A$ is invertible.
	\begin{proof}
		\begin{align*}
		(I-A)\lim_{t\rightarrow \infty}\sum_{i=0}^{t} A^t &= \lim_{t \rightarrow \infty}[\sum_{i=0}^{t} A^t - \sum_{i=1}^{t+1}A^{t}] \\
		&= I - \lim_{t \rightarrow \infty} A^{t+1} \\
		&= I
		\end{align*}
		Since $I$ is invertible, by the rank inequality $\text{rank}(AB)\leq \min(\text{rank}(A), \text{rank}(B))$, 
		$I-A$ and $\lim_{t\rightarrow \infty}\sum_{i=0}^{t} A^t$ will be full rank and thereby invertible.
	\end{proof}
\end{lemma}

\section{Supporting lemmas}\label{appendix:supporting-lemma}
Before proceeding to the analysis of critic variables, we justify the uniqueness of fix point for critic and reward estimator variables under the update (\ref{eq:critic-update}) and (\ref{eq:reward-estimator-update}), respectively. Define the following notations
\begin{align*}
A_{\theta, \phi} &:= \EE[\phi(s)(\gamma\phi(s')^T-\phi(s)^T)], \numberthis \label{eq:def-Atheta}\\
A_{\theta, \varphi} &:= \EE[\varphi(s,a)\varphi(s,a)^T], \\
b_{\theta, \phi} &:= \EE[\phi(s)\bar{r}(s,a)], \\
b_{\theta, \varphi} &:= \EE[\varphi(s,a)\bar{r}(s,a)],
\end{align*}
with expectation taken from $s \sim \mu_\theta(s), a \sim \pit, s' \sim \cP$. 
The optimal critic and reward estimator variables given policy $\theta$ will satisfy $A_{\theta, \phi} \omega^*(\theta) + b_{\theta, \phi}  = 0; A_{\theta, \varphi} \lambda^*(\theta) + b_{\theta, \varphi} = 0.$ By Assumption \ref{assump:negative-definiteness}, $A_{\theta, \phi}$ and $A_{\theta, \varphi}$ are negative definite with largest eigenvalue $\lambda_\phi$ and $\lambda_{\varphi}$, which ensures the unique solution $\omega^*(\theta) = -A_{\theta, \phi}^{-1} b_{\theta, \phi}; \lambda^*(\theta) = -A_{\theta, \varphi}^{-1} b_{\theta, \varphi}.$ Let $R_\omega:=\frac{r_{\max}}{\lambda_{\phi}}, R_\lambda:=\frac{r_{\max}}{\lambda_{\varphi}}$. Then the norm of optimal solutions will be bounded as $\|\omega^*(\theta)\| \leq R_\omega, \ \|\lambda^*(\theta)\| \leq R_\lambda$, which justifies the projection step of the Algorithm~\ref{algorithm:dec-ac-re}. { In practice, the knowledge of $\lambda_{\phi}$ and $\lambda_{\varphi}$ may not be available. One can estimate projection radius online using the methods proposed in Section 8.2 of \citep{bhandari2018finite}}.
%\begin{align*}
%	A_{\theta, \phi} \omega^*(\theta) + b_{\theta, \phi}  = 0, \\
%	A_{\theta, \varphi} \lambda^*(\theta) + b_{\theta, \varphi} = 0.
%\end{align*}
%
%By Assumption \ref{assump:negative-definiteness}, $A_{\theta, \phi}$ and $A_{\theta, \varphi}$ are negative definite with largest eigenvalue $\lambda_\phi$ and $\lambda_{\varphi}$, which ensures the unique solution
%\begin{align*}
%	\omega^*(\theta) &= -A_{\theta, \phi}^{-1} b_{\theta, \phi}, \\
%	\lambda^*(\theta) &= -A_{\theta, \phi}^{-1} b_{\theta, \phi}.
%\end{align*}
%Let $R_\omega:=\frac{r_{\max}}{\lambda_{\phi}}, R_\lambda:=\frac{r_{\max}}{\lambda_{\varphi}}$. Then the norm of optimal solutions will be bounded as $\|\omega^*(\theta)\| \leq R_\omega, \ \|\lambda^*(\theta)\| \leq R_\lambda$, which justifies the projection step of the Algorithm~\ref{algorithm:dec-ac-re}.

We slightly abuse the notation by overwriting $V_{\pi_\theta}$ as $V_\theta$. To study the error of critic, we introduce the following notations (crf. $\xi := (s,a,s')$)
\begin{align*}
\delta^i(\xi, \theta) &:= r^i(s,a) + \gamma V_\theta(s') - V_\theta(s) \\
\delta(\xi, \theta) &:= \bar{r}(s,a) + \gamma V_\theta(s') - V_\theta(s)\\
\tilde{\delta}(\xi, \omega) &:= \bar{r}(s,a) + \gamma \phi(s')^T\omega - \phi(s)^T\omega \\
\hat{\delta}(\xi, \omega, \lambda) &:= \varphi(s,a)^T \lambda + \gamma \phi(s')^T\omega - \phi(s)^T\omega,\numberthis \label{eq:definition-delta}
\end{align*}

%Then the update of $\theta^i, \omega^i$, and $\bar{\omega}$ can be described by the following expressions, respectively
For the ease of expression, we further define
\begin{align*}
g_a^i(\xi, \omega, \lambda)&:= \hat{\delta}(\xi, \omega, \lambda) \psi_{\theta^i}(s,a^i), \\
g_c^i(\xi, \omega) &:= \delta^i(\xi, \omega) \phi(s), \\
\bar{g}_c(\xi, \omega) &:= \tilde{\delta}(\xi, \omega) \phi(s), \\
g_c(\theta, \omega) &:= \EE_{\xi \sim \mu_\theta}[\bar{g}_c(\xi, \omega)]. \numberthis \label{eq:definition-step}
\end{align*}

% We will start with the error of averaged critic parameter first. The following lemma characterizes the descent of averaged critic variables under i.i.d. sampling.

\subsection{Error of critic}\label{appendix:critic-error}

The following lemmas and propositions serves as the preliminary for establishing the approximate descent property of the critic variables' optimal gap.

\begin{proposition}[Lipschitz continuity of $\omega^*(\theta)$~\citep{wu2020finite}] \label{proposition:lipschitz-continuous-omega}
	Suppose Assumptions \ref{assump:boundedness}, \ref{assump:negative-definiteness}, \ref{assump:lipschitz-policy}, and \ref{assump:markov-chain} hold, then there exists a positive constant $L_\omega$ such that for any $\theta_1, \theta_2 \in \RR^{Nd_\theta}$, we have 
	\begin{align*}
	\| \omega^*(\theta_1) - \omega^*(\theta_2) \| \leq L_\omega\|\theta_1 - \theta_2\|.
	\end{align*}
\end{proposition}

\begin{lemma}[smoothness of stationary distribution]
	For any $\theta, \theta' \in \RR^{d}$, there exists a positive constant $L_{\mu, 2}$ such that $\|\nabla \mu_{\theta}(s) - \nabla \mu_{\theta'}(s)\| \leq L_{\mu, 2} \|\theta - \theta'\|$. \label{lemma:smooth-stationary}
\end{lemma}

The proof of this Lemma consists of two main steps: 1) Derive the expression of the gradient  and 2) establish that the gradient is Lipschitz continuous. For the first part, we follow the main idea in \citep{baxter2001infinite}. 

\begin{proof}
	For a given policy $\pi_{\theta}$, we define the transition probability $P_{\theta}(s|s'):=\sum_{a}\pi_{\theta}(a|s') P(s|s', a)$. By the Assumption \ref{assump:markov-chain}, there exists a stationary distribution $\mu_{\theta}(s)$ which satisfies for all state $s$
	
	\begin{align*}
	\mu_{\theta}(s) = \sum_{s' \in \cS}	\mu_{\theta}(s') P_{\theta}(s|s') \numberthis \label{eq:lip-statdis-1}
	\end{align*}
	
	Define the following notations 
	\begin{align*}
	\mu_{\theta} &:= [\mu_{\theta}(s_1), \mu_{\theta}(s_2), \cdots, \mu_{\theta}(s_n)]^T &\RR^{|\cS| \times 1} \\
	P_{\theta}(s) &:= [P_{\theta}(s|s_1), P_{\theta}(s|s_2), \cdots, P_{\theta}(s|s_n)]^T &\RR^{|\cS| \times 1} \\
	P(\theta) &:= [P_{\theta}(s_1), P_{\theta}(s_2), \cdots, P_{\theta}(s_n)] & \RR^{|\cS|\times|\cS|} \\
	\nabla \mu_{\theta} &:= [\nabla \mu_{\theta}(s_1), \nabla \mu_{\theta}(s_2), \cdots, \nabla \mu_{\theta}(s_n)] & \RR^{d_\theta \times |\cS|} \\
	\nabla P_{\theta}(s) &:= [\nabla P_{\theta}(s|s_1), \nabla P_{\theta}(s|s_2), \cdots, \nabla P_{\theta}(s|s_n)] & \RR^{d_\theta \times |\cS|}\\
	\end{align*}
	
	Upon taking derivative with respect to $\theta$ on both sides of (\ref{eq:lip-statdis-1}), we have
	\begin{align*}
	\nabla\mu_{\theta}(s) &= \sum_{s'\in \cS} \nabla \mu_{\theta}(s') P_{\theta}(s|s') + \mu_{\theta}(s')\nabla_{\theta}P_{\theta}(s|s') \\
	&= \nabla \mu_{\theta} P_{\theta}(s) + \nabla P_{\theta}(s) \mu_{\theta} \numberthis \label{eq:lip-statdis-2}
	\end{align*}
	
	(\ref{eq:lip-statdis-2}) can be written in compact form as
	\begin{align*}
	\nabla \mu_{\theta} &= \nabla \mu_{\theta} P(\theta) + [\nabla P_{\theta}(s_1)\mu_{\theta}, \cdots, \nabla P_{\theta}(s_n)\mu_{\theta}] \numberthis \label{eq:lip-statdis-3}
	\end{align*}
	
	Therefore, we have
	\begin{align*}
	[\nabla P_{\theta}(s_1)\mu_{\theta}, \cdots, \nabla P_{\theta}(s_n)\mu_{\theta}] &= \nabla \mu_{\theta} (I-P(\theta)) \\
	&= \nabla \mu_{\theta} (I-(P(\theta)-e\mu_{\theta}^T)),
	\end{align*}
	where the second inequality is due to $\nabla \mu_{\theta} e = \nabla (\mu_{\theta}e) = \nabla 1 = 0$. 
	
	We now show that $I-(P(\theta) - e\mu_{\theta}^T)$ is invertible. The first step is to show $\lim_{t \rightarrow \infty} (P(\theta) - e\mu_{\theta}^T)^t = 0$. Let $P, \mu$ represent $P(\theta), \mu_{\theta}$ for simplicity, we first show $(P - e\mu^T)^t = P^t - P^{t-1}e\mu^T$ by induction. Observe that when $t=1$, this is trivially satisfied. Suppose the equality holds for $t=k$, then 
	\begin{align*}
	(P-e\mu^T)^{k+1} &= (P^{k} - P^{k-1}e\mu^T) P - (P^{k} - P^{k-1}e\mu^T) e\mu^T \\
	&= P^{k+1} - P^{k-1}e\mu^T - P^{k}e\mu^T + P^{k-1}(e\mu^{T})^2 \\
	&= P^{k+1} - P^{k}e\mu^T,
	\end{align*}
	where the second equality is due to (\ref{eq:lip-statdis-1}) such that $e\mu^TP=e\mu^T$ and 
	the last equality is due to $\mu^Te = 1$.
	
	% Following the property of the stationary distribution, we have $\lim_{t\rightarrow\infty}P(\theta)^t = e\mu(\theta)^T$.
	Therefore, we have 
	\[\lim_{t \rightarrow \infty} (P(\theta) - e\mu_{\theta}^T)^t = \lim_{t\rightarrow \infty}(P(\theta)^{t} - P(\theta)^{t-1}e\mu_{\theta}^T) = e\mu_{\theta}^T - e\mu_{\theta}^T = 0, \]
	which together with Lemma~\ref{lemma:invertible-matrix} justifies that $I-(P(\theta) - e\mu_{\theta}^T)$ is invertible.
	Thus, we have 
	\begin{align*}
	\nabla \mu_{\theta} = (I-(P(\theta) - e\mu_{\theta}^T))^{-1} [\nabla P_{\theta}(s_1)\mu_{\theta}, \cdots, \nabla P_{\theta}(s_n)\mu_{\theta}]. \numberthis \label{eq:lip-statdis-4}
	\end{align*}
	We will utilize Lemma~\ref{lemma:lipschitz-multi} to prove the Lipschitz property of $\nabla \mu_{\theta}$. We first show the Lipschitz continuous of the first term. Let $A(\theta)$ to represent $I-(P(\theta) - e\mu_{\theta}^T)$, then we have
	\begin{align*}
	\|A(\theta_1) - A(\theta_2)\| &= \|P(\theta_1) - P(\theta_2) + e(\mu_{\theta_2} - \mu_{\theta_1})^T\| \\
	&\leq \|P(\theta_1) - P(\theta_2)\| + \|e(\mu_{\theta_2} - \mu_{\theta_1})^T\| \\
	&= \sqrt{\sum_{s,s'\in\cS}|\sum_{a\in\cA}(\pi_{\theta_1}(a|s') - \pi_{\theta_2}(a|s'))P(s|s',a)|^2} + \sqrt{|\cS|} \|\mu_{\theta_2} - \mu_{\theta_1}\| \\
	&\leq \sqrt{\sum_{s, s'\in \cS}(\sum_{a\in\cA}|(\pi_{\theta_1}(a|s') - \pi_{\theta_2}(a|s'))P(s|s',a)|)^2} + \sqrt{|\cS|} \|\mu_{\theta_2} - \mu_{\theta_1}\| \\
	% 	&\leq \sqrt{\sum_{s'\in\cS}|\cA|^2L_\pi^2\|\theta_1-\theta_2\|^2\sum_{s\in\cS}P(s|s',a)} + \sqrt{|\cS|}L_{\mu}\|\theta_1 - \theta_2\| \\
	&\leq \sqrt{\sum_{s'\in \cS}|\cA|^2L_{\pi}^2\|\theta_1 - \theta_2\|^2 \sum_{s\in\cS}P(s|s',a)^2} + \sqrt{|\cS|}L_{\mu}\|\theta_1 - \theta_2\| \\
	&= \sqrt{|\cS|}(|\cA|L_\pi+L_\mu)\|\theta_1-\theta_2\|.
	\end{align*}
	where the second inequality uses triangle inequality. The last inequality is due to Lipschitz continuous of the policy specified in Assumption~\ref{assump:lipschitz-policy}, and Lipschitz continuous of $\mu_{\theta}$ implied by Lemma~\ref{lemma:lipschitz-continuous-value}. 
	
	To see that $A^{-1}(\theta)$ is Lipschitz continuous and bounded, observe that
	\begin{align*}
	\|A^{-1}(\theta_1) - A^{-1}(\theta_2)\| &= \|A^{-1}(\theta_2) (A(\theta_2) - A(\theta_1) ) A^{-1}(\theta_1)\| \\
	&\leq \|A^{-1}(\theta_2) \| \|A^{-1}(\theta_1)\| \|A(\theta_2) - A(\theta_1)\| \\
	&\leq \sqrt{|\cS|}({|\cA|}L_\pi+L_\mu) \|A^{-1}(\theta_2)\| \|A^{-1}(\theta_1)\|\|\theta_2 - \theta_1\|, \numberthis \label{eq:lip-statdis-5}
	\end{align*}
	where the first inequality uses Cauchy-Schwartz inequality, and the last inequality uses the Lipschitz continuous of $A(\theta)$ in (\ref{eq:lip-statdis-5}). Since $\|A(\theta)\|$ is bounded, $\|A^{-1}(\theta)\|$ is also bounded (due to invertibility), which justifies that the first term in (\ref{eq:lip-statdis-4}) is Lipschitz continuous and bounded. 
	
	We now consider the second term in (\ref{eq:lip-statdis-4}). For any state $s$
	\begin{align*}
	\|\nabla P_{\theta_1}(s)\mu_{\theta_1} - \nabla P_{\theta_2}(s)\mu_{\theta_2} \| &= \|\nabla P_{\theta_1}(s)(\mu_{\theta_1}- \mu_{\theta_2}) + (\nabla P_{\theta_1}(s)-\nabla P_{\theta_2}(s))\mu_{\theta_2}\| \\
	&\leq \|\nabla P_{\theta_1}(s)(\mu_{\theta_1}- \mu_{\theta_2})\| + \| (\nabla P_{\theta_1}(s)-\nabla P_{\theta_2}(s))\mu_{\theta_2} \| \\
	&\leq \|\nabla P_{\theta_1}(s)\| \|\mu_{\theta_1}-\mu_{\theta_2}\| + \|\nabla P_{\theta_1}(s)-\nabla P_{\theta_2}(s)\| \|\mu_{\theta_2}\| \\
	&\leq \sum_{s' \in \cS} \sum_{a \in \cA} \|\nabla \pi_{\theta_1}(a|s') P(s|s',a) \| L_{\mu}\|\theta_1-\theta_2\| \\
	&\quad + \sum_{s'\in\cS}\sum_{a\in\cA} \|(\nabla \pi_{\theta_1}(a|s') - \nabla \pi_{\theta_2}(a|s')) P(s|s',a)\| \\ %\|\sum_{s'\in\cS}\sum_{a\in\cA}(\pi_{\theta_1}(a|s')-\pi_{\theta_2}(a|s')P(s|s',a))\| \\
	% 	&\leq \sum_{s' \in \cS} \sum_{a \in \cA} L_{\mu} \| \nabla \pi_{\theta_1}(a|s') P(s|s',a)\| \|\theta_1-\theta_2\| + \sqrt{|\cS|}|\cA| L_{\pi} \|\theta_1 - \theta_2\| \\
	%\sum_{a\in\cA} \| \pi_{\theta_1}(a|s')-\pi_{\theta_2}(a|s') \| \\
	&\leq |\cS||\cA|(C_\pi L_\mu+L_\pi) \|\theta_1 - \theta_2\|,
	\end{align*}
	which justifies the Lipschitz continuous of $\nabla P_{\theta}(s)\mu_{\theta}$. Define $B(\theta):= [\nabla P_{\theta}(s_1)\mu_{\theta}, \cdots, \nabla P_{\theta}(s_n)\mu_{\theta}]$, we have 
	\begin{align*}
	\| B(\theta_1) - B(\theta_2) \| \leq |\cS|^{3/2}|\cA|(C_\pi L_\mu + L_\pi) \|\theta_1 - \theta_2\|.
	\end{align*}
	Since $\nabla \mu_{\theta} = A^{-1}(\theta)B(\theta)$, with $A^{-1}(\theta)$ and $B(\theta)$ being Lipschitz continuous and bounded. Therefore, according to Lemma~\ref{lemma:lipschitz-multi}, there exists a positive constant $L_{\mu, 2}$ which satisfies
	\begin{align*}
	\|\nabla \mu_{\theta_1} - \nabla \mu_{\theta_2}\| \leq L_{\mu, 2} \|\theta_1 - \theta_2\|. 
	\end{align*}
\end{proof}

\begin{proposition}[restatement of Lemma~\ref{lemma:opt-critic-smoothness}, Lipschitz continuity of $\nabla_\theta \omega^*(\theta)$~\citep{chen2021closing}] \label{lemma:lipschitz-smooth-omega}
	Suppose Assumptions \ref{assump:boundedness}-\ref{assump:markov-chain} holds, then there exists a positive constant $L_{\omega, 2}$ such that 
	\begin{align*}
	\|\nabla_\theta \omega^*(\theta_1) - \nabla_\theta \omega^*(\theta_2)\|_F \leq L_{\omega, 2}\|\theta_1 - \theta_2\|.
	\end{align*}
\end{proposition}
\begin{proof}
	The proof follows the derivation of Proposition 8 of \citep{chen2021closing}. However, they make assumption that $\mu_{\theta}(s)$ is Lipschitz continuous, which we have justified in Lemma~\ref{lemma:smooth-stationary}. We present the proof for the completeness.
	
	We have $\omega^*(\theta) = -A_{\theta,\phi}^{-1}b_{\theta,\phi}$, where $A_{\theta, \phi}$ is defined in (\ref{eq:def-Atheta}). The Jacobian of $\omega^*(\theta)$ can be calculated as 
	\begin{align*}
	\nabla_{\theta}\omega^*(\theta) &= -\nabla_{\theta}(A_{\theta,\phi}^{-1}b_{\theta, \phi})\\
	&= -A_{\theta, \phi}^{-1}(\nabla_{\theta}A_{\theta, \phi})A_{\theta, \phi}^{-1}b_{\theta, \phi} - A_{\theta, \phi}(\nabla_{\theta}b_{\theta, \phi}). \numberthis\label{eq:smooth-stationary-1}
	\end{align*}
	We can utilize Lemma~\ref{lemma:lipschitz-multi} to show the Lipschitz continuity of $\nabla \omega^*(\theta)$. We have to verify the Lipschitz continuity and boundedness of $A_{\theta,\phi}^{-1}, b_{\theta, \phi}, \nabla_{\theta}A_{\theta, \phi}$, and $\nabla_{\theta}b_{\theta,\phi}$.
	
	The Lipschitz continuity and boundedness of $A_{\theta,\phi}^{-1}$ has been shown in (\ref{eq:lip-statdis-5}. 
	Let $b_1$ and $b_2$ represent $b_{\theta_1,\phi}$, $b_{\theta_2,\phi}$, we have
	\begin{align*}
	\|b_1-b_2\| &= \|\EE[\bar{r}(s,a,s')\phi(s)] - \EE[r(\tilde s, \tilde a, \tilde s')\phi(\tilde s)]\| \\
	&\leq \sup_{s,a,s'}\|r(s,a,s')\phi(s)\| \|\PP((s,a,s' \in \cdot)) - \PP((\tilde s,\tilde a, \tilde s' \in \cdot))\|_{TV} \\
	&\leq r_{\max}\|\PP((s,a,s' \in \cdot)) - \PP((\tilde s,\tilde a, \tilde s' \in \cdot))\|_{TV} \\
	&\leq 2 |\cA|L_{\pi}(1+\log_{\rho}\kappa^{-1} + (1-\rho)^{-1} \|\theta_1-\theta_2\|,
	\end{align*}
	where the last inequality follows Lemma~\ref{lemma:error-total-variation}.
	
	We now analyze $\nabla_{\theta} A_{\theta, \phi}$. We first define
	\begin{align*}
	A(s,s'):=\phi(s)(\gamma\phi(s')-\phi(s))^T, \quad b(s,a,s'):=r(s,a,s')\phi(s).
	\end{align*}
	as
	\begin{align*}
	\nabla_{\theta}A_{\theta, \phi} &= \nabla_{\theta}\left(\sum_{s,a,s'}\mu_{\theta}(s) \pi_{\theta}(a|s) P(s'|s,a)A(s,s')\right) \\
	&=\sum_{s,a,s'} \left[\nabla_{\theta}\mu_{\theta}(s)\pi_{\theta}(a|s)P(s'|s,a)A(s,s') + \mu_{\theta}\nabla_{\theta}\pi_{\theta}(a|s)P(s'|s,a)A(s,s') \right].
	\end{align*}
	By Lemma~\ref{lemma:smooth-stationary} and Lemma~\ref{lemma:error-total-variation}, and Assumption~\ref{assump:lipschitz-policy}, $\mu_{\theta}(s),$ $\pi_{\theta}(a|s), \nabla_{\theta}\mu_{\theta}(s)$, $\nabla_{\theta}\pit(a|s)$ are Lipschitz continuous and bounded. Therefore, $\nabla_{\theta}A_{\theta,\phi}$ is Lipschitz and bounded. 
	
	Finally, we analyze $\nabla_{\theta}b_{\theta, \phi}$ by following the same technique.
	\begin{align*}
	\nabla_{\theta} b_{\theta, \phi} &= \nabla_{\theta} \left(\sum_{s,a,s'}\mu_{\theta}(s)\pi_{\theta}(a|s)P(s'|s,a)b(s,a,s')\right) \\
	&= \sum_{s,a,s'}\left[\nabla_{\theta}\mu_{\theta}(s)\pit(a|s) P(s'|s,a) b(s,a,s') + \mu_{\theta}(s)\nabla_{\theta}\pit(a|s)P(s'|s,a)b(s,a,s') \right].
	\end{align*}
	By Lemma~\ref{lemma:smooth-stationary} and Lemma~\ref{lemma:error-total-variation}, and Assumption~\ref{assump:lipschitz-policy}, $\mu_{\theta}(s),$ $\pi_{\theta}(a|s), \nabla_{\theta}\mu_{\theta}(s)$, $\nabla_{\theta}\pit(a|s)$ are Lipschitz continuous and bounded. Thus, $\nabla_{\theta}b_{\theta,\phi}$ is bounded and Lipschitz continuous. 
	
	We have shown the Lipschitz continuity and boundedness of $A_{\theta, \phi}^{-1}$, $b_{\theta,\phi}$, $\nabla_{\theta}A_{\theta,\phi}$, and $\nabla_{\theta}b_{\theta,\phi}$. Therefore, by applying Lemma~\ref{lemma:lipschitz-multi}, we conclude that there exists a positive constant $L_{\omega, 2}$ such that $\nabla_{\theta}\omega^*(\theta)$ in (\ref{eq:smooth-stationary-1}) is $L_{\omega, 2}$-Lipschitz continuous.
\end{proof}

\begin{lemma}[descent of critic's optimal gap (Markovian sampling)]\label{lemma:critic-error-markov-1}
Under Assumptions \ref{assump:boundedness}-\ref{assump:doub_stoch}, with $\omega_{k+1}$ generated by Algorithm~\ref{algorithm:dec-ac-re} given $\omega_{k}$ and $\theta_k$ under Markovian sampling, then the following holds
\begin{align*}
\EE[\|\bar{\omega}_{k+1} - \omega^*(\theta_{k+1})\|^2|\theta_k] &\leq \left(1 + 4L_{\omega}N \alpha_k + \frac{L_{\omega, 2}^2}{2} C_\theta^2N\sqrt{d_\theta}\alpha_k^2\right)\EE\|\bar{\omega}_{k+1} - \omega^*(\theta_k)\|^2 \\
&\quad +\left(\frac{L_{\omega, 2}^2 }{2} C_{\theta}^2 N + L_\omega^2C_\theta^2N\right) \alpha_k^2 + \frac{\alpha_k}{4}\sum_{i=1}^{N} \| \EE[g_a^i(\xi_k, \omega_{k+1}^{i}, \lambda_{k+1}^{i})] \|^2. \numberthis \label{eq:critic-error-7}
\end{align*}
\begin{align*}
\EE [\| \bar{\omega}_{k+1} - \omega^*(\theta_k) \|^2|\theta_k] \leq (1-2\lambda_{\phi} \beta_k) \EE \| \bar{\omega}_k - \omega^*(\theta_k) \|^2 + C_{K_1} \beta_k\beta_{k-Z_K} + C_{K_2}\alpha_{k-Z_K} \beta_k. \numberthis \label{eq:critic-error-8}
\end{align*}
where the constants are defined as
$C_{K_1}:= 4C_2C_\delta Z_K + C_\delta^2,\ C_{K_2}:= 
4C_1C_{\theta}Z_K + 2C_3C_{\theta}Z_K^2 + C_8, Z_K:= \min\{z \in \NN^+ | \kappa\rho^{z-1} \leq \min \{\alpha_K, \beta_K, \eta_K\}\}$.
\begin{proof}

We begin with the optimality gap of averaged critic variables
\begin{align*}
&\| \bar{\omega}_{k+1} - \omega^*(\theta_{k+1}) \|^2 \\
&= \| \bar{\omega}_{k+1} - \omega^*(\theta_k) + \omega^*(\theta_k) - \omega^*(\theta_{k+1}) \|^2 \\
&= \| \bar{\omega}_{k+1} - \omega^*(\theta_k)\|^2 + \|\omega^*(\theta_k) - \omega^*(\theta_{k+1})\|^2 + 2 \langle \bar{\omega}_{k+1} - \omega^*(\theta_k), \omega^*(\theta_k) - \omega^*(\theta_{k+1}) \rangle \\
%	&\leq \bar{\omega}_{k+1} - \omega^*(\theta_k)\|^2 + L_\omega^2 \|\theta_k - \theta_{k+1} \|^2 + 2 \langle \bar{\omega}_{k+1} - \omega^*(\theta_k), \omega^*(\theta_k) - \omega^*(\theta_{k+1}) \rangle \\
&\leq \| \bar{\omega}_{k+1} - \omega^*(\theta_k)\|^2 + NL_\omega^2 C_\theta^2 \alpha_k^2 + 2 \langle \bar{\omega}_{k+1} - \omega^*(\theta_k), \nabla\omega^*(\theta_k)^T(\theta_k - \theta_{k+1}) \rangle \\
&\quad + 2\langle \bar{\omega}_{k+1} - \omega^*(\theta_k) , \omega^*(\theta_k) - \omega^*(\theta_{k+1}) - \nabla\omega^*(\theta_k)^T(\theta_k - \theta_{k+1}) \rangle \numberthis \label{eq:critic-error-3},
\end{align*}
where the inequality is based on the Lipschitz of $\omega^*(\theta)$ implied by Proposition~\ref{proposition:lipschitz-continuous-omega}
\begin{align*}
\|\omega^*(\theta_k) - \omega^*(\theta_{k+1})\|^2 &\leq L_\omega^2\|\theta_{k} - \theta_{k+1}\|^2, \\
\|\theta_k - \theta_{k+1}\|^2 &=  \sum_{i=1}^{N}\|\alpha_k\hat{\delta}(\xi_k, \omega_k^i, \lambda_k^i)\psi_{\theta_k^i}(s_k, a_k^i)\|^2 \leq N \alpha_k^2 C_\theta^2, \numberthis \label{eq:bound-theta-update}
\end{align*}
with $C_\theta:=C_\delta C_\psi$, and $C_\delta$ is defined in (\ref{eq:def-c-delta}).

The third term in (\ref{eq:critic-error-3}) can be bounded as 
\begin{align*}
&\langle \bar{\omega}_{k+1} - \omega^*(\theta_k), \nabla\omega^*(\theta_k)^T(\theta_k - \theta_{k+1}) \rangle \\
&\leq \|\bar{\omega}_{k+1} - \omega^*(\theta_k)\| \|\nabla\omega^*(\theta_k)^T(\theta_k - \theta_{k+1})\| \\
&\leq \|\nabla \omega*(\theta_k)\| \|\bar{\omega}_{k+1} - \omega^*(\theta_k)\| \|\theta_k - \theta_{k+1}\| \\
&\leq L_{\omega} \|\bar{\omega}_{k+1} - \omega^*(\theta_k)\| \|\theta_k - \theta_{k+1}\| \\
&\leq \sum_{i=1}^{N} L_{\omega}\alpha_k \|\bar{\omega}_{k+1} - \omega^*(\theta_k)\| \|g_a^i(\xi_k, \omega_{k+1}^{i}, \lambda_{k+1}^{i})\| \\
&\leq \sum_{i=1}^{N} (2L_{\omega}\alpha_k\|\bar{\omega}_{k+1} - \omega^*(\theta_k)\|^2 + \frac{\alpha_k}{8}\|g_a^i(\xi_k, \omega_{k+1}^{i}, \lambda_{k+1}^{i})\|^2), \numberthis \label{eq:critic-error-tmp}
\end{align*}
where the second inequality follows Proposition~\ref{proposition:lipschitz-continuous-omega}, the third inequality uses triangle inequality, and the last inequality uses Young's inequality.

The last term in (\ref{eq:critic-error-3}) can be bounded as 
\begin{align*}
& \EE\langle \bar{\omega}_{k+1} - \omega^*(\theta_k) , \omega^*(\theta_k) - \omega^*(\theta_{k+1}) - \nabla\omega^*(\theta_k)^T(\theta_k - \theta_{k+1}) \rangle \\
&\leq \frac{L_{\omega, 2}^2}{2}\sqrt{d_{\theta}}\EE \| \bar{\omega}_{k+1} - \omega^*(\theta_k)\| \| \theta_{k+1}-\theta_k\|^2 \\
&\leq \frac{L_{\omega, 2}^2}{4}\sqrt{d_{\theta}} \EE\|\bar{\omega}_{k+1} - \omega^*(\theta_k)\|^2 \|\theta_{k+1} - \theta_k\|^2 + \frac{L_{\omega, 2}^2}{4} \| \theta_{k+1}- \theta_k\|^2 \\
&\leq \frac{L_{\omega, 2}^2}{4}\sqrt{d_{\theta}} NC_\theta^2\alpha_k^2 \EE\|\bar{\omega}_{k+1} - \omega^*(\theta_k)\|^2 + \frac{L_{\omega, 2}^2 }{4}N C_{\theta}^2 \alpha_k^2. \numberthis \label{eq:critic-error-4}
\end{align*}
The first inequality uses Lemma~\ref{lemma:descent-lemma-multi}. The second inequality is induced by Young's inequality. The last inequality follows (\ref{eq:bound-theta-update}).
	
Plug (\ref{eq:critic-error-tmp}) and (\ref{eq:critic-error-4}) into (\ref{eq:critic-error-3}) will yield (\ref{eq:critic-error-7}).

% (\ref{eq:critic-error-7}) has already been derived in the proof of i.i.d. sampling setting, please check the derivation of (\ref{eq:critic-error-1}).

We now prove (\ref{eq:critic-error-8}). By the critic update rule, we have (crf. $g_c(\theta, \omega):= \EE_{\xi \sim \mu_{\theta}}[\bar{g}_c(\xi, \omega)]$.)
\begin{align*}
	\EE \| \bar{\omega}_{k+1} - \omega^*(\theta_k)\|^2 &= 
	\EE \|\Pi_{R_\omega} (\bar{\omega}_k + \beta_k \bar{g}_c(\xi_k, \bar{\omega}_k)) - \Pi_{R_\omega} \omega^*(\theta_k) \|^2 \\
	& \overset{(i)}{\leq} \EE \| \bar{\omega}_k + \beta_k \bar{g}_c(\xi, \bar{\omega}_k) - \omega^*(\theta_k) \|^2 \\
	& = \| \bar{\omega}_k - \omega^*(\theta_k) \|^2 + \beta_k^2 \EE\| \bar{g}_c(\xi_k, \bar{\omega}_k)\|^2 + 2 \beta_k \EE[ \langle \bar{\omega}_k - \omega^*(\theta_k), \EE[\bar{g}_c(\xi_k, \bar{\omega}_k)] \rangle] \\
	& \overset{(ii)}{\leq} \| \bar{\omega}_k - \omega^*(\theta_k) \|^2 + \beta_k^2 C_\delta^2 + 2 \beta_k  \langle \bar{\omega}_k - \omega^*(\theta_k), \EE[\bar{g}_c(\xi_k, \bar{\omega}_k)] \rangle \\
	& = \| \bar{\omega}_k - \omega^*(\theta_k) \|^2 + \beta_k^2 C_\delta^2 + 2 \beta_k \langle \bar{\omega}_k - \omega^*(\theta_k), g_c(\theta_k, \bar{\omega}_k) \rangle \\
    &\quad + 2 \beta_k \EE\langle \bar{\omega}_k - \omega^*(\theta_k) , \EE[\bar{g}_c(\xi_k, \bar{\omega}_k)] - g_c(\theta_k, \bar{\omega}_k) \rangle, 
	\numberthis \label{eq:critic-error-5}
	\end{align*}
	where (i) is due to the non-expansiveness of projection to convex set, and (ii) follows 
	\[\|\bar{g}_c(\xi, \omega)\| \leq |r(s,a) + \gamma\phi(s')^T\omega - \phi(s)^T\omega| \leq r_{\max}+(1+\gamma)R_\omega:=C_\delta. \numberthis \label{eq:def-c-delta}\]

The product in the third term in (\ref{eq:critic-error-5}) can be bounded as 
	\begin{align*}
	\langle \bar{\omega}_k - \omega^*(\theta_k), g_c(\theta_k, \bar{\omega}_k) \rangle &= 
	\langle \bar{\omega}_k - \omega^*(\theta_k), \EE_{\xi \sim \mu_{\theta_k}}[\bar{g}_c(\xi_k, \bar{\omega}_k)] \rangle\\
	&= \langle \bar{\omega}_k - \omega^*(\theta_k) , \EE_{\xi \sim \mu_{\theta_k}} [\bar{g}_c(\xi_k, \bar{\omega}_k) - g_c(\theta_k, \omega^*(\theta_k))] \rangle \\
	&= \beta_k \langle \bar{\omega}_k - \omega^*(\theta_k), \EE_{\xi \sim \mu_{\theta_k}}[\phi(s)(\gamma\phi(s') - \phi(s))^T|\theta_k] (\bar{\omega}_k - \omega^*(\theta_k)) \rangle \\
	&= \beta_k \langle \bar{\omega}_k - \omega^*(\theta_k), A_{\theta_k, \phi}(\bar{\omega}_k - \omega^*(\theta_k))  \rangle \\
	&\leq -\lambda_{\phi} \beta_k \| \bar{\omega}_k - \omega^*(\theta_k)\|^2. \numberthis \label{eq:critic-error-6}
	\end{align*}
	Here the first equality is due to critic's optimality condition $g_c(\theta_k, \omega^*(\theta_k)) = \EE_{\xi_k \sim \mu_{\theta_k}}[\bar{g}_c(\xi_k, \omega^*(\theta_k))|\theta_k] = 0$. The last inequality uses the negative definiteness of $A_{\theta_k, \phi}$ of Assumption~\ref{assump:negative-definiteness}.

Plug (\ref{eq:critic-error-6}) to (\ref{eq:critic-error-5}) gives 
\begin{align*}
    \EE \| \bar{\omega}_{k+1} - \omega^*(\theta_k)\|^2 &\leq (1-2\lambda_\phi\beta_k) \| \bar{\omega}_k - \omega^*(\theta_k) \|^2 + \beta_k^2 C_\delta^2 \\
    &\quad + 2 \beta_k \langle \bar{\omega}_k - \omega^*(\theta_k) , \EE[\bar{g}_c(\xi_k, \bar{\omega}_k)] - g_c(\theta_k, \bar{\omega}_k) \rangle. \numberthis \label{eq:critic-error-9} 
\end{align*}

% Follow the derivation of
% (\ref{eq:critic-error-5}), we have
% \begin{align*}
% \EE \| \bar{\omega}_{k+1} - \omega^*(\theta_k)\|^2 &\leq \| \bar{\omega}_k - \omega^*(\theta_k) \|^2 + \beta_k^2 C_\delta^2 + 2 \beta_k \EE[ \langle \bar{\omega}_k - \omega^*(\theta_k), \bar{g}_c(\xi_k, \bar{\omega}_k) \rangle]\\
% & = \| \bar{\omega}_k - \omega^*(\theta_k) \|^2 + \beta_k^2 C_\delta^2 + 2 \beta_k \EE\langle \bar{\omega}_k - \omega^*(\theta_k), g_c(\theta_k, \bar{\omega}_k) \rangle \\
% &\quad + 2 \beta_k \EE\langle \bar{\omega}_k - \omega^*(\theta_k) , \bar{g}_c(\xi_k, \bar{\omega}_k) - g_c(\theta_k, \bar{\omega}_k) \rangle \\
% &\leq (1-2\lambda_\phi\beta_k) \| \bar{\omega}_k - \omega^*(\theta_k) \|^2 + \beta_k^2 C_\delta^2 \\
% &\quad + 2 \beta_k \EE\langle \bar{\omega}_k - \omega^*(\theta_k) , \bar{g}_c(\xi_k, \bar{\omega}_k) - g_c(\theta_k, \bar{\omega}_k) \rangle. \numberthis \label{eq:critic-error-9}
% \end{align*}
% Here, the last inequality bound the third term using the same technique of (\ref{eq:critic-error-6}).

We now bound the last term in (\ref{eq:critic-error-9}). By Lemma~\ref{lemma:critic-markov-error}, for any  $z \in \NN^{+}$, we have
\begin{align*}
&\EE \langle \bar{\omega}_k - \omega^*(\theta_k) , \bar{g}_c (\xi_k, \bar{\omega}_k) - g_c(\theta_k, \bar{\omega}_k) \rangle \\
& \leq C_1 \EE\|\theta_k - \theta_{k-z}\| + C_2 \EE\| \bar{\omega}_k - \bar{\omega}_{k-z} \| + C_3 \sum_{m=0}^{z-1} \EE\|\theta_{k-m} - \theta_{k-z}\| + C_8 \kappa \rho^{z-1} \\
& \overset{(i)}{\leq} C_1 \sum_{n=1}^{z} \EE \| \theta_{k-n+1} - \theta_{k-n} \| + C_2 \sum_{n=1}^{z}\EE\|\bar{\omega}_{k-n+1} - \bar{\omega}_{k-n}\| \\
&\quad + C_3 \sum_{m=0}^{z-1}\sum_{n=1}^{z-m} \EE \| \theta_{k-m-n+1} - \theta_{k-m-n}\| + C_8 \kappa\rho^{z-1} \\
&\leq 2C_1 C_{\theta} \sum_{n=1}^{z}\alpha_{k-n} + 2 C_2 C_\delta\sum_{n=1}^{z}\beta_{k-n} + C_3C_{\theta} \sum_{m=0}^{z-1}\sum_{n=1}^{z-m}\alpha_{k-m-n} + C_8 \kappa\rho^{z-1} \\
&\overset{(ii)}{\leq} 2C_1C_{\theta}z\alpha_{k-z} + 2C_2C_\delta z\beta_{k-z} + C_3C_{\theta}z(z-1)\alpha_{k-z} + C_8\kappa\rho^{z-1}, \numberthis
\end{align*}
where the $(i)$ uses triangle inequality, $(ii)$ uses the non-increasing property of step sizes.

Let $z= Z_K$, we have (crf. $Z_K:= \min\{z \in \NN^+ | \kappa\rho^{z-1} \leq \min \{\alpha_K, \beta_K, \eta_K\}\}$)
\begin{align*}
&\EE\langle \bar{\omega}_k - \omega^*(\theta_k), \bar{g}_c(\xi_k, \bar{\omega}_k) - g_c(\theta_k, \bar{\omega}_k) \rangle \\
&\leq 2C_1 C_{\theta} Z_K\alpha_{k-Z_K} + 2 C_2 C_\delta Z_K \beta_{k-Z_K} + C_3 C_{\theta} Z_K^2 \alpha_{k-Z_K} + C_8 \alpha_{k-Z_K}. \numberthis \label{eq:critic-error-10}
\end{align*}

Plug (\ref{eq:critic-error-10}) into (\ref{eq:critic-error-9}) will yield
\begin{align*}
\| \bar{\omega}_{k+1} - \omega^*(\theta_k) \|^2 &\leq 
(1-2\lambda_{\phi} \beta_k) \| \bar{\omega}_k - \omega^*(\theta_k) \|^2 + C_\delta^2\beta_k^2 \\
&\quad + 4C_1 C_{\theta} Z_K\alpha_{k-Z_K} + 4 C_2 C_\delta Z_K \beta_{k-Z_K} + 2C_3 C_{\theta} Z_K^2 \alpha_{k-Z_K} + 2C_8 \alpha_{k-Z_K}. 
%	&\leq (1-2\lambda_{\phi} \beta_k) \| \bar{\omega}_k - \omega^*(\theta_k) \|^2 + C_{K_1} \beta_k\beta_{k-Z_K} + C_{K_2}\alpha_{k-Z_K} \beta_k,
\end{align*}
%where we uses $\|\bar{g}_c(\xi_k, \bar{\omega}_k)\|^2 \leq C_\delta^2$ in the last inequality. 
By defining $C_{K_1}:= 4C_2C_\delta Z_K + C_\delta^2,\ C_{K_2}:= 
4C_1C_{\theta}Z_K + 2C_3C_{\theta}Z_K^2 + C_8$, we complete the proof.
\end{proof}
	
\end{lemma}

\begin{lemma}\label{lemma:critic-markov-error}
	Consider the sequence generated by Algorithm~\ref{algorithm:dec-ac-re}, for any $z \in \NN^+$, we have
	\begin{align*}
	\EE \langle \bar{\omega}_k - \omega^*(\theta_k), \bar{g}_c(\xi_k, \bar{\omega}_k) - g_c(\theta_k, \bar{\omega}_k) \rangle &\leq C_1 \|\theta_k - \theta_{k-z}\| + C_2 \|\bar{\omega}_k - \bar{\omega}_{k-z} \| \\ 
	&+ C_3 \sum_{m=0}^{z-1} \| \theta_{k-m} - \theta_{k-z}\|  + C_8 \kappa\rho^{z-1}, \numberthis \label{eq:critic-markov-product-error}
	\end{align*}
	where $C_1 := 4 R_\omega C_\delta |\cA| L_\pi (1+\log_{\rho}\kappa^{-1} + (1-\rho)^{-1}) + 2 C_\delta L_\omega,  \
	C_2 := 4(1+\gamma) R_\omega + 2 C_\delta, \
	C_3 := 4 R_\omega C_\delta |\cA| L_\pi, \
	C_8 := 8 R_\omega C_\delta.$
\end{lemma}
Basically, this lemma shows that the term on the left hand side of (\ref{eq:critic-markov-product-error}) is of order $\tilde\cO({\alpha_k + \beta_k})$.
\begin{proof}
	Consider the Markov chain since timestep $k-z$:
	%\[s_{k-z}\xrightarrow{\theta_{k-z}}\]
	%\[s_{k-z}\xrightarrow{\theta_{k-z}}a_{k-z}\xrightarrow{\cP}s_{k-z+1}\]
	\[s_{k-z}\xrightarrow{\theta_{k-z}}a_{k-z}\xrightarrow{\cP}s_{k-z+1}
	\xrightarrow{\theta_{k-z+1}} a_{k-z+1} \cdots \xrightarrow{\theta_{k-1}}a_{k-1} \xrightarrow{\cP}s_k\xrightarrow{\theta_k}a_k\xrightarrow{\cP}s_{k+1}.\]
	Also consider the auxiliary Markov chain with fixed policy since timestep $k-z$:
	\[s_{k-z}\xrightarrow{\theta_{k-z}}a_{k-z}\xrightarrow{\cP}s_{k-z+1}
	\xrightarrow{\theta_{k-z}} \tilde{a}_{k-z+1} \cdots \xrightarrow{\theta_{k-z}}\tilde{a}_{k-1} \xrightarrow{\cP}\tilde{s}_k\xrightarrow{\theta_{k-z}}\tilde{a}_k\xrightarrow{\cP}\tilde{s}_{k+1}.\]
	Throughout the proof of this lemma, we will use $\theta, \theta', \bar{\omega}, \bar{\omega}', \xi, \tilde{\xi}$ as shorthand notations of $\theta_k, \theta_{k-z}, \bar{\omega}_k, \bar{\omega}_{k-z},\xi_{k}, \tilde{\xi}_{k}$.
	
	For the ease of expression, define
	\[\Delta_1(\xi, \theta, \omega):= \langle \omega - \omega^*(\theta), \bar{g}_c(\xi, \omega)-g_c(\theta,\omega) \rangle.\]
	
	Therefore, we have 
	\begin{align*}
	\langle \bar{\omega}_k - \omega^*(\theta_k), \bar{g}_c(\xi_k, \bar{\omega}_k) - g_c(\theta_k, \bar{\omega}_k) \rangle &= \Delta_1(\xi, \theta, \bar{\omega}) \\
	&= \underbrace{\Delta_1(\xi,\theta,\bar{\omega}) - \Delta_1(\xi,\theta',\bar{\omega})}_{I_1}  + \underbrace{\Delta_1(\xi,\theta',\bar{\omega}) - \Delta_1(\xi,\theta',\bar{\omega}')}_{I_2}  \\
	&\quad + \underbrace{\Delta_1(\xi,\theta',\bar{\omega}') - \Delta_1(\tilde{\xi},\theta',\bar{\omega}')}_{I_3} + \underbrace{\Delta_1(\tilde{\xi},\theta',\bar{\omega}')}_{I_4}. \numberthis \label{eq:critic-markov-lemma-1}
	\end{align*}
	
	$I_1$ can be expressed as 
	\begin{align*}
	I_1 &= \langle\bar{\omega} - \omega^*(\theta), \bar{g}_c(\xi, \bar{\omega}) - g_c(\theta, \bar{\omega}) \rangle - \langle\bar{\omega} - \omega^*(\theta'), \bar{g}_c(\xi, \bar{\omega}) - g_c(\theta', \bar{\omega}) \rangle \\
	& = \langle\bar{\omega} - \omega^*(\theta), \bar{g}_c(\xi, \bar{\omega}) - g_c(\theta, \bar{\omega}) \rangle - \langle\bar{\omega} - \omega^*(\theta), \bar{g}_c(\xi, \bar{\omega}) - g_c(\theta', \bar{\omega}) \rangle \\
	&\quad + \langle  \omega^*(\theta) - \omega^*(\theta'), \bar{g}_c(\xi, \bar{\omega}) - g_c(\theta', \bar{\omega}) \rangle \\
	&\leq \|\bar{\omega} - \omega^*(\theta)\| \|g_c(\theta', \bar{\omega}) - g_c(\theta, \bar{\omega})\| + \|\omega^*(\theta) - \omega^*(\theta')\|\| \bar{g}_c(\xi, \bar{\omega}) - g_c(\theta', \bar{\omega}) \|. \numberthis \label{eq:critic-markov-lemma-2}
	\end{align*}
	
	%	Let $\xi \sim \theta$ denotes $s \sim \mu_\theta, a \sim \pi_{\theta}, s' \sim \cP$.
	The first term can be bounded as 
	\begin{align*}
	\|\bar{\omega} - \omega^*(\theta)\| \|g_c(\theta', \bar{\omega}) - g_c(\theta, \bar{\omega})\| &\leq 2 R_\omega \|\EE_{\xi \sim \mu_\theta'}[\bar{g}_c(\xi, \bar{\omega})] - \EE_{\xi\sim\mu_\theta}[\bar{g}_c(\xi,\bar{\omega})]\| \\
	&\leq 4R_\omega \sup_{\xi} \| \bar{g}_c(\xi, \bar{\omega}) \| d_{TV}(\mu_\theta'\otimes\pi_\theta'\otimes\cP, \mu_\theta\otimes\pi_\theta\otimes\cP) \\
	&\leq 4R_\omega C_\delta d_{TV}(\mu_\theta'\otimes\pi_\theta'\otimes\cP, \mu_\theta\otimes\pi_\theta\otimes\cP) \\
	%		&= 4 R_\omega C_\delta d_{TV}(\mu_\theta'\otimes\pi_\theta'\otimes\cP, \mu_\theta\otimes\pi_\theta\otimes\cP) \\
	&\leq 4 R_\omega C_\delta |\cA| L_\pi (1+\log_{\rho}\kappa^{-1} + (1-\rho)^{-1}) \|\theta - \theta'\|, \numberthis \label{eq:critic-markov-lemma-3}
	\end{align*}
	where the first inequality is due to $\omega \leq R_{\omega}$ induced by the projection step of critic's update. The third inequality is due to $\|\bar{g}_c(\xi, \bar{\omega})\| \leq C_\delta$, and the last inequality follows Lemma~\ref{lemma:error-total-variation}.
	
	By the Lipschitz conitinuous of $\omega^*(\theta)$ proposed in Proposition~\ref{proposition:lipschitz-continuous-omega}, the second term in (\ref{eq:critic-markov-lemma-2}) can be bounded as 
	\begin{align*}
	\|\omega^*(\theta) - \omega^*(\theta')\|\| \bar{g}_c(\xi, \bar{\omega}) - g_c(\theta, \bar{\omega}) \|
	&\leq L_{\omega}\|\theta-\theta'\| \| \bar{g}_c(\xi, \bar{\omega}) - g_c(\theta, \bar{\omega}) \|\\
	&\leq 2 C_\delta L_\omega \|\theta-\theta'\| \numberthis \label{eq:critic-markov-lemma-4}
	\end{align*}
	%, and $\|\bar{g}_c(\xi,\bar{\omega})\| \leq C_\delta$. 
	
	Plug (\ref{eq:critic-markov-lemma-3}) and (\ref{eq:critic-markov-lemma-4}) into (\ref{eq:critic-markov-lemma-2}), we can bound $I_1$ as
	\begin{align*}
	I_1 \leq (4 R_\omega C_\delta |\cA| L_\pi (1+\log_{\rho}\kappa^{-1} + (1-\rho)^{-1}) + 2 C_\delta L_\omega) \|\theta - \theta'\|. \numberthis \label{eq:critic-markov-lemma-5}
	\end{align*}
	
	$I_2$ can be decomposed by
	\begin{align*}
	I_2 &= \langle \bar{\omega} - \omega^*(\theta'), \bar{g}_c(\xi, \bar{\omega}) - g_c(\theta', \bar{\omega}) \rangle - \langle \bar{\omega}' - \omega^*(\theta'), \bar{g}_c(\xi, \bar{\omega}') - g_c(\theta', \bar{\omega}') \rangle \\
	&= \langle \bar{\omega} - \omega^*(\theta'), \bar{g}_c(\xi, \bar{\omega}) - g_c(\theta', \bar{\omega}) \rangle - \langle \bar{\omega}' - \omega^*(\theta'), \bar{g}_c(\xi, \bar{\omega}) - g_c(\theta', \bar{\omega}) \rangle \\
	&\quad + \langle \bar{\omega}' - \omega^*(\theta'), \bar{g}_c(\xi, \bar{\omega}) - \bar{g}_c(\xi, \bar{\omega}') - g_c(\theta', \bar{\omega}) + g_c(\theta', \bar{\omega}') \rangle \\
	&= \langle \bar{\omega} - \omega^*(\theta'), \bar{g}_c(\xi, \bar{\omega}) - g_c(\theta', \bar{\omega}) \rangle \\ &\quad + \langle \bar{\omega}' - \omega^*(\theta'), \bar{g}_c(\xi, \bar{\omega}) - \bar{g}_c(\xi, \bar{\omega}') - g_c(\theta', \bar{\omega}) + g_c(\theta', \bar{\omega}') \rangle \\
	&\leq 2C_\delta \|\bar{\omega} - \omega^*(\theta')\|  + \langle \bar{\omega}' - \omega^*(\theta'), \bar{g}_c(\xi, \bar{\omega}) - \bar{g}_c(\xi, \bar{\omega}') - g_c(\theta', \bar{\omega}) + g_c(\theta', \bar{\omega}') \rangle.
	\end{align*}
% 	The first term can be bounded as 
% 	\begin{align*}
% 	&\langle \bar{\omega} - \bar{\omega}' , \bar{g}_c(\xi, \bar{\omega}) - g_c(\theta', \bar{\omega}) \rangle  \leq 2C_\delta \| \bar{\omega} - \bar{\omega}' \|. \numberthis \label{eq:critic-markov-lemma-6} 
% 	\end{align*}
	The last term can be bounded as 
	\begin{align*}
	&\langle \bar{\omega}' - \omega^*(\theta'), \bar{g}_c(\xi, \bar{\omega}) - \bar{g}_c(\xi, \bar{\omega}') - g_c(\theta', \bar{\omega}) + g_c(\theta', \bar{\omega}') \rangle \\
	&\leq \| \bar{\omega} - \omega^*(\theta')\|( \| \bar{g}_c(\xi, \bar{\omega}) - \bar{g}_c(\xi, \bar{\omega}')\| + \|g_c(\theta', \bar{\omega}') - g_c(\theta', \bar{\omega}) \|) \\
	&\leq 2R_\omega( \| \bar{g}_c(\xi, \bar{\omega}) - \bar{g}_c(\xi, \bar{\omega}')\| + \|g_c(\theta', \bar{\omega}') - g_c(\theta', \bar{\omega}) \|) \\
	&\leq 2R_\omega( \| \bar{g}_c(\xi, \bar{\omega}) - \bar{g}_c(\xi, \bar{\omega}')\| + \EE_{\xi \sim \mu_{\theta'}}\|\bar{g}_c(\xi, \bar{\omega}') - g_c(\xi, \bar{\omega}) \|) \\
	%	&\leq 2R_\omega(\|\gamma\phi(s')^T(\bar{\omega}-\bar{\omega'})-\phi(s)^T(\bar{\omega}-\bar{\omega'}))\phi(s) \\
	%	&\leq 2R_\omega \|2(1+\gamma) (\bar{\omega} - \bar{\omega}') \| \\
	&\leq 4 R_\omega (1+\gamma) \| \bar{\omega} - \bar{\omega}' \|, \numberthis \label{eq:critic-markov-lemma-7} 
	\end{align*}
	where the first inequality applies Cauchy-Schwartz inequality and triangle inequality, the second inequality follows the projection of each critic step. The last inequality is due to 
	\begin{align*}
	\| \bar{g}_c(\xi, \bar{\omega}) - \bar{g}_c(\xi, \bar{\omega}')\| &= \|\phi(s) (\gamma\phi(s')^T(\bar{\omega}-\bar{\omega}') - \phi(s)^T(\bar{\omega}-\bar{\omega}'))\| \\
	&\leq \gamma \|\phi(s')^T(\bar{\omega}-\bar{\omega}')\| + \|\phi(s)^T(\bar{\omega}-\bar{\omega}')\| \\
	&\leq (1+\gamma) \|\bar{\omega}-\bar{\omega}'\|.
	\end{align*}
	
	Thus, $I_2$ can be bounded as
	\begin{align*}
	I_2 \leq (4(1+\gamma) R_\omega + 2 C_\delta) \| \bar{\omega} - \bar{\omega}' \|. \numberthis \label{eq:critic-markov-lemma-8}
	\end{align*}
	
	We bound $I_3$ as
	\begin{align*}
	\EE[I_3|\theta', s_{k-z+1}] &= \EE[\Delta_1(\xi, \theta', \bar{\omega}') - \Delta_1(\tilde{\xi}, \theta', \bar{\omega}') | \theta', s_{k-z+1}] \\
	& \leq 2\sup_{\xi} |\Delta_1(\xi,\theta', \bar{\omega}')| \ d_{TV}(\PP(\xi\in \cdot| \theta', s_{k-z+1}), \PP(\tilde{\xi}\in \cdot| \theta', s_{k-z+1})) \\
	&\leq 8 R_\omega C_\delta d_{TV}(\PP(\xi\in \cdot| \theta', s_{k-z+1}), \PP(\tilde{\xi}\in \cdot| \theta', s_{k-z+1})) \\
	&\leq 4 R_\omega C_\delta |\cA| L_\pi \sum_{m=0}^{z-1}\|\theta_{k-m} - \theta_{k-z}\|. \numberthis \label{eq:critic-markov-lemma-9}
	\end{align*}
	Here, the second inequality is due to $\|\Delta_1(\xi, \theta', \bar{\omega}')\|\leq \|\omega' - \omega^*(\theta')\| \|\bar{g}_c(\xi, \omega') - g_c(\theta', \omega') \|\leq 4R_\omega C_\delta$, and the last inequality is according to Lemma~\ref{lemma:mismatch-markov-stationary}.
	
	%By lemma.~\ref{}, 
	%\[d_{TV}(\PP(\xi\in \cdot| \theta' \bar{\omega}', s_{k-z+1}), \PP(\tilde{\xi}\in \cdot| \theta' \bar{\omega}', s_{k-z+1})) \leq \frac{1}{2}|\cA| L_\pi \sum_{i=0}^{z}\|\theta_{k-i} - \theta_{k-z}\|.\] 
	
	%Therefore, we have
	%\[\EE[I_3] \leq 4 R_\omega C_\delta |\cA| L_\pi \sum_{i=0}^{z}\|\theta_{k-i} - \theta_{k-z}\|.\]
	We now bound $I_4$
	\begin{align*}
	\EE[I_4|\theta', \bar{\omega}', s_{k+z-1}] &= \EE[\Delta_1(\tilde{\xi}, \theta', \bar{\omega}') | \theta', \bar{\omega}', s_{k-z+1}] \\
	& \leq \sup_{\xi} |\Delta_1(\xi, \theta', \bar{\omega}')| \| \PP(\xi \in \cdot|\theta', s_{k-z+1}) - \mu_{\theta'} \otimes \pi_{\theta'}\otimes\cP\| \\
	&\leq 8 R_\omega C_\delta d_{TV} (\PP(\tilde{x}\in \cdot| \theta', s_{t-z+1}), \mu_{\theta'}\otimes\pi_{\theta'}\otimes\cP) \\
	&\leq 8 R_\omega C_\delta \kappa \rho^{z-1}, \numberthis \label{eq:critic-markov-lemma-10}
	\end{align*}
	where the last inequality follows Lemma~\ref{lemma:mix-markov-chain}.
	%
	%By lemma.~\ref{}, we have
	%\[d_{TV}(\PP(\tilde{x}\in \cdot| \theta', s_{t-z+1}), \mu_\theta'\otimes\pi_\theta'\otimes \cP) \leq \kappa\rho^{z-1}.\]
	%Therefore, $I_4$ can be bounded as 
	%\[\EE[I_4] \leq 8 R_\omega C_\delta \kappa \rho^{z-1}.\]
	
	Plug (\ref{eq:critic-markov-lemma-5}), (\ref{eq:critic-markov-lemma-8}), (\ref{eq:critic-markov-lemma-9}), and (\ref{eq:critic-markov-lemma-10}) into (\ref{eq:critic-markov-lemma-1}), we get
	\begin{align*}
	\EE[\Delta_1(\xi, \theta, \bar{\omega})] &\leq (4 R_\omega C_\delta |\cA| L_\pi (1+\log_{\rho}\kappa^{-1} + (1-\rho)^{-1}) + 2 C_\delta L_\omega) \EE\| \theta_k - \theta_{k-z} \| \\
	& + (4(1+\gamma) R_\omega + 2 C_\delta)\EE \| \bar{\omega}_k - \bar{\omega}_{k-z} \| \\
	&+ (4 R_\omega C_\delta |\cA| L_\pi) \sum_{m=0}^{z-1} \EE\| \theta_{k-m} - \theta_{k-z} \| \\
	&+ 8 R_\omega C_\delta \kappa\rho^{z-1},
	\end{align*}
	which completes the proof.
\end{proof}

\subsection{Error of reward estimator}\label{appendix:reward-estimator-error}
The analysis for the error of reward estimator is similar to critic. To see this, we only need to change $\bar{g}_c(\xi, \bar{\omega})$ into $\bar{g}_r(\xi, \bar{\lambda}):=(r(s,a)-\varphi(s,a)^T\bar{\lambda})\varphi(s,a)$ to recover most of the proofs.

%We provide the reward estimator's analysis for the completeness. For the ease of discussion, we define
%\begin{align*}
%g_r^i(\xi, \lambda) &:= \varphi(s,a)(r^i(s,a) - \varphi(s,a)^T\lambda), \\
%\bar{g}_r(\xi, \lambda) &:= \varphi(s,a)(\bar{r}(s,a) - \varphi(s,a)^T\lambda), \\
%g_r(\theta, \lambda) &:= \EE_{\xi \sim \mu_\theta}[\bar{g}_r(\xi, \lambda)].
%\end{align*}
%Note here $g_r^i(\xi, \lambda)$ and $\bar{g}_r(\xi, \lambda)$ do not depend on the next state $s'$. We use $\xi$ for notational convience.

Lemma~\ref{lemma:lambda-error-markov-1} and \ref{lemma:lambda-markov-error} are the counter parts of Lemma~\ref{lemma:critic-error-markov-1} and \ref{lemma:critic-markov-error} for reward estimator.
\begin{lemma}\label{lemma:lambda-error-markov-1}
	Suppose Assumptions \ref{assump:boundedness} , \ref{assump:negative-definiteness},  \ref{assump:lipschitz-policy}, \ref{assump:markov-chain} hold, with $\lambda_{k+1}$ generated by Algorithm~\ref{algorithm:dec-ac-re} given $\lambda_{k}$ and $\theta_k$ under Markovian sampling, then the following holds
	\begin{align*}
	\EE[\|\bar{\lambda}_{k+1} - \lambda^*(\theta_{k+1})\|^2|\theta_k] &\leq \left(1 + 4L_{\lambda}N \alpha_k + \frac{L_{\lambda, 2}^2}{2} C_\theta^2N\sqrt{d_\theta}\alpha_k^2\right)\EE\|\bar{\lambda}_{k+1} - \lambda^*(\theta_k)\|^2 \\
	&\quad +\left(\frac{L_{\lambda, 2}^2 }{2} C_{\theta}^2 N + L_\lambda^2C_\theta^2N\right) \alpha_k^2 + \frac{\alpha_k}{4}\sum_{i=1}^{N} \| \EE[g_a^i(\xi_k, \lambda_{k+1}^{i}, \lambda_{k+1}^{i})] \|^2. \numberthis \label{eq:lambda-error-7}
	\end{align*}
	\begin{align*}
	\EE[\|\bar{\lambda}_{k+1} - \lambda^*(\theta_k)\|^2|\theta_k] \leq (1-2\eta_k \lambda_{\varphi}) \|\bar{\lambda}_k - \lambda^*(\theta_k)\|^2 + C_{K_3}\eta_k\eta_{k-Z_K} + C_{K_4} \eta_k\alpha_{k-Z_K}, \numberthis \label{eq:lambda-error-8}
	\end{align*}
	where $C_{K_3}:= 4C_6C_\lambda Z_K + C_\lambda^2,\ C_{K_4}:= 
	4C_5C_{\theta}Z_K + 2C_7C_{\theta}Z_K^2 + C_8, Z_K:= \min\{z \in \NN^+ | \kappa\rho^{z-1} \leq \min \{\alpha_K, \beta_K, \eta_K\}\}$, $C_\lambda:= r_{\max} + R_\lambda \geq \max_{s,a,\lambda}\|(r(s,a) - \lambda^T\varphi(s,a)\varphi(s,a))\|$.
\end{lemma}

\begin{lemma}\label{lemma:lambda-markov-error}
	Consider the sequence generated by Algorithm~\ref{algorithm:dec-ac-re}, for any $z \in \NN^+$, we have
	\begin{align*}
	\EE[\langle \bar{\lambda}_k - \lambda^*(\theta), \bar{g}_r(\xi_k,\bar{\lambda}_k) - g_r(\theta_k, \bar{\lambda}_k) \rangle] 
	&\leq 
	C_5\|\theta_k - \theta_{k-z}\| + C_6 \|\lambda_k - \lambda_{k-z} \| \\
	&\quad + C_7 \sum_{m=0}^{z-1} \| \theta_{k-m} - \theta_{k-z} \| + C_8 \kappa\rho^{z-1}, \numberthis \label{eq:lambda-markov-lemma-1}
	\end{align*}
	where $C_5:=4 R_\lambda C_\lambda |\cA| L_\pi (1+\log_{\rho}\kappa^{-1} + (1-\rho)^{-1}) + 2 C_\lambda L_\lambda, C_6:=4 R_\lambda + 2 C_\lambda, C_7:=4 R_\lambda C_\lambda |\cA| L_\pi, C_8:=8 R_\lambda C_\lambda$.
\end{lemma}

\subsection{Consensus error} \label{appendix:consensus-error}
\begin{lemma}[restatement of Lemma~\ref{lemma:consensus-error-main}, bound of consensus error]\label{lemma:consensus-error}
	Define the matrix representation of critics and reward estimators' parameters as $\bomega_k:=[\omega_k^1, \cdots, \omega_k^N]^T$, $\blambda_k:=[\lambda_k^1, \cdots, \lambda_k^N]^T$.
	%	\begin{align*}
	%	\boldsymbol{\omega_k} := \begin{bmatrix}(\omega_k^1)^T \\ \vdots \\ (\omega_k^N)^T \end{bmatrix}, \qquad
	%	\boldsymbol{\lambda_k} := \begin{bmatrix}(\lambda_k^1)^T \\ \vdots \\ (\lambda_k^N)^T \end{bmatrix}.
	%	\end{align*}
	Let $\normalfont{Q:= I-\frac{1}{N}\bone\bone^T}$, then the consensus error can be expressed as $\sum_{i=1}^{N} \| \omega_k^i - \bar{\omega}_k \|^2=\|Q\bomega_k\|^2$, $\sum_{i=1}^{N} \| \lambda_k^i - \bar{\lambda}_k \|^2=\|Q\blambda_k\|^2$. 
	Suppose Asssumption ~\ref{assump:doub_stoch} holds. Let $\{\omega_k^i\}_k, \{\lambda_k^i\}_k$ be the sequence generated by the Algorithm~\ref{algorithm:dec-ac-re},  then for any $k \geq 1$, the following inequalities hold
	\begin{align*}
	\|Q\bomega_k\| \leq \nu^{\frac{k}{K_c}-1}\|\bomega_0\| + 4\sqrt{N}C_\delta \sum_{t=0}^{k}\beta_t \nu^{ \frac{k-t}{K_c}-1} \numberthis \label{eq:consensus-error-1}\\
	\|Q\blambda_k\| \leq \nu^{\frac{k}{K_c}-1}\|\blambda_0\| + 4\sqrt{N}C_\lambda \sum_{t=0}^{k}\beta_t \nu^{ \frac{k-t}{K_c}-1}, \numberthis \label{eq:consensus-error-2}	
	\end{align*}
	where $\nu \in (0,1)$ is the second largest singular value of $W$.
	
\end{lemma}
\begin{proof}
	We will prove the bound in (\ref{eq:consensus-error-1}) for the critic variables. The analysis for reward estimator in (\ref{eq:consensus-error-2}) follows the same routine. To simplify the notation, we will use $g_k^i$ to represent $g_c^i(\xi_k, \omega_k^i)$ throughout the proof of this lemma. We also use $e_k^i$ to represent the projection update $e_k^i:=\Pi_{R_\omega}(\omega_k^i - \beta_k g_k^i) - (\omega_k^i - \beta_k g_k^i)$. Define  $\bar{g}_k:=\frac{1}{N}\sum_{i=1}^{N}g_k^i$; $\bar{e}_k:=\frac{1}{N}\sum_{i=1}^{N}e_k^i$, and the corresponding matrix exressions as
	\begin{align*}
	G_k := 
	\begin{bmatrix}
	(g_k^1)^T, \\ \vdots \\ (g_k^N)^T
	\end{bmatrix},
	E_k :=
	\begin{bmatrix}
	(e_k^1)^T, \\ \vdots \\ (e_k^N)^T
	\end{bmatrix}.
	\end{align*}
	
	According to the update rule of critic variables, the following equalities holds
	%	\[\bomega_{k+1} = W\bomega_k - \beta_k G_k + E_k.\numberthis \label{eq:consensus-error-3}\]
	\begin{align*}
	\bomega_{k+1} = 
	\begin{cases}
	W\bomega_k - \beta_k G_k + E_k, \quad \text{if} \quad  k \hspace{-0.2cm} \mod K_c = 0 \\
	\bomega_k - \beta_k G_k + E_k, \quad \text{otherwise}.
	\end{cases}
	\label{eq:consensus-error-3} \numberthis
	\end{align*}
	To bound the consensus error, We first bound the consensus error of critic's update as
	\begin{align*}
	\|QG_k\| &= \sqrt{\sum_{i=1}^{N}\|g_k^i - \bar{g}_k\|^2} \overset{(i)}{\leq}
	\sqrt{\sum_{i=1}^{N} 2\|g_k^i\|^2 + 2\|\bar{g}_k\|^2} \leq 2\sqrt{N}C_\delta. \numberthis \label{eq:consensus-error-4} \\
	\|QE_k\| &= \sqrt{\sum_{i=1}^{N}\|e_t^i - \bar{e}_t\|^2} \leq \sqrt{\sum_{i=1}^{N}2\|e_k^i\|^2+2\|\bar{e}_k\|^2} \overset{(ii)}{\leq}
	\sqrt{\sum_{i=1}^{N} 2\beta_k^2\|g_k^i\|^2 + 2\beta_k^2\|\bar{g}_k\|^2} \leq 2\beta_k\sqrt{N}C_\delta \numberthis \label{eq:consensus-error-5},
	\end{align*}
	where $(i)$ is due to $\|g_k^i\|\leq C_\delta$; $(ii)$ is ensured by the convexity of the projection set.
	
	We now study the consensus error of critic variables. Let $k'=\lfloor \frac{k}{K_c} \rfloor * K_c$. By the update rule in (\ref{eq:consensus-error-3}), we have 
	\begin{align*}
	Q\bomega_{k'} &= QW\bomega_{k'-1} - \beta_{k'-1}QG_{k'-1} + QE_{k'-1} \\
	&= WQ\bomega_{k'-1} - \beta_{k'-1}QG_{k'-1} + QE_{k'-1} \\
	&= WQ\bomega_{k'-K_c} - \sum_{t=k'-K_c}^{k'-1} \beta_tW^{\lceil k'-1-t \rceil}QG_t + \sum_{k'-K_c}^{t=k'-1}W^{\lceil k'-1-t \rceil}QE_t, \numberthis \label{eq:consensus-error-6}
	\end{align*}
	where the second equality is due to the doubly stochasticity of matrix $W$ implied by Assumption~\ref{assump:doub_stoch}: $\normalfont{QW=W-\frac{1}{N}\bone\bone^TW=W-\frac{1}{N}W\bone\bone^T=WQ.}$ The last equality is indicated by the update rule that
	\begin{align*}
	\bomega_{k'-1} = \bomega_{k'-K_c} - \sum_{t=k'-K_c}^{k'-2} \beta_tG_t + \sum_{t=k'-K_c}^{k'-2} E_t.
	\end{align*}
	Expand the recursion in (\ref{eq:consensus-error-6}), we have
	\begin{align*}
	Q\bomega_{k'} &= W^{\frac{k'}{K_c}}Q\bomega_0 - \sum_{t=0}^{k'-1}W^{c_t}\beta_t Q G_t + \sum_{t=0}^{k'-1}W^{c_t}QE_t, \\
	% &= W^{\frac{k'}{K_c}}Q\bomega_0 - \sum_{t=0}^{k'}W^{\lfloor \frac{k'-t}{K_c}\rfloor}\beta_t Q G_t + \sum_{t=0}^{k'}W^{\lfloor \frac{k'-t}{K_c}\rfloor}QE_t. \\
	\end{align*}
 where $c_t:=\lceil \frac{k'-1-t}{K_c} \rceil$. 
	Therefore, the $k_{th}$ iteration's consensus error can be expressed as 
	\begin{align*}
	Q\bomega_k &= Q\bomega_{k'} - \sum_{t=k'}^{k-1}\beta_tQG_t + \sum_{t=k'}^{k-1} QE_t \\
	&= W^{\frac{k'}{K_c}}Q\bomega_0 - \sum_{t=0}^{k} W^{c_t}\beta_tQG_t + \sum_{t=0}^{k} W^{c_t}QE_t. \numberthis \label{eq:consensus-error-7}
	\end{align*}
    Take norm on the each side of (\ref{eq:consensus-error-7}) and apply triangle inequality, we get
	\begin{align*}
	\|Q\bomega_k\| &\leq \|W^{\frac{k'}{K_c}}Q\bomega_0\| + \sum_{t=0}^{k} \beta_t \|W^{c_t}QG_t\| + \sum_{t=0}^{k} \|W^{c_t}QE_t\| \\
	&\overset{(i)}{\leq} \nu^{\frac{k'}{K_c}}\|\bomega_0\| + \sum_{t=0}^{k}\beta_t \nu^{c_t}\|G_t\| + \sum_{t=0}^{k}\nu^{c_t}\|E_t\| \\
	&\overset{(ii)}{\leq} \nu^{\frac{k'}{K_c}}\|\bomega_0\| + 4\sqrt{N}C_\delta\sum_{t=0}^{k} \beta_t \nu^{c_t} \\
	&\overset{(iii)}{\leq} \nu^{\frac{k}{K_c}-1}\|\bomega_0\| + 4\sqrt{N}C_\delta \sum_{t=0}^{k}\beta_t \nu^{ \frac{k-t}{K_c}-1}.
	\end{align*}
	where $(i)$ inequality uses Lemma~\ref{lemma:consensus-matrix} and the fact that the spectral of $Q$ is less than 1; $(ii)$ is due to (\ref{eq:consensus-error-4}) and (\ref{eq:consensus-error-5}); $(iii)$ uses the fact that $\frac{k'}{K_c} \geq \frac{k}{K_c}-1$ and $\lceil \frac{k'-1-t}{K_c} \rceil = \lceil \frac{k-t}{K_c} + \frac{k'-k-1}{K_c} \rceil \geq \lceil \frac{k-t}{K_c} \rceil - 1 \geq \frac{k-t}{K_c} -1 $. Thus, the proof for (\ref{eq:consensus-error-1}) is completed. The proof of (\ref{eq:consensus-error-2}) follows a similar procedure, we leave it as an exercise to reader.

\end{proof}

\subsection{Error of actor}
The following lemma characterizes the sampling error of actor.

\begin{lemma}\label{lemma:actor-gradient-bias}
	Consider the sequence generated by Algorithm~\ref{algorithm:dec-ac-re}, for any $z \geq 1$ we have 
	\begin{align*}
	&\| \EE_{\xi \sim \mu_{\theta_k}}[\delta(\xi,\theta_k) \psi_{\theta_k^i}(s_k, a_k^i)] - \EE[\delta(\xi_k,\theta_k) \psi_{\theta_k^i}(s_k, a_k^i)]\| \\
	&\leq 2C_{\theta}\kappa\rho^{z-1} + C_{12}\sum_{m=0}^{z-1} \|\theta_{k-m}- \theta_{k-z}\| + C_{13} \|\theta_k - \theta_{k-z}\| +  C_{14} \|\theta_{k}^{i} - \theta_{k-z}^{i}\|, \numberthis \label{eq:grad-bias-1}
	\end{align*}
	where $C_{12}:=2C_\theta |\cA|L_\pi, \ C_{13}:=|\cA| L (\log_{\rho}\kappa^{-1} + (1-\rho)^{-1})C_\theta + 2(1+\gamma)L_V, \ C_{14}:=2 C_{\delta}L_\psi$.
	\begin{proof}
		Consider the Markov chain since timestep $k-z$:
		%\[s_{k-z}\xrightarrow{\theta_{k-z}}\]
		%\[s_{k-z}\xrightarrow{\theta_{k-z}}a_{k-z}\xrightarrow{\cP}s_{k-z+1}\]
		\[s_{k-z}\xrightarrow{\theta_{k-z}}a_{k-z}\xrightarrow{\cP}s_{k-z+1}
		\xrightarrow{\theta_{k-z+1}} a_{k-z+1} \cdots \xrightarrow{\theta_{k-1}}a_{k-1} \xrightarrow{\cP}s_k\xrightarrow{\theta_k}a_k\xrightarrow{\cP}s_{k+1}.\]
		Also consider the auxiliary Markov chain with fixed policy since timestep $k-z$:
		\[s_{k-z}\xrightarrow{\theta_{k-z}}a_{k-z}\xrightarrow{\cP}s_{k-z+1}
		\xrightarrow{\theta_{k-z}} \tilde{a}_{k-z+1} \cdots \xrightarrow{\theta_{k-z}}\tilde{a}_{k-1} \xrightarrow{\cP}\tilde{s}_k\xrightarrow{\theta_{k-z}}\tilde{a}_k\xrightarrow{\cP}\tilde{s}_{k+1}.\]
		Throughout the proof of this lemma, we wil use $\psi_{\theta^i}$ to represent $\psi_{\theta^i}(s_k, a_k^i)$ for brevity. 
		%and we use $\theta, \theta', \xi, \tilde{\xi}$ as shorthand notations of $\theta_k, \theta_{k-z},\xi_{k}, \tilde{\xi}_{k}$. 
		
		We define the following notation for the ease of discussion
		\begin{align*}
		\Delta_3(\xi, \theta) := \EE_{\xi \sim \mu_\theta}[\delta(\xi, \theta)\psi_{\theta^i}] - \delta(\xi, \theta) \psi_{\theta^i}].
		\end{align*}
		Then our objective is to bound
		\begin{align*}
		\EE[\|\Delta_3(\xi_k,\theta_k)\| | \ \theta_{k-z}].
		\end{align*}
		We decompose $\|\Delta_3(\xi_k, \theta_k)\|$ by applying triangle inequality
		%\begin{align*}
		%	\Delta_3(\xi_k, \theta_k) &= \Delta_3(\xi_k, \theta_k) - \Delta_3(\xi_k, \theta_{k-z}) \\
		%	&\quad + \Delta_3(\xi_k, \theta_{k-z}) - \Delta_3(\tilde{\xi}_k, \theta_{k-z}) \\
		%	&\quad + \Delta_3(\tilde{\xi}_k, \theta_{k-z}).
		%\end{align*}
		%Apply triangle inequality, we have
		\begin{align*}
		\|\Delta_3(\xi_k, \theta_k)\| &\leq \underbrace{\|\Delta_3(\xi_k, \theta_k) - \Delta_3(\xi_k, \theta_{k-z})\|}_{I_1} \\
		&\quad + \underbrace{\|\Delta_3(\xi_k, \theta_{k-z}) - \Delta_3(\tilde{\xi}_k, \theta_{k-z})\|}_{I_2} \\
		&\quad + \underbrace{\|\Delta_3(\tilde{\xi}_k, \theta_{k-z})\|}_{I_3}. \numberthis \label{eq:grad-bias-2}
		\end{align*}
		
		We apply triangle inequality again to bound $I_1$ as
		\begin{align*}
		I_1 &\leq \underbrace{\| \delta(\xi_k, \theta_{k-z})\psi_{\theta_{k-z}^i} - \delta(\xi_k, \theta_{k})\psi_{\theta_{k}^i}\|}_{I_1^{(1)}} \\
		&\quad + \underbrace{\| \EE_{\xi \sim \mu_{\theta_k}}[\delta(\xi, \theta_k)\psi_{\theta_k^i}] - \EE_{\xi \sim \mu_{\theta_{k-z}}}[\delta(\xi, \theta_{k-z})\psi_{\theta_{k-z}^i}] \|}_{I_1^{(2)}} \numberthis \label{eq:grad-bias-3}
		\end{align*}
		
		$I_1^{(1)}$ can be bounded as 
		\begin{align*}
		I_1^{(1)} &= \| \delta(\xi_k, \theta_{k-z})\psi_{\theta_{k-z}^i} - \delta(\xi_k, \theta_{k})\psi_{\theta_{k}^i}\| \\
		&\leq \| \delta(\xi_k, \theta_{k-z})\psi_{\theta_{k-z}^i} - \delta(\xi_k, \theta_{k})\psi_{\theta_{k-z}^i}\| \\
		&\quad + \| \delta(\xi_k, \theta_{k})\psi_{\theta_{k-z}^i} - \delta(\xi_k, \theta_{k})\psi_{\theta_{k}^i}\| \\
		&\leq \| |\gamma(V_{\theta_{k-z}}(s') - V_{\theta_k}(s')) + (V_{\theta_{k-z}}(s) - V_{\theta_{k-z}}(s')) | \psi_{k-z}^{i} \| \\
		&\quad +\| \delta(\xi_k, \theta_k)\psi_{\theta_{k-z}^{i}} - \delta(\xi_k, \theta_k)\psi_{\theta_k^i}\| \\
		&\leq (1+\gamma)L_V \|\theta_k - \theta_{k-z}\| + \| \delta(\xi_k, \theta_k)\psi_{\theta_{k-z}^{i}} - \delta(\xi_k, \theta_k)\psi_{\theta_k^i}\| \\
		&\leq (1+\gamma)L_V \|\theta_k - \theta_{k-z}\| + C_\delta L_{\psi}\|\theta_k^i - \theta_{k-z}^{i}\|, \numberthis \label{eq:grad-bias-4}
		\end{align*}
		where the second last inequality follows the Lipschitz continuous of value function in Lemma~\ref{lemma:lipschitz-continuous-value}, and the last inequality uses Lipschitz continuous of $\psi_{\theta^i}$.
		
		$I_1^{(2)}$ can be bounded as 
		\begin{align*}
		I_1^{(2)} &=
		\| \EE_{\xi \sim \mu_{\theta_k}}[\delta(\xi, \theta_k)\psi_{\theta_k^i}] - \EE_{\xi \sim \mu_{\theta_{k-z}}}[\delta(\xi, \theta_{k-z})\psi_{\theta_{k-z}^i}] \| \\
		&= \| \EE_{\xi \sim \mu_{\theta_k}}[\delta(\xi, \theta_{k-z})\psi_{\theta_{k-z}^i}] - \EE_{\xi \sim \mu_{\theta_{k-z}}}[\delta(\xi, \theta_{k-z})\psi_{\theta_{k-z}^i}] \\
		&\quad + \EE_{\xi\sim\mu_{\theta_k}}[\delta(\xi, \theta_{k})\psi_{\theta_{k}^i} - \delta(\xi, \theta_{k-z})\psi_{\theta_{k-z}^i}] \| \\
		&\leq |\cA| L (\log_{\rho}\kappa^{-1} + (1-\rho)^{-1})C_\theta \|\theta_k - \theta_{k-z}\| \\
		&\quad + \|\EE_{\xi\sim\mu_{\theta_k}}[\delta(\xi, \theta_{k})\psi_{\theta_{k}^i} - \delta(\xi, \theta_{k-z})\psi_{\theta_{k-z}^i}] \| \\
		&\leq |\cA| L (\log_{\rho}\kappa^{-1} + (1-\rho)^{-1})C_\theta \|\theta_k - \theta_{k-z}\| \\
		&\quad + (1+\gamma)L_V \|\theta_k - \theta_{k-z}\| + C_\delta L_{\psi}\|\theta_k^i - \theta_{k-z}^{i}\|, \numberthis \label{eq:grad-bias-5}
		\end{align*}
		where the first inequality applies Lemma~\ref{lemma:error-total-variation}, and the last inequality uses the derivation in (\ref{eq:grad-bias-4}).
		
		Combine (\ref{eq:grad-bias-4}) and (\ref{eq:grad-bias-5}), we have
		\begin{align*}
		I_1 &\leq |\cA| L (\log_{\rho}\kappa^{-1} + (1-\rho)^{-1})C_\theta \|\theta_k - \theta_{k-z}\| \\
		&\quad + 2(1+\gamma)L_V \|\theta_k - \theta_{k-z}\| + 2C_\delta L_{\psi}\|\theta_k^i - \theta_{k-z}^{i}\| \numberthis \label{eq:grad-bias-6}
		\end{align*}
		
		We now bound $I_2$ as 
		\begin{align*}
		\EE[I_2] &= \EE\|\delta(\tilde{\xi}_k, \theta_{k-z})\psi_{\theta_{k-z}}^i - \delta(\xi_k, \theta_{k-z})\psi_{\theta_{k-z}}^i \| \\
		&\leq 2 \sup_{\xi} \|\delta(\xi, \theta_{k-z})\psi_{\theta_{k-z}^{i}} \| d_{TV}(P(\tilde{\xi}_k \in \cdot | \theta_{k-z}, s_{k-z}), P(\xi_k \in \cdot | \theta_{k-z}, s_{k-z})) \\
		&\leq 2 C_\theta \sum_{m=0}^{z-1} |\cA| L_\pi\|\theta_{k-m}-\theta_{k-z}\|, \numberthis \label{eq:grad-bias-7}
		\end{align*}
		where the last inequality follows Lemma~\ref{lemma:mismatch-markov-stationary}.
		
		$I_3$ can be bounded as 
		\begin{align*}
		I_3 &= \EE\|\EE_{\xi \sim \mu_{\theta_{k-z}}}[\delta(\xi, \theta_{k-z}) \psi_{k-z}^{i}] - \delta(\tilde{\xi}_k, \theta_{k-z}\psi_{\theta_{k-z}}^i) \| \\
		&\leq 2 \sup_{\xi} \|\delta(\xi, \theta_{k-z})\psi_{\theta_{k-z}}^i\| d_{TV}(P(\tilde{\xi}\in \cdot|\theta_{k-z},s_{k-z}), \mu_{\theta_{k-z}} \otimes \pi_{\theta_{k-z}} \otimes \cP) \\
		&\leq 2 C_\theta \kappa \rho^{z-1}, \numberthis \label{eq:grad-bias-8}
		\end{align*}
		where the last inequality follows Lemma~\ref{lemma:mix-markov-chain}.
		
		Plug (\ref{eq:grad-bias-6}), (\ref{eq:grad-bias-7}), and (\ref{eq:grad-bias-8}), we have
		\begin{align*}
		&\| \EE_{\xi \sim \mu_{\theta_k}}[\delta(\xi,\theta_k) \psi_{\theta_k^i}(s_k, a_k^i)] - \EE[\delta(\xi_k,\theta_k) \psi_{\theta_k^i}(s_k, a_k^i)]\| \\
		&\leq  2 C_\theta \kappa \rho^{z-1} + 2C_\delta L_{\psi}\|\theta_k^i - \theta_{k-z}^{i}\| + 2 C_\theta \sum_{m=0}^{z-1} |\cA| L_\pi\|\theta_{k-m}-\theta_{k-z}\|\\
		&\quad + (|\cA| L (\log_{\rho}\kappa^{-1} + (1-\rho)^{-1})C_\theta + 2(1+\gamma)L_V) \|\theta_k - \theta_{k-z}\|,
		\end{align*}
		which completes the proof.

	\end{proof}
\end{lemma}
%\clearpage

\section{Proof of main results}

\subsection{Proof of Theorem~\ref{thm:main-theorem-markovian}}\label{appendix:proof-thm-markovian}

Let $\theta_k \in \RR^{Nd_\theta}$ be the stack of actors' parameter at timestep $k$. By Lemma~\ref{lemma:smooth-objective}, we have
\begin{align*}
\EE\left[J(\theta_{k+1})\right] - J(\theta_k) &\geq \EE\left[\langle \nabla J(\theta_k) , \theta_{k+1} - \theta_k\rangle\right] - \frac{L}{2}\|\theta_{k+1} - \theta_k\|^2 \\
& = \sum_{i=1}^{N} \EE\left[\langle \nabla_{\theta^i} J(\theta_k), \theta_{k+1}^{i} - \theta_{k}^{i} \rangle\right] - \frac{L}{2} \sum_{i=1}^{N} \| \theta_{k+1}^i - \theta_k^i \|^2 \\
&= \sum_{i=1}^{N} \EE\left[\alpha_k \langle \nabla_{\theta^i} J(\theta_k), g_a^i(\xi_k , \omega_{k+1}^i, \lambda_{k+1}^i) \rangle\right] - \frac{L}{2}\alpha_k^2 \sum_{i=1}^{N} \EE\| g_a^i(\xi_k, \omega_{k+1}^i, \lambda_{k+1}^i) \|^2 \\
&\geq \sum_{i=1}^{N} \left[\frac{\alpha_k}{2} \| \nabla_{\theta^i} J(\theta_k)\|^2 + \frac{\alpha_k}{2}\|\EE\left[g_a^i(\xi_k, \omega_{k+1}^i, \lambda_{k+1}^i)\right]\|^2 \right. \\
&\quad - \left. \frac{\alpha_k}{2}\|\nabla_{\theta^i}J(\theta_k) - \EE\left[g_a^i(\xi_k, \omega_{k+1}^i, \lambda_{k+1}^i)\right] \|^2\right] - \frac{L}{2} N C_{\theta}^2 \alpha_k^2, \numberthis \label{eq:markov-thm-err-decouple}
\end{align*}
where the expectation is taken over $\xi_k$ under Markovian sampling. The last inequality is due to
\[\left \|g_a^i(\xi_k, \omega_{k+1}^i, \lambda_{k+1}^i) \right\| = \left\|\hat{\delta}(\xi_k, \omega_k^i, \lambda_k^i)\psi_{\theta_k^i}(s_k, a_k^i) \right\| \leq C_\delta C_\psi:=C_\theta \numberthis \label{eq:def-c-theta}.\]

For brevity, we will use $\psi_{\theta_k^i}$ to represent $\psi_{\theta_k^i}(s_k, a_k^i)$.
The gradient bias can be bounded as 
\begin{align*}
&\left\| \nabla_{\theta^i} J(\theta_k) - \EE\left[g_a^i(\xi_k, \omega_{k+1}^i, \lambda_{k+1}^i) | \omega_{k+1}^i , \lambda_{k+1}^i\right] \right\|^2 \\
&\leq 4 \underbrace{\left\| \nabla_{\theta^i} J(\theta_k) - \EE\left[\delta(\xi_k, \theta_k) \psi_{\theta_k^i}\right]\right\|^2}_{I_1} +  4 \underbrace{\left\| \EE\left[(\delta(\xi_k, \theta_k) - \tilde{\delta}(\xi_k, \omega^*(\theta_k))) \psi_{\theta_k^i}\right] \right\|^2}_{I_2} \\
&\quad + 4 \underbrace{\left\| \EE\left[(\tilde{\delta}(\xi_k, \omega^*(\theta_k)) -\tilde{\delta}(\xi_k, \omega_{k+1}^{i})) \psi_{\theta_k^i}\right]\right\|^2}_{I_3} +  4 \underbrace{\left\| \EE\left[(\tilde{\delta}(\xi_k, \omega_{k+1}^{i}) -\hat{\delta}(\xi_k, \omega_{k+1}^{i}, \lambda_{k+1}^{i})) \psi_{\theta_k^i}\right] \right\|^2}_{I_4}, \numberthis \label{eq:markov-thm-gradbias}
\end{align*}
where the inequality uses $\|a+b+c+c\|^2 \leq 4\|a\|^2 + 4\|b\|^2 + 4\|c\|^2 + 4\|d\|^2$.

% By following the derivation of (\ref{eq:iid-thm-2}), the gradient bias can be bounded as (crf. $\psi_{\theta_k^i}:=\psi_{\theta_k^i}(s_k, a_k^i)$)

% \begin{align*}
% &\left \| \nabla_{\theta^i} J(\theta_k) - \EE\left[g_a^i(\xi_k, \omega_{k+1}^i, \lambda_{k+1}^i) | \omega_{k+1}^i , \lambda_{k+1}^i\right] \right \|^2 \\
% &\leq 4 \underbrace{\left\| \nabla_{\theta^i} J(\theta_k) - \EE\left[\delta(\xi_k, \theta_k) \psi_{\theta_k^i}\right]\right\|^2}_{I_1} 
% + 4 \underbrace{\left\| \EE\left[\left(\delta(\xi_k, \theta_k) - \tilde{\delta}\left(\xi_k, \omega^*(\theta_k)\right)\right) \psi_{\theta_k^i}\right] \right\|^2}_{I_2} \\
% &\quad + 4 \underbrace{\left\| \EE\left[\left(\tilde{\delta}(\xi_k, \omega^*(\theta_k)) -\tilde{\delta}(\xi_k, \omega_{k+1}^{i})\right) \psi_{\theta_k^i}\right]\right\|^2}_{I_3} 
% + 4 \underbrace{\left\| \EE\left[\left(\tilde{\delta}(\xi_k, \omega_{k+1}^{i}) -\hat{\delta}(\xi_k, \omega_{k+1}^{i}, \lambda_{k+1}^{i})\right) \psi_{\theta_k^i}\right] \right\|^2}_{I_4}, \numberthis \label{eq:markov-thm-2}
% \end{align*}

We bound $I_1$ as
\begin{align*}
I_1 &= \left\| \nabla_{\theta^i} J(\theta_k) - \EE\left[\delta(\xi_k, \theta_k) \psi_{\theta_k^i} | \theta_k \right] \right\|^2 \\
&= \left\| \EE_{\xi \sim d_{\theta_k}} \left[\delta(\xi, \theta_k) \psi_{\theta_k^i}| \theta_k\right] - \EE\left[\delta(\xi_k, \theta_k) \psi_{\theta_k^i}|\theta_k\right] \right\|^2 \\
&\leq 2 \underbrace{\left\| \EE_{\xi \sim d_{\theta_k}} \left[\delta(\xi, \theta_k) \psi_{\theta_k^i}| \theta_k\right] - \EE_{\xi \sim \mu_{\theta_k}} \left[\delta(\xi, \theta_k) \psi_{\theta_k^i}| \theta_k\right]\right\|^2}_{I_{1}^{(1)}} + 2\underbrace{\left\|\EE_{\xi \sim \mu_{\theta}} \left[\delta(\xi, \theta_k) \psi_{\theta_k^i}| \theta_k\right] - \EE \left[\delta(\xi_k, \theta_k) \psi_{\theta_k^i}| \theta_k\right]\right\|^2}_{I_{1}^{(2)}} \numberthis \label{eq:markov-thm-3}
\end{align*}

From now on, we will use $\xi \sim d_{\theta}$ to denote $s \sim d_{\pit}, a \sim \pi(\cdot|s), s' \sim \cP$ for notational simplicity. $I_1$ is the sampling error under perfect value function estimation of critic. It can be bounded as 
\begin{align*}
\EE\left[I_{1}^{(1)} | \theta_k\right] &= \left\| \nabla_{\theta^i} J(\theta_k) - \EE\left[\delta(\xi_k, \theta_k) \psi_{\theta_k^i} | \theta_k\right] \right\|^2 \\
&= \left\| \EE_{\xi \sim d_{\theta_k}} \left[\delta(\xi, \theta_k) \psi_{\theta_k^i}| \theta_k\right] - \EE_{\xi \sim \mu_{\theta_k}} \left[\delta(\xi, \theta_k) \psi_{\theta_k^i}| \theta_k\right] \right\|^2 \\
&\leq \left(2 \sup_{\xi}\left \|\delta(\xi, \theta_k)\psi_{\theta_k^i} \right\| \ d_{TV}(\mu_{\theta_k}\otimes \pi_{\theta_k} \otimes \cP, d_{\theta_k}\otimes \pi_{\theta_k} \otimes \cP)\right)^2 \\
&\overset{(i)}{\leq} \left(2 C_\theta d_{TV}(\mu_{\theta_k}, d_{\theta_k})\right)^2 \\
&\overset{(ii)}{\leq} 16 C_{\theta}^2 (\log_\rho\kappa^{-1} + \frac{1}{\rho})^2 (1-\gamma^2),
\end{align*}
where $(i)$ uses (\ref{eq:def-c-theta}); $(ii)$ follows Lemma~\ref{lemma:mismatch-visitation-stationary}. Define $\varepsilon_{sp}:=4C_{\theta}^2 (\log_\rho\kappa^{-1} + \frac{1}{\rho})^2 (1-\gamma)^2$, then we have
\[I_{1}^{(1)} \leq 4\varepsilon_{sp}. \numberthis \label{eq:markov-thm-i11-bound}\]

% Follow the derivation of (\ref{eq:iid-thm-tmp}), we have
% \[I_1^{(1)} \leq 4\varepsilon_{sp}. \]

By Lemma~\ref{lemma:actor-gradient-bias}, $I_1^{(2)}$ can be bounded as 
\begingroup
\allowdisplaybreaks
\begin{align*}
I_1^{(2)} & \leq \left(2C_{\theta}\kappa\rho^{z-1} + C_{12}\sum_{m=0}^{z-1} \|\theta_{k-m}- \theta_{k-z}\| + C_{13} \|\theta_k - \theta_{k-z}\| +  C_{14} \|\theta_{k}^{i} - \theta_{k-z}^{i}\|\right)^2 \\
&\leq \left(2C_{\theta}\kappa\rho^{z-1} + C_{12}\sum_{m=0}^{z-1}\sum_{n=1}^{z-m}\|\theta_{k-m-n+1} - \theta_{k-m}\| + C_{13} \sum_{n=1}^{z}\|\theta_{k-n+1} - \theta_{k-n}\| + C_{14} \sum_{n=1}^{z}\|\theta_{k-n+1}^i - \theta_{k-n}^i\|\right)^2 \\
&\leq \left(2C_{\theta}\kappa\rho^{z-1} + C_{12}NC_\theta \frac{z(z+1)}{2}\alpha_{k-z} + C_{13}Nz C_\theta \alpha_{k-z} + C_{14} zC_\theta \alpha_{k-z}\right)^2 \\
&\leq 16C_\theta^2\kappa^2\rho^{2z-2}+2C_{12}^2C_\theta^2z^2\alpha_{k-z}^2 + 4C_{13}^2N^2z^2C_\theta^2\alpha_{k-z}^2 + 4C_{14}^2z^2C_\theta^2\alpha_{k-z}^2, \numberthis \label{eq:markov-thm-4}
\end{align*}
\endgroup
where the second inequality uses triangle inequality, and the last inequality applies $(a+b+c+d)^2\leq 4a^2 + 4b^2 + 4c^2 + 4d^2$. Let $z=Z_K:= \min\{z\in \NN^+| \kappa\rho^{z-1} \leq \min \{\alpha_K, \beta_K, \eta_K\}\}$. Then we have
\begin{align*}
I_1^{(2)} \leq C_{K_5} \alpha_{k-Z_K}^2, \numberthis \label{eq:markov-thm-5}
\end{align*}
%where $C_{K_5}:=48C_{\theta}^2 + 12 N^2 | \cA |^2 L_\pi^2Z_K^4C_{\theta}^2 + 12 C_{\theta}^4Z_K^2$.
where we define $C_{K_5}:=16C_\theta^2+2C_{12}^2C_\theta^2Z_K^2+4C_{13}^2N^2Z_K^2C_\theta^2+4C_{14}^2Z_K^2C_\theta^2$. Thus, we have
\[I_1 \leq 4\varepsilon_{sp} + C_{K_5}\alpha_{k-Z_K}^2 \numberthis \label{eq:markov-thm-i1-bound}.\]

The term $I_2$ describes the approximation quality of linear function class, it can be bounded as
\begin{align*}
I_2 &= \left \| \EE\left[(\delta(\xi_k, \theta_k) - \tilde{\delta}(\xi_k, \omega^*(\theta_k))) \psi_{\theta_k^i}\right] \right\|^2 \\
&\overset{(i)}{\leq} \EE\left[\left|\delta(\xi_k, \theta_k) - \tilde{\delta}(\xi_k, \omega^*(\theta_k))\right|^2 \left \| \psi_{\theta_k^i} \right \|^2\right] \\
&\overset{(ii)}{\leq} C_{\psi}^2 \EE\left[\gamma \left| V_{\pi_{\theta_k}}(s_{k+1}) - \hat{V}_{\omega^*(\theta_k)}(s_{k+1})\right| + \left| V_{\pi_{\theta_k}}(s_{k}) - \hat{V}_{\omega^*(\theta_k)}(s_{k})\right|\right] \\
&\overset{(iii)}{\leq} 2C_\psi^2 \left(\gamma^2\EE\left[\left|V_{\theta_k}(s_{k+1}) - \hat{V}_{\omega^*(\theta_k)}(s_{k+1})\right|^2\right] + \EE\left[\left|V_{\theta_k}(s_{k}) - \hat{V}_{\omega^*(\theta_k)}(s_{k})\right|^2\right]\right) \\
&\overset{(iiii)}{\leq} 2C_\psi^2(1+\gamma^2) \varepsilon_{app}^{c} \leq 4C_\psi^2\varepsilon_{app}^c. \numberthis \label{eq:markov-thm-i2-bound}
\end{align*}
where $(i)$ applies Cauchy Schwarz inequality and triangle inequality; $(ii)$ is due to $\|\psi_{\theta_k^i}\| \leq C_\psi$, which is ensured by Assumption~\ref{assump:lipschitz-policy}; $(iii)$ uses $|a+b|^2 \leq 2|a|^2 + 2|b|^2$; $(iiii)$ follows the definition of the critic's approximation error:
\[\varepsilon_{app}^c:=\max_{\theta}\sqrt{\EE_{s\sim\mu_{\theta}}\left[\left|V_{\pit}(s) - \hat{V}_{\omega^*(\theta)}(s)\right|^2\right]}. \numberthis \label{eq:def-approx-error-critic} \]

$I_3$ captures the error of critic's estimator, which can be bounded as
\begin{align*}
\EE[I_3] &= \left\| \EE\left[\left(\tilde{\delta}(\xi_k, \omega^*(\theta_k)) - \tilde{\delta}(\xi_k, \omega_{k+1}^i)\right) \psi_{\theta_k^i}\right] \right\|^2 \\
&\leq \EE\left[\left|\tilde{\delta}(\xi_k, \omega^*(\theta_k)) - \tilde{\delta}(\xi_k, \omega_{k+1}^i)\right|^2 \left\| \psi_{\theta_k^i}\right\|^2 \right] \\
&\leq C_\psi^2 \EE\left[\left|\gamma \phi(s_{k+1})^T\left(\omega^*(\theta_k) - \omega_{k+1}^i\right) - \phi(s_k)^T\left(\omega^*(\theta_k) - \omega_{k+1}^i\right)\right|^2 \right] \\
&\leq C_\psi^2 \left(2\EE\left[\left|\gamma \phi(s_{k+1})^T\left(\omega^*(\theta_k)-\omega_{k+1}^i\right)\right|^2\right] + 2\EE\left[\left|\phi(s_k)^T\left(\omega^*(\theta_k) - \omega_{k+1}^i\right)\right|^2\right]\right) \\
&\leq C_\psi^2\left(2\gamma^2 \EE\left[\left\|\phi(s_{k+1})\right\|^2 \left\|\omega^*(\theta_k)-\omega_{k+1}^i\right\|^2\right] + 2 \EE\left[\left\|\phi(s_k)\right\|^2 \left\| \omega^*(\theta_k) - \omega_{k+1}^i\right\|^2\right]\right) \\
&{\leq} 2C_\psi^2(1+\gamma^2) \left\| \omega^*(\theta_k) - \omega_{k+1}^i\right\|^2 \leq 4C_\psi^2\left\| \omega^*(\theta_k) - \omega_{k+1}^i\right\|^2, \numberthis \label{eq:markov-thm-i3-bound}
\end{align*}
where the last inequality is due to $\|\phi(s)\|\leq 1$, which is specified by Assumption~\ref{assump:boundedness}. 

$I_4$ characterizes the error of reward estimator, which can be bounded as
\begin{align*}
\EE[I_4] &= \left\| \EE\left[\left(\tilde{\delta}(\xi_k, \omega_{k+1}^{i}) -\hat{\delta}(\xi_k, \omega_{k+1}^{i}, \lambda_{k+1}^{i})\right) \psi_{\theta_k^i} | \lambda_{k+1}^i\right] \right\|^2 \\
&\leq \EE\left[\left|\tilde{\delta}(\xi_k, \omega_{k+1}^{i}) -\hat{\delta}(\xi_k, \omega_{k+1}^{i}, \lambda_{k+1}^{i})\right|^2 \left\| \psi_{\theta_k^i} \right\|^2 | \lambda_{k+1}^i\right] \\
&\leq C_\psi^2 \EE\left[\left|\bar{r}(s_k, a_k) - \varphi(s_k, a_k)^T \lambda_{k+1}^i\right|^2 | \lambda_{k+1}^i\right] \\
&\leq C_\psi^2 \left(2 \EE\left[\left|\bar{r}(s_k, a_k) - \varphi(s_k, a_k)^T \lambda^*(\theta_k)\right|^2\right] + 2 \EE\left[\left|\varphi(s_k, a_k)^T \lambda^*(\theta_k) - \varphi(s_k, a_k)^T \lambda_{k+1}^i \right|^2 | \lambda_{k+1}^i\right]\right) \\
&\leq 2 C_\psi^2 \varepsilon_{app}^{r} + 2C_\psi^2\left\|\lambda^*(\theta_k) - \lambda_{k+1}^i\right\|^2, \numberthis \label{eq:markov-thm-i4-bound}
\end{align*}
where the $\varepsilon_{app}^{r}$ in the last inequality is the approximation error of reward estimator, which is defined as
\[\varepsilon_{app}^{r} := \max_{\theta, a} \sqrt{\EE_{s\sim \mu_{\theta}}\left[\left|\bar{r}(s,a) - \hat{r}_{\lambda^*(\theta)}(s,a)\right|^2\right]} .\]

% The bound of $I_2, I_3,$ and $I_4$ follows the analysis under i.i.d. sampling. Plug in (\ref{eq:iid-thm-tmp2}), (\ref{eq:iid-thm-tmp3}), and (\ref{eq:iid-thm-tmp4}) will give us the bound of gradient bias
Combining (\ref{eq:markov-thm-i1-bound}), (\ref{eq:markov-thm-i2-bound}), (\ref{eq:markov-thm-i3-bound}), and (\ref{eq:markov-thm-i4-bound}) gives us the bound of the gradient bias error as
\begin{align*}
\|\nabla_{\theta^i} F(\theta_k) - \EE[g_a^i(\xi_k, \omega_{k+1}^i, \lambda_{k+1}^i)]\|^2
&\leq 16(\varepsilon_{sp}+C_\psi^2\varepsilon_{app}) + 16 C_\psi^2 \| \omega^*(\theta_k) - \omega_{k+1}^i\|^2 \\
&\quad + 8C_\psi^2\|\lambda^*(\theta_k) - \lambda_{k+1}^{i}\|^2 + 4C_{K_5}\alpha_{k-Z_K}^2. \numberthis \label{eq:markov-thm-gradbias-bound}
\end{align*}
Plug (\ref{eq:markov-thm-gradbias-bound}) into (\ref{eq:markov-thm-err-decouple}), we get
\begin{align*}
\EE[J(\theta_{k+1})] - J(\theta_k) 
&\geq \sum_{i=1}^{N} \left(\frac{\alpha_k}{2}\EE\|\nabla_{\theta^i}J(\theta_k)\|^2 + \frac{\alpha_k}{2} \EE \|g_a^i(\xi_k, \omega_{k+1}^i, \lambda_{k+1}^i)\|^2 \right. \\
&\quad \left. - 8 C_\psi^2\alpha_k \EE\| \omega^*(\theta_k) - \omega_{k+1}^i\|^2 - 4 C_\psi^2\alpha_k\EE\|\lambda^*(\theta_k) - \lambda_{k+1}^{i}\|^2\right) \\
&\quad - \frac{L}{2}NC_{\theta}^2\alpha_k^2- 2NC_{K_5}\alpha_{k-Z_K}^2 - 8(\varepsilon_{sp} + C_\psi^2\varepsilon_{app}) N\alpha_k. \numberthis \label{eq:markov-thm-7}
\end{align*}

Consider the Lyapunov function 
\[\VV_k := -J(\theta_k) + \| \bar{\omega}_k - \omega^*(\theta_k)\|^2 + \|\bar{\lambda}_k - \lambda^*(\theta_k) \|^2. \numberthis \label{eq:markov-thm-8}\]

The difference between two Lyapunov functions will be
\begin{align*}
\EE[\VV_{k+1}] - \EE[\VV_k] &= \EE[J(\theta_k)] - \EE[J(\theta_{k+1})] + \EE\|\bar{\omega}_{k+1} - \omega^*(\theta_{k+1}) \|^2 - \EE\| \bar{\omega}_k - \omega^*(\theta_k) \|^2 \\
&\quad + \EE\|\bar{\lambda}_{k+1} - \lambda^*(\theta_k)\|^2 - \EE\|\bar{\lambda}_k - \lambda^*(\theta_k)\|^2 \\
&\leq \sum_{i=1}^{N} \left(-\frac{\alpha_k}{2}\|\nabla_{\theta^i}J(\theta_k)\|^2 - \frac{\alpha_k}{2} \EE \|g_a^i(\xi_k, \omega_{k+1}^i)\|^2\right)
\\
&\quad + 2NC_{K_5}\alpha_{k-Z_K} + \frac{L}{2}NC_{\theta}^2\alpha_k^2 + 8(\varepsilon_{sp} + C_\psi^2\varepsilon_{app}) N\alpha_k \\
&\quad + \underbrace{\sum_{i=1}^{N} 8 C_\psi^2\alpha_k \EE\| \omega^*(\theta_k) - \omega_{k+1}^i\|^2 + \EE\|\bar{\omega}_{k+1} - \omega^*(\theta_{k+1}) \|^2 - \EE\| \bar{\omega}_k - \omega^*(\theta_k) \|^2}_{I_5} \\
&\quad + \underbrace{\sum_{i=1}^{N} 4 C_\psi^2\alpha_k\EE\|\lambda^*(\theta_k) - \lambda_{k+1}^{i}\|^2 + \EE\|\bar{\lambda}_{k+1} - \lambda^*(\theta_{k+1})\|^2 - \EE\|\bar{\lambda}_k - \lambda^*(\theta_k) \|^2}_{I_6}. \numberthis \label{eq:markov-thm-9}
\end{align*}

The first two terms of $I_5$ can be bounded as 
\begin{align*}
&\sum_{i=1}^{N} 8 C_\psi^2\alpha_k \EE\| \omega^*(\theta_k) - \bar{\omega}_{k+1} + \bar{\omega}_{k+1} - \omega_{k+1}^i\|^2 + \EE\|\bar{\omega}_{k+1} - \omega^*(\theta_{k+1}) \|^2 \\
&= \sum_{i=1}^{N} 8 C_\psi^2\alpha_k\EE\|\bar{\omega}_{k+1} - \omega_{k+1}^i\|^2 + 8 C_\psi^2 \alpha_k\EE\|\bar{\omega}_{k+1} - \omega^*(\theta_k) \|^2 + \EE\|\bar{\omega}_{k+1} - \omega^*(\theta_{k+1}) \|^2 \\
%	&\leq \sum_{i=1}^{N}8C_\psi^2\alpha_k(\nu^{2k} \frac{\|\boldsymbol{\omega}_0\|_F^2}{\zeta^2} + \frac{4L\|\boldsymbol{\omega}_0\|_F\nu^k}{\zeta^2}\alpha_k + \frac{4L^2}{\zeta^2} \alpha_k^2) \\
%&\leq 8C_\psi^2\alpha_k(\nu^{2k} \| \bomega_0 \|_F +    \frac{16NC_\delta^2}{1-\nu}\beta_k^2 + \frac{8\sqrt{N}C_\delta\|\bomega_0\|}{1-\nu} \nu^k \beta_k) \\
%&\quad + 8 C_\psi^2 \alpha_k\EE\|\bar{\omega}_{k+1} - \omega^*(\theta_k) \|^2 + \EE\|\bar{\omega}_{k+1} - \omega^*(\theta_{k+1}) \|^2 \\
&\leq 8 C_\psi^2 \alpha_k\EE\|\bar{\omega}_{k+1} - \omega^*(\theta_k) \|^2 + \EE\|\bar{\omega}_{k+1} - \omega^*(\theta_{k+1}) \|^2 + \alpha_k M_{k_1} \\
&\leq \left(1 +  4L_{\omega}N \alpha_k + 8 C_\psi^2 \alpha_k + 
\frac{L_{\omega, 2}^{2}}{2}C_\theta^2N\sqrt{d_\theta}\alpha_k^2\right)\EE\|\bar{\omega}_{k+1} - \omega^*(\theta_k)\|^2 \\
&\quad + \left(\frac{L_{\omega,2}^2C_\theta^2N}{2}+ L_\omega^2C_\theta^2N\right) \alpha_k^2 + \frac{\alpha_k}{4}\sum_{i=1}^{N}\left\|\EE\left[g_a^i(\xi_k, \omega_{k+1}^i, \lambda_{k+1}^{i})\right]\right\|^2 + \alpha_k M_{k_1},
\numberthis \label{eq:markov-thm-10}
\end{align*}
where the equality is due to 
\[\sum_{i=1}^{N} \left\langle \omega^*(\theta_k) - \bar{\omega}_{k+1} , \bar{\omega}_{k+1} - \omega_{k+1}^i \right\rangle = \left\langle \omega^*(\theta_k) - \bar{\omega}_{k+1} , \bar{\omega}_{k+1} - \bar{\omega}_{k+1} \right\rangle = 0.\]
%\[\sum_{i=1}^{N}\|\omega^*(\theta_{k})-\omega_{k+1}^{i}\|^2 = \|\omega^*(\theta_k)-\bar{\omega}_{k+1}\|^2 + \| \bar{\omega}_{k+1}-\omega_{k+1} \]
The first inequality follows the Lemma~\ref{lemma:consensus-error}, where $M_{k_1}$ is defined as
\[M_{k_1}:= \nu^{\frac{2k}{K_c}-2}\|\bomega_0\|^2 + 16 NC_\delta^2\left(\sum_{t=0}^{k}\beta_t \nu^{\frac{k-t}{K_c}-1} \right)^2 + 8 \sqrt{N}C_\delta \nu^{\frac{k}{K_c}-1}\sum_{t=0}^{k}\beta_t \nu^{\frac{k-t}{K_c}-1}. \]

% \[M_{k_1}:= 8C_\psi^2\left(\nu^{2k} \| \bomega_0 \|_F +    \frac{16NC_\delta^2}{1-\nu}\beta_k^2 + \frac{8\sqrt{N}C_\delta\|\bomega_0\|_F}{1-\nu} \nu^k \beta_k\right). \numberthis \label{eq:def-mk1}\]
The last inequality follows (\ref{eq:critic-error-7}) in Lemma~\ref{lemma:critic-error-markov-1}.

%Recall
%%\[M_{k_1}:= 8NC_\psi^2(\nu^{2k} \frac{\|\boldsymbol{\omega}_0\|_F^2}{\zeta^2} + \frac{4L\|\boldsymbol{\omega}_0\|_F\nu^k}{\zeta^2}\alpha_k + \frac{4L^2}{\zeta^2} \alpha_k^2).\]
%\[M_{k_1}:= 8C_\psi^2(\nu^{2k} \| \bomega_0 \|_F +    \frac{16NC_\delta^2}{1-\nu}\beta_k^2 + \frac{8\sqrt{N}C_\delta\|\bomega_0\|_F}{1-\nu} \nu^k \beta_k). \numberthis \label{eq:markov-thm-11}\]

%Plug (\ref{eq:markov-thm-11}) into (\ref{eq:markov-thm-10}), we have
%\begin{align*}
%I_5 &\leq 8 C_\psi^2 \alpha_k\EE\|\bar{\omega}_{k+1} - \omega^*(\theta_k) \|^2 + \EE\|\bar{\omega}_{k+1} - \omega^*(\theta_{k+1}) \|^2 + \alpha_k M_{k_1} \\
%&\leq (1 +  4L_{\omega, 2}^{2}N \alpha_k + 8 C_\psi^2 \alpha_k + 
%\frac{L_{\omega, 2}^{2}}{2}C_\theta^2N^2\alpha_k^2)\EE\|\bar{\omega}_{k+1} - \omega^*(\theta_k)\|^2 \\
%&\quad + (\frac{L_{\omega,2}^2C_\theta^2N^2}{2}+ L_\omega^2) \alpha_k^2 + \frac{\alpha_k}{4}\sum_{i=1}^{N}\|\EE[g_a^i(\xi_k, \omega_{k+1}^i, \lambda_{k+1}^{i})]\|^2 + \alpha_k M_{k_1}, \numberthis \label{eq:markov-thm-12}
%\end{align*}
%where the second inequality follows (\ref{eq:critic-error-7}) in Lemma~\ref{lemma:critic-error-markov-1}, and $M_{k_1}$ is defined in (\ref{eq:iid-thm-7}).

Plug (\ref{eq:markov-thm-10}) into (\ref{eq:markov-thm-9}), and define $C_9:=\min\left\{c \ | \ 4L_{\omega}N \alpha_k + 8 C_\psi^2 \alpha_k + 
\frac{L_{\omega, 2}^{2}}{2}C_\theta^2N\sqrt{d_\theta}\alpha_k^2 \leq c \alpha_k\right\}$, we get 
\begin{align*}
I_5 &\leq 
(1+C_9\alpha_k)\EE\|\bar{\omega}_{k+1} - \omega^*(\theta_k)\|^2 + \left(\frac{L_{\omega,2}^2C_\theta^2N^2}{2}+ L_\omega^2C_\theta^2N\right) \alpha_k^2 \\
&\quad + \frac{\alpha_k}{4}\sum_{i=1}^{N}\|\EE[g_a^i(\xi_k, \omega_{k+1}^i, \lambda_{k+1}^{i})]\|^2 + \alpha_k M_{k_1} \\
&\leq [(1+C_9\alpha_k)(1-2\lambda_{\phi}\beta_k)-1]\EE\|\bar{\omega}_{k} - \omega^*(\theta_k)\|^2 \\
&\quad + (1+C_9\alpha_k)(C_{K_1}\beta_k\beta_{k-Z_K}+C_{K_2}\beta_k\alpha_{k-Z_K}) \\
&\quad  + \left(\frac{L_{\omega,2}^2C_\theta^2N}{2}+ L_\omega^2C_\theta^2N\right) \alpha_k^2 + \frac{\alpha_k}{4}\sum_{i=1}^{N}\left\|\EE\left[g_a^i\left(\xi_k, \omega_{k+1}^i, \lambda_{k+1}^{i}\right)\right]\right\|^2 + \alpha_k M_{k_1}, \numberthis \label{eq:markov-thm-13}
\end{align*}
where the last inequality follows (\ref{eq:critic-error-8}) in Lemma~\ref{lemma:critic-error-markov-1}.

By letting $\beta_k=\frac{C_9}{2\lambda_{\phi}}\alpha_k$, we can ensure
\[(1+C_9\alpha_k)(1-2\lambda_{\phi}\beta_k) < 1.\]
Therefore, $I_5$ can be bounded as
\begin{align*}
I_5 &\leq \frac{\alpha_k}{4}\sum_{i=1}^{N}\|\EE[g_a^i(\xi_k, \omega_{k+1}^i, \lambda_{k+1}^{i})]\|^2 + \alpha_k M_{k_1} + \left(\frac{L_{\omega,2}^2C_\theta^2N^2}{2}+ L_\omega^2C_\theta^2N\right) \alpha_k^2 \\
&\quad + (1+C_9\alpha_k)(C_{K_1}\beta_k\beta_{k-Z_K}+C_{K_2}\beta_k\alpha_{k-Z_K}). \numberthis \label{eq:markov-thm-14}
\end{align*}

By applying Lemma~\ref{lemma:lambda-error-markov-1} and following the similar procedure, we can bound $I_6$ as
\begin{align*}
I_6 &\leq \frac{\alpha_k}{4}\sum_{i=1}^{N}\|\EE[g_a^i(\xi_k, \omega_{k+1}^i, \lambda_{k+1}^{i})]\|^2 + \alpha_k M_{k_2} + \left(\frac{L_{\lambda,2}^2C_\theta^2N^2}{2}+ L_\lambda^2C_\theta^2N\right) \alpha_k^2 \\
&\quad + (1+C_{10}\alpha_k)(C_{K_3}\eta_k\eta_{k-Z_K}+C_{K_4}\eta_k\alpha_{k-Z_K}). \numberthis \label{eq:markov-thm-15}
\end{align*}
with $\eta_k = \frac{C_{10}}{2\lambda_{\varphi}}\alpha_k$.
$C_{10}$ and $M_{k_2}$ are defined as
\begin{align*}
C_{10}&:=\min\left\{c \ | \ 4{L_{\lambda}} N \alpha_k + 4 C_\psi^2 \alpha_k + \frac{L_{\lambda, 2}^2}{2}C_\delta^2 N\sqrt{d_\theta} \alpha_k^2 \leq c \alpha_k\right\}, \\
M_{k_2}&:= \nu^{\frac{2k}{K_c}-2}\|\blambda_0\|^2 + 16 NC_\lambda^2\left(\sum_{t=0}^{k}\eta_t \nu^{\frac{k-t}{K_c}-1} \right)^2 + 8 \sqrt{N}C_\lambda \nu^{\frac{k}{K_c}-1} \sum_{t=0}^{k}\eta_t \nu^{\frac{k-t}{K_c}-1}. \numberthis \label{eq:def-c10-mk2}
\end{align*}

Plug (\ref{eq:markov-thm-14}) and (\ref{eq:markov-thm-15}) into (\ref{eq:markov-thm-9}), we have
\begin{align*}
\EE[\VV_{k+1}] - \EE[\VV_k] 
&\leq \sum_{i=1}^{N} -\frac{\alpha_k}{2}\|\nabla_{\theta^i}J(\theta_k)\|^2 + (M_{k_1} + M_{k_2}) \alpha_k \\
%&\quad + (1+C_9\alpha_k)C_\delta^2\beta_k^2 + (1+C_{10}\alpha_k)C_\lambda^2\eta_k^2 +(\frac{L}{2}NC_{\theta}^2 + C_{11})\alpha_k^2 \\
&\quad + (1+C_9\alpha_k)(C_{K_1}\beta_k\beta_{k-Z_K}+C_{K_2}\beta_k\alpha_{k-Z_K}) \\
&\quad + (1+C_{10}\alpha_k)(C_{K_3}\eta_k\eta_{k-Z_K}+C_{K_4}\eta_k\alpha_{k-Z_K})  \\
&\quad + \left(\frac{L}{2}NC_{\theta}^2 + C_{11}\right)\alpha_k^2 + 8(\varepsilon_{sp}+C_\psi^2\varepsilon_{app}N)\alpha_k,
%&= \sum_{i=1}^{N}(-\frac{\alpha_k}{2}\|\nabla_{\theta^i}J(\theta_k)\|^2) + (M_{k_1} + M_{k_2}) \alpha_k + 8(\varepsilon_{sp}+C_\psi^2\varepsilon_{app}N)\alpha_k \\
%&\quad + (1+C_9\alpha_k)C_\delta^2\beta_k^2 + (1+C_{10}\alpha_k)C_\lambda^2\eta_k^2 +(\frac{L}{2}NC_{\theta}^2 + C_{11})\alpha_k^2,
\numberthis \label{eq:markov-thm-16}
\end{align*} 
where $C_{11}:=C_\theta^2N(\frac{L_{\omega,2}^2+L_{\lambda,2}}{2} + L_\omega^2 + L_\lambda^2$).

By letting $\alpha_k = \frac{\bar{\alpha}}{\sqrt{K}}$ for some positive constant $\bar{\alpha}$, and recall $\beta_k=\frac{C_9}{2\lambda_{\phi}}\alpha_k, \eta_k = \frac{C_{10}}{2\lambda_{\varphi}}\alpha_k$, we can telescope (\ref{eq:markov-thm-16}) as
\begin{align*}
\frac{1}{K}\sum_{k=0}^{K}\sum_{i=1}^{N} \EE\|\nabla_{\theta^i}J(\theta_k)\|^2 &\leq \frac{2\EE[\VV_0]}{K\alpha_k} + 16(\varepsilon_{sp}+C_\psi^2\varepsilon_{app}N) + \frac{2}{K}\sum_{k=0}^{K}(M_{k_1} + M_{k_2}) \\
&\quad + (2+2C_9\alpha_k)(C_{K_1}\frac{\beta_k}{\alpha_k}\beta_{k-Z_K} + C_{K_2}\frac{\beta_k}{\alpha_k}\alpha_{k-Z_K}) \\
&\quad + (2+2C_{10}\alpha_k)(C_{K_3}\frac{\eta_k}{\alpha_k}\eta_{k-Z_K} + C_{K_4}\frac{\eta_k}{\alpha_k}\alpha_{k-Z_K}) \\
&\quad + \left(LNC_{\theta}^2 + 2C_{11}\right)\alpha_k. \numberthis \label{eq:markov-thm-17}
\end{align*}

The summation of $M_{k_1}$ can be bounded as
\begin{align*}
    \sum_{k=0}^{K} M_{k_1} &= \sum_{k=0}^{K} \left( \nu^{\frac{2k}{K_c}-2}\|\bomega_0\|^2 + 16 NC_\delta^2\left(\sum_{t=0}^{k}\beta_t \nu^{\frac{k-t}{K_c}-1} \right)^2 + 8 \sqrt{N}C_\delta \nu^{\frac{k}{K_c}-1}\sum_{t=0}^{k}\beta_t \nu^{\frac{k-t}{K_c}-1} \right) \\
    &= \sum_{k=0}^{K} \left( \nu^{\frac{2k}{K_c}-2}\|\bomega_0\|^2 + 16 NC_\delta^2 \beta_k^2\left(\sum_{t=0}^{k} \nu^{\frac{k-t}{K_c}-1} \right)^2 + 8 \sqrt{N}C_\delta \nu^{\frac{k}{K_c}-1}\beta_k\sum_{t=0}^{k}\nu^{\frac{k-t}{K_c}-1} \right) \\
    &\overset{(i)}{\leq} \sum_{k=0}^{K} \left( \nu^{\frac{2k}{K_c}-2}\|\bomega_0\|^2 + 16 NC_\delta^2 \beta_k^2\frac{K_c^2}{\nu^2(1-\nu)^2} + 8 \sqrt{N}C_\delta \nu^{\frac{k}{K_c}-1}\beta_k\frac{K_c}{\nu(1-\nu)} \right) \\
    &\overset{(ii)}{\leq} \frac{\|\bomega_0\|^2}{\nu^2(1-\nu^{2/K_c})^2} + \frac{16NC_\delta^2\beta_k^2K_c^2 K}{\nu^2(1-\nu)^2} + 8\sqrt{N}C_\delta\beta_k\frac{K_c}{(1-\nu^{K_c})\nu(1-\nu)} \\
    &= \cO(\beta_k^2K_c^2) = \cO(\sqrt{K}), \numberthis \label{eq:markov-thm-18}
\end{align*}
where the second equality is according to the step size choice. $(i)$ is due to 
\[\sum_{t=0}^{k}\nu^{\frac{k-t}{K_c}-1} \leq K_c \sum_{z=0}^{\lceil \frac{k}{K_c}\rceil} \nu^{z-1} \leq K_c \frac{1}{\nu(1-\nu)} .\]
$(ii)$ is due to $\sum_{k=0}^{K} \nu^{\frac{k}{K_c}-1} = \frac{1}{\nu(1-\nu^{1/K_c})}$. The last equality uses $K_c=\cO(K^{1/4})$. By following similar arguments, we can show that $\sum_{k=0}^{K}M_{k_2}=\cO(\sqrt{K})$. Therefore, the third term in (\ref{eq:markov-thm-17}) is of order $\cO(\frac{1}{\sqrt{K}})$.

Finally, by 
noticing $C_{K_1}=\cO(\log\frac{1}{\alpha_k}), C_{K_2}=\cO(\log^2\frac{1}{\alpha_k}), C_{K_3}=\cO(\log\frac{1}{\alpha_k}), C_{K_4}=\cO(\log^2\frac{1}{\alpha_k})$, we obtain the desired iteration complexity of $\tilde \cO(\frac{1}{\sqrt{K}})$, or equivalently, the sample complexity of $\tilde \cO(\varepsilon^{-2})$.

\subsection{Proof of Theorem~\ref{thm:main-theorem-markov-noisy}}\label{appendix:proof-thm-noisy}

Define the update of actor $i$ using the noisy reward as 
\begin{align*}
g_a^i(\xi_k, \omega_{k+1}^i) := \tilde{r}_{k, K_r}^i(s_k, a_k) + \gamma \phi(s')^T\omega_{k+1}^{i}  - \phi(s)^T \omega_{k+1}^{i}. \numberthis \label{eq:noi-thm-1}
\end{align*}

Following the derivation of (\ref{eq:markov-thm-err-decouple}), we have
\begin{align*}
\EE[J(\theta_{k+1}] - J(\theta_k) &\geq \sum_{i=1}^{N} \left[\frac{\alpha_k}{2} \| \nabla_{\theta^i} J(\theta_k)\|^2 + \frac{\alpha_k}{2}\|\EE[g_a^i(\xi_k, \omega_{k+1}^i)]\|^2 \right. \\
&\quad \left. - \frac{\alpha_k}{2}\|\nabla_{\theta^i}J(\theta_k) - \EE[g_a^i(\xi_k, \omega_{k+1}^i)] \|^2\right] - \frac{L}{2} N C_{\theta}^2 \alpha_k^2. \numberthis \label{eq:noi-thm-2}
\end{align*}
Similarly to the proof of Theorem~ \ref{thm:main-theorem-markovian}, the gradient bias term can be decomposed as as
\begin{align*}
\|\nabla_{\theta^i}J(\theta_k) - \EE[g_a^i(\xi_k, \omega_{k+1}^i)] \|^2 
&\leq 4 \underbrace{\| \nabla_{\theta^i} J(\theta_k) - \EE[\delta(\xi_k, \theta_k) \psi_{\theta_k^i}]\|^2}_{I_1} \\
&\quad + 4 \underbrace{\| \EE[(\delta(\xi_k, \theta_k) - \tilde{\delta}(\xi_k, \omega^*(\theta_k))) \psi_{\theta_k^i}] \|^2}_{I_2} \\
&\quad + 4 \underbrace{\| \EE[(\tilde{\delta}(\xi_k, \omega^*(\theta_k)) -\tilde{\delta}(\xi_k, \omega_{k+1}^{i})) \psi_{\theta_k^i}]\|^2}_{I_3} \\
&\quad + 4 \underbrace{\| \EE[(\bar{r}_k(s_k, a_k) - \tilde{r}_{k,K_r}(s_k, a_k)) \psi_{\theta_k^i}] \|^2}_{I_4} \numberthis \label{eq:noi-thm-3}
\end{align*}
$I_1$, $I_2$, $I_3$ can be bounded following the derivation of (\ref{eq:markov-thm-gradbias-bound}), (\ref{eq:markov-thm-i2-bound}), and (\ref{eq:markov-thm-i3-bound}), respectively. Plug these bounds into (\ref{eq:noi-thm-2}), we have
\begin{align*}
\EE[J(\theta_{k+1})] - J(\theta_k) 
&\geq \sum_{i=1}^{N} \left(\frac{\alpha_k}{2}\EE\|\nabla_{\theta^i}J(\theta_k)\|^2 + \frac{\alpha_k}{2} \EE \|g_a^i(\xi_k, \omega_{k+1}^i)\|^2 - 8 C_\psi^2\alpha_k \EE\| \omega^*(\theta_k) - \omega_{k+1}^i\|^2\right) \\
&\quad - \sum_{i=1}^{N} \frac{\alpha_k}{2} C_{\psi}^2 \|\bar{r}_k(s_k,a_k) - \tilde{r}_{k, K_r}^i(s_k, a_k)\|^2 - \frac{L}{2}NC_{\theta}^2\alpha_k^2 \\
&\quad - 2NC_{K_5}\alpha_{k-Z_K}^2 - 8(\varepsilon_{sp} + C_\psi^2\varepsilon_{app}^r) N\alpha_k. \numberthis \label{eq:noi-thm-4}
\end{align*}
Define $\tilde{r}_{k, K_r} := [r_{k, K_r}^1, \cdots, r_{k, K_r}^N]^T$. The reward bias can be bounded as
\begin{align*}
\sum_{i=1}^{N} \|\bar{r}_k(s_k,a_k) - \tilde{r}_{k, K_r}^i(s_k, a_k)\|^2 &= \|Q\tilde{r}_{k, K_r}\|^2 \\
&= \|QW^{K_r}\tilde{r}_{k, 0}(s_k, a_k)\|^2 \\
&\leq \nu^{2K_r}\|\tilde{r}_{k, 0}(s_k, a_k)\|^2 \\
&= \nu^{2K_r} \sum_{i=1}^{N}\left(\|\tilde{r}_{k,0}^i(s_k, a_k) - \bar{r}_{k}(s_k,a_k)\|^2 + \|\bar{r}_k(s_k, a_k)\|^2\right) \\
&\leq \nu^{2K_r}N(\sigma^2 + r_{\max}), \numberthis \label{eq:noi-thm-5}
\end{align*}
where $\sigma^2$ is the variance of the reward noise. Let $K_r=\frac{1}{2}\log_{\nu}\alpha_k$ and define $C_{15}:=\sigma^2 + r_{\max}^2$. Plug (\ref{eq:noi-thm-5}) back to (\ref{eq:noi-thm-4}), we have 
\begin{align*}
\EE[J(\theta_{k+1})] - J(\theta_k) 
&\geq \sum_{i=1}^{N} \left(\frac{\alpha_k}{2}\EE\|\nabla_{\theta^i}J(\theta_k)\|^2 + \frac{\alpha_k}{2} \EE \|g_a^i(\xi_k, \omega_{k+1}^i)\|^2 - 8 C_\psi^2\alpha_k \EE\| \omega^*(\theta_k) - \omega_{k+1}^i\|^2\right) \\
&\quad + \frac{N}{2}(C_{15}+C_\theta^2L)\alpha_k^2
- 2NC_{K_5}\alpha_{k-Z_K}^2 - 8(\varepsilon_{sp} + C_\psi^2\varepsilon_{app}^r) N\alpha_k.
\end{align*}
Consider the Lyapunov function
\[\VV_k:=-J(\theta_k) + \|\bar{\omega}_k - \omega^*(\theta_k)\|^2. \]
The difference between two Lyapunov functions is
\begin{align*}
\EE[\VV_{k+1}] - \EE[\VV_k] &\leq \sum_{i=1}^{N} \left(-\frac{\alpha_k}{2}\|\nabla_{\theta^i}J(\theta_k)\|^2 - \frac{\alpha_k}{2} \EE \|g_a^i(\xi_k, \omega_{k+1}^i)\|^2\right)
\\
&\quad + \frac{N}{2}C_{16}\alpha_k^2
- 2NC_{K_5}\alpha_{k-Z_K}^2 - 8(\varepsilon_{sp} + C_\psi^2\varepsilon_{app}^r) N\alpha_k \\
&\quad + \underbrace{\sum_{i=1}^{N} 8 C_\psi^2\alpha_k \EE\| \omega^*(\theta_k) - \omega_{k+1}^i\|^2 + \EE\|\bar{\omega}_{k+1} - \omega^*(\theta_{k+1}) \|^2 - \EE\| \bar{\omega}_k - \omega^*(\theta_k) \|^2}_{I_5}. \\
\end{align*} 
$I_5$ can be bounded by following the derivation of (\ref{eq:markov-thm-14}). Thus, we have
\begin{align*}
&\EE[\VV_{k+1}] - \EE[\VV_k] \\
&\leq \sum_{i=1}^{N} -\frac{\alpha_k}{2}\|\nabla_{\theta^i}J(\theta_k)\|^2 + \frac{N}{2}C_{16}\alpha_k^2
- 2NC_{K_5}\alpha_{k-Z_K}^2 - 8(\varepsilon_{sp} + C_\psi^2\varepsilon_{app}^r) N\alpha_k \\
&\quad + (1+C_9\alpha_k)(C_{K_1}\beta_k\beta_{k-Z_K}+C_{K_2}\beta_k\alpha_{k-Z_K}) + M_{k_1}\alpha_k,\numberthis \label{eq:noi-thm-6}
\end{align*}
where $C_{16}:=C_{15}+C_\theta^2L+ \frac{L_{\omega,2}^2C_\theta^2N^2}{2} + L_\omega^2$.

Telescoping (\ref{eq:noi-thm-6}), we have 
\begin{align*}
\frac{1}{K}\sum_{k=0}^{K}\sum_{i=1}^{N} \EE\|\nabla_{\theta^i}J(\theta_k)\|^2 &\leq \frac{2\EE[\VV_0]}{K\alpha_k} +  16(\varepsilon_{sp}+C_\psi^2\varepsilon_{app}^rN) + \frac{2}{K}\sum_{k=0}^{K}M_{k_1} + C_{16}\alpha_k\\
&\quad + (1+C_9\alpha_k)\left(C_{K_1}\frac{\beta_k}{\alpha_k}\beta_{k-Z_K} + C_{K_2}\frac{\beta_k}{\alpha_k}\alpha_{k-Z_K}\right).
\end{align*} 

The term $\frac{2}{K}\sum_{k=0}^{K}M_{k_1}$ has been bounded in (\ref{eq:markov-thm-18}). Let $\alpha_k = \frac{\bar{\alpha}}{\sqrt{K}}$ for some positive constant $\bar{\alpha}$, $\beta_k=\frac{C_9}{2\lambda_\phi}\alpha_k$ will yield the desired rate.

\section{Natural Actor-Critic variant and its convergence} \label{appendix:nac-convergence}
In this section, we propose a natural Actor-Critic variant of Algorithm~\ref{algorithm:dec-ac-re}, where the approach of calculating the natural policy graident under the decentralized setting is mainly inspired by~\citep{chen2021sample}. We show that the gradient norm square of such an algorithm will converge with the optimal sample complexity of $\widetilde{\cO}(\varepsilon^{-3})$. Moreover, the algorithm will converge to the \emph{global optimum} with the sample complexity of $\widetilde{\cO}(\varepsilon^{-6})$. 
In the rest of this section, we first explain the update of the algorithm, and then prove its convergence.

\subsection{Decentralized natural Actor-Critic}

The natural policy gradient (NPG) algorithm~\citep{kakade2002natural} can be viewed as a preconditioned policy gradient algorithm, which updates as follow:
\[\theta_{k+1} = \theta_k - \alpha_k F(\theta_k)^{\dagger}\nabla J(\theta_k), \numberthis \label{eq:npg-update} \]
where $F(\theta):=\EE_{s\sim d_{\pit}, a \sim \pit}\left[\psi_{\theta}(s,a)\psi_{\theta}(s,a)^T \right]$ is the Fisher information matrix (FIM). The natural Actor-Critic (NAC) uses the critic variable to estimate the gradient. The main challenge for implementing NAC lies in the estimation of the matrix-vector product $F(\theta_k)^{\dagger}\nabla J(\theta_k)$, especially under the decentralized setting. The work~\citep{chen2021sample} proposes to solve the following subproblem in order to estimate the product in a decentralized way:
\[h(\theta_k) = \argmin_{h} f_{\theta_k}(h) := \frac{1}{2}h^TF(\theta_k)h - \nabla J(\theta_k)^Th. \numberthis \label{eq:nac-subproblem}\]
Such a problem can be solved by using (stochastic) gradient descent, where the gradient is calculated by $F(\theta_k)h - \nabla J(\theta_k)$. For the centralized setting, the gradient w.r.t. each agent can be approximated as $\frac{1}{N_a}\sum_{n=1}^{N_a}\psi_{\theta_k}^i(s_n, a_n^i)\psi_{\theta_k}(s_n, a_n)^Th - g_a^i(\xi_n, \omega_{k+1}, \lambda_{k+1})$. However, when considering the decentralized setting, the term $\bar{z}_{n}:=\psi_{\theta_k}(s_n, a_n)^Th = \sum_{i=1}^{N}\psi_{\theta_k}^i(s_n, a_n)^Th^i$ is not accessible for each agent. To approximate this value under the decentralized setting, agents compute $z_{n, 0}^i := \psi_{\theta_k}^i(s_n, a_n)^Th^i$ locally and then perform the following communication step for $K_z$ steps:
\[z_{n, k'+1}^i = \sum_{j=1}^{N} W^{ij} z_{n, k'}^i , \ \forall n \in [N_a], \ k'=0,\cdots,K_z-1. \numberthis \label{eq:nac-product-estimation} \]
As we will see, the value $Nz_{n, k'}^i$ converges to $\bar{z}_{n}$ linearly. Thus, the gradient of the subproblem (\ref{eq:nac-subproblem}) for agent $i$ can be approximated as:
\[\widetilde{\nabla} f_{\theta_k}^i(h_{k, k'}) := \frac{N}{N_a} \sum_{n=1}^{N_a}\psi_{\theta_k}^{i}(s_n, a_n^i)z_{n, K_z}^i - g_a^i(\xi_n, \omega_{k+1}, \lambda_{k+1}). \numberthis \label{eq:nac-direction-gradient} \]
Then, each agent $i$ performs the following update for $K_a$ steps to estimate the natural policy gradient direction:
\[h_{k, k'+1}^i = \Pi_{C_h}(h_{k, k'}^i - \varrho \widetilde{\nabla}f_{\theta_k}^{i}(h_{k, k'})), \numberthis \label{eq:nac-direction-update} \]
where $\varrho$ is a positive constant step size. Since the norm of optimal direction is bounded by $C_h:=\lambda_{\max}(F(\theta)^{-1})C_\theta$, we project the vector into a ball of norm $C_h$ for each update. %\qijun{deal with the projection} 
Finally, we perform the approximate natural policy gradient step as:
\[\theta_{k+1}^i = \theta_{k}^{i} - \alpha_k h_{k, K_a}^i. \numberthis \label{eq:nac-update} \]
% \qijun{To simplify the subscript}

\begin{algorithm}[t]
	\null
	\caption{Decentralized single-timescale NAC}
	\label{algorithm:dec-nac-re}
	\small
	\begin{algorithmic}[1]
		\STATE {\bfseries Initialize:} Actor parameter $\theta_0$, critic parameter $\omega_0$, reward estimator parameter $\lambda_0$, initial state $s_0$, natural policy gradient estimation $h_{k, 0}$.
		\FOR{$k=0,\cdots,K-1$}
		
		\STATE \textbf{{Option 1: i.i.d. sampling:}}
		\STATE $s_k \sim \mu_{\theta_k}(\cdot), a_k \sim \pi_{\theta_k}(\cdot|s_k), s_{k+1} \sim \cP(\cdot | s_k, a_k)$.
		
		\STATE \textbf{{Option 2: Markovian sampling:}}
		\STATE $a_k \sim \pi_{\theta_k}(\cdot|s_k), s_{k+1} \sim \cP(\cdot | s_k, a_k)$.
		
		\STATE
		
		\STATE \textbf{{Periodical consensus:}} Compute $\tilde{\omega}_k^i$ and $\tilde{\lambda}_k^i$ by (\ref{eq:critic-consensus}) and (\ref{eq:reward-estimator-consensus}).
		
		\STATE
		
		\FOR{$i=0, \cdots, N$ \textbf{in parallel}}
		
		\STATE {\textbf{Reward estimator update:}} Update $\lambda_{k+1}^i$ by (\ref{eq:reward-estimator-update}).
		\STATE {\textbf{Critic update:}} Update $\omega_{k+1}^{i}$ by (\ref{eq:critic-update}).
		\STATE {\textbf{Actor update:}} %Update $\theta_{k+1}^{i}$ by (\ref{eq:actor-update}).
		
		\begin{ALC@g}
			\STATE Collect $N_a$ transition samples based on Markovian/i.i.d sampling.
			%			\STATE Initialize  as $h_{k, 0}=\boldsymbol{0}$.
			\FOR{$k'=1, \cdots, K_a$}
			\STATE Estimate $\bar{z}_{k', n}, \ \forall n \in [N_a]$ using (\ref{eq:nac-product-estimation}).
			%				\STATE Calculate the gradient $\nabla f_{k, k'}(h_{k, k'})$ by \eqref{}.
			\STATE Update $h_{k, k'+1}$ by (\ref{eq:nac-direction-update}).
			\ENDFOR
			\STATE Update $\theta_{k+1}^{i}$ by (\ref{eq:nac-update}).
		\end{ALC@g}
		\ENDFOR
		\ENDFOR
	\end{algorithmic}
\end{algorithm}

\subsection{Convergence of natural Actor-Critic}
In this section, we establish the sample complexity of Algorithm~\ref{algorithm:dec-nac-re}. We first introduce an additional assumption.

\begin{assumption}{(invertible FIM)}\label{assump:fim-invertible}
	There exists a positive constant $\lambda_F$ such that for all policy $\theta$,  $\lambda_{\min}(F(\theta)) \geq \lambda_F$.
\end{assumption}
Assumption~\ref{assump:fim-invertible} ensures that $F(\theta)$ is positive definite so that the problem~(\ref{eq:nac-subproblem}) is strongly convex for all policy. Such an assumption is also adopted by~\citep{chen2021sample, xu2021doubly, liu2020improved}. 

We now show the sample complexity of the Algroithm~\ref{algorithm:dec-nac-re} in terms of gradient norm and the global optimality gap. To keep the analysis concise, we will consider the i.i.d. sampling scheme where we can directly sample transition tuples $(s, a, s')$ from the stationary distribution $\mu_{\pit}$. Extending the analysis to the Markovian sampling scheme essentially follows the similar technique as in AC's analysis, which introduces an additional $\cO(\log(\varepsilon^{-1}))$ error terms caused by Markov chain mixing, and an error of order $\cO(\frac{1}{1-\gamma})$ due to the mismatch between $\mu_{\pit}$ and $d_{\pit}$.

\begin{theorem} \label{thm:nac-convergence}
	Suppose Assumptions \ref{assump:boundedness}-\ref{assump:fim-invertible} hold. Consider the update of Algorithm~\ref{algorithm:dec-nac-re} under the i.i.d. sampling. 
	Let $\alpha_k = \frac{\bar{\alpha}}{\sqrt{K}}$ for some positive constant $\bar{\alpha}$, $\beta_k=\frac{C_9}{2\lambda_{\phi}}\alpha_k$, $\varrho \leq \frac{1}{2C_\psi^2}$, $N_a=\cO(\sqrt{K})$, $K_a = \cO(\log(K^{1/2})), K_c=\cO(\log(K^{1/4}))$. Then, the following hold
	\begin{align*}
	\frac{1}{K} \sum_{k=1}^{K} \sum_{i=1}^{N} \EE \left[\| \nabla_{\theta^i} {J}(\theta_k) \|^2\right] \leq \mathcal{O}\left(\frac{1}{\sqrt{K}}\right) + \cO(\varepsilon_{app}+\varepsilon_{sp}) \numberthis \label{eq:nac-convergence-gradient}\\
	\frac{1}{K}\sum_{k=0}^{K}J(\theta^*) - J(\theta_k) \leq \cO\left(\frac{1}{K^{1/4}}\right) + \cO(\varepsilon_{app} + \varepsilon_{sp} + \varepsilon_{actor}). \numberthis \label{eq:nac-convergence-optgap}
	\end{align*}
\end{theorem}
The error $\varepsilon_{app}$ and $\varepsilon_{sp}$ are defined in (\ref{eq:approximation-error-def}) and (\ref{eq:sp-error-def}), respectively. The error $\varepsilon_{actor}$ is referred as "compatible function approximation error", which is defined as:
\[ \varepsilon_{actor}:= \max_{\theta} \min_{d} \EE_{s \sim d_{\pit}, a \sim \pit} [(\psi_{\theta}(s,a)^Td - A_{\pit}(s,a))^2].\label{eq:compatible-approx-error} \]
Such an error captures the expressivity of the policy parameterization class: it measures the error of approximating $A_{\pit}(s,a)$ using $\psi_{\theta}(s,a)$ as feature. The error becomes 0 when using the softmax-tabular parameterization; see more discussions in Section 6 of~\citep{agarwal2019theory}.

Based on Theorem~\ref{thm:nac-convergence},  Algorithm~\ref{algorithm:dec-nac-re} needs $K=\cO(\varepsilon^{-2})$ iterations to achieve $\varepsilon$-error for gradient norm square, and thus attains the sample complexity of $KN_aK_a=\widetilde{\cO}(\varepsilon^{-3})$, which matches the best existing sample complexity of NAC~\citep{xu2020improving, chen2021sample}. In terms of the global optimality gap, the algorithm requires $K=\cO(\varepsilon^{-4})$ iterations to achieve $\varepsilon$-error, and thus has the sample complexity of $KN_aK_a = \widetilde{\cO}(\varepsilon^{-6})$. Such a sample complexity is  worse than the best existing sample complexity of $\widetilde \cO(\varepsilon^{-3})$~\citep{xu2020improving, chen2021sample}.

We now explain the gap for the sub-optimal sample complexity. Mimicking  the analysis of \citep{chen2021sample} allows to establish the following inequality:
\begin{align*}
\frac{1}{K}\sum_{k=0}^{K} J\left(\theta^*\right) - \EE[J(\theta_k)] &\leq  \cO\left(\frac{1}{K}\sum_{k=1}^{K}\sum_{i=1}^{N}\EE[\|\nabla_{\theta^i} J(\theta_k)\|^2] \right) \\ 
&\quad + \cO\left(\frac{1}{K}\sum_{k=1}^{K}\sum_{i=1}^{N}\EE\|\omega_k^i - \omega^*(\theta_k)\|\right) + \cO\left(\frac{1}{K\alpha_k}\right).
\numberthis \label{eq:nac-convergence-explain}
\end{align*}
While our analysis can obtain the iteration complexity of $\cO(\frac{1}{\sqrt{K}})$ for $\|\nabla J(\theta_k)\|^2$, we can only achieve $\cO(\frac{1}{K^{1/4}})$ iteration complexity for critic's error $\|\omega_k - \omega^*(\theta_k)\|$. This is because our algorithm uses single-timescale update, where the critic's error inevitably converges slower than that of double-loop based algorithms which have $\cO(\frac{1}{\sqrt{K}})$ complexity for the critic's error at each iteration. Therefore, the sample complexity in terms of global optimality gap of our single-timescale NAC is dominated by this critic's error term, resulting in the final complexity of $\widetilde \cO(\varepsilon^{-6})$. Nevertheness, the bound (\ref{eq:nac-convergence-explain}) is not necessarily tight. We leave the research on the tight bound of single-timescale NAC as a future work.

%Based on (\ref{eq:nac-convergence-explain}), in order for the NAC algorithm to reach a global optimum rather than a stationary point, the convergence rate of the critic variables is essential. In this sense, one approach is to apply the double-loop/mini-batch technique~\citep{chen2021sample} to attain a better critic for each iteration. 

%it may be more promising to use the double-loop/mini-batch technique for the critic variables' update in order to obtain more accurate value estimation. Nevertheless, we remark that this sample complexity result is based on a straightforward application of the analysis of \citep{chen2021sample}, which is designed for double-loop algorithm. Therefore, such a proof technique may not be the tightest one for our single-timescale NAC. We leave the research on the improvement of such highly sub-optimal results of single-timescale NAC as a future work. 

\subsection{Proof of Theorem~\ref{thm:nac-convergence}}
By Lemma~\ref{lemma:smooth-objective}, we have
\begin{align*}
\EE[J(\theta_{k+1})] - J(\theta_k) &\geq \EE \langle \nabla_{\theta}J(\theta_k), \theta_{k+1} - \theta_{k} \rangle - \frac{L}{2}  \|\theta_{k+1} - \theta_k\|^2 \\
&\overset{(i)}{\geq} \alpha_k \EE \langle \nabla_{\theta}J(\theta_k), h_{k} \rangle - \frac{L}{2}NC_h^2\alpha_k^2 \\
&=\alpha_k \EE\langle\nabla_{\theta}J(\theta_k) , F(\theta_k)^{-1}g_a(\xi_k, \omega_{k+1}, \lambda_{k+1}) \rangle\\
&\quad + \alpha_k \EE\langle \nabla_{\theta}J(\theta_k), h_k - F(\theta_k)^{-1}g_a(\xi_k, \omega_{k+1}, \lambda_{k+1})\rangle - \frac{L}{2}NC_h^2\alpha_k^2 \\
&\overset{(ii)}{=} \alpha_k \EE\langle F(\theta_k)^{-1/2} \nabla_{\theta}J(\theta_k) , F(\theta_k)^{-1/2}g_a(\xi_k, \omega_{k+1}, \lambda_{k+1}) \rangle\\
&\quad + \alpha_k \EE\langle \nabla_{\theta}J(\theta_k), h_k - F(\theta_k)^{-1}g_a(\xi_k, \omega_{k+1}, \lambda_{k+1})\rangle - \frac{L}{2}NC_h^2\alpha_k^2 \\
&=
\frac{\alpha_k}{2}\|F(\theta_k)^{-1/2}\nabla_{\theta}J(\theta_k)\|^2 + \frac{\alpha_k}{2}\|F(\theta_k)^{-1/2}\EE[g_a(\xi_k, \omega_{k+1}, \lambda_{k+1})]\|^2 \\
&\quad-\frac{\alpha_k}{2} \|F(\theta_k)^{-1/2}\nabla_{\theta}J(\theta_k) - F(\theta_k)^{-1/2}\EE[g_a(\xi_k, \omega_{k+1}, \lambda_{k+1})]\|^2 \\
&\quad + \alpha_k\EE\langle \nabla_{\theta}J(\theta_k), h_k - F(\theta_k)^{-1}g_a(\xi_k, \omega_{k+1}, \lambda_{k+1}) \rangle - \frac{L}{2}NC_h^2 \alpha_k^2\\
&\overset{(iii)}{\geq} \frac{\alpha_k}{4}C_{\psi}^{-2}\|\nabla_{\theta}J(\theta_k)\|^2 + \frac{\alpha_k}{2}\lambda_{F}^{-1}\|\EE[g_a(\xi_k, \omega_{k+1}, \lambda_{k+1})]\|^2 \\
&\quad - \frac{\alpha_k}{2}\lambda_F^{-1}\underbrace{\|\nabla_{\theta}J(\theta_k) - \EE[g_a(\xi_k, \omega_{k+1}, \lambda_{k+1})]\|^2}_{I_1} \\
&\quad - 4\alpha_k C_{\psi}^2\underbrace{\|\EE[h_k] - F(\theta_k)^{-1}\EE[g_a(\xi_k, \omega_{k+1}, \lambda_{k+1})] \|^2}_{I_2} - \frac{L}{2}NC_h^2 \alpha_k^2, \numberthis \label{eq:nac-iid-thm-0}
\end{align*}
where $(i)$ is due to $\|\theta_{k+1}^i - \theta_k^i\| \leq C_h := \lambda_FC_\theta$. Note that we use $h_{k}^{i}$ to represent $h_{k, K_a}^{i}$ for simplifying the notation. $(ii)$ uses decomposition of positive definite (PD) matrix. Specifically, let $A$ be PD matrix, then by eigenvalue decomposition, $A=V\Lambda V^{T}$ for some orthonormal matrix $V$. Define $A^{-1/2}:= V\Lambda^{-1/2}V^T$, then $\langle x, A^{-1}y \rangle = \langle A^{-1/2}x, A^{-1/2}y\rangle$ for any $x$ and $y$. $(iii)$ uses $C_{\psi}^{-2} \leq \lambda(F(\theta)^{-1}) \leq \lambda_{F}^{-1}$ and Young's inequality.

$I_1$ represents the error of gradient bias, which we have bounded when analyzing the error of AC. By (\ref{eq:markov-thm-gradbias-bound}), we have
\begin{align*}
I_1 \leq \sum_{i=1}^{N} \left[16(\varepsilon_{sp}+C_\psi^2\varepsilon_{app}) + 16 C_\psi^2 \| \omega^*(\theta_k) - \omega_{k+1}^i\|^2 + 8C_\psi^2\|\lambda^*(\theta_k) - \lambda_{k+1}^{i}\|^2\right]. \numberthis \label{eq:nac-iid-thm-1}
\end{align*}
To bound $I_2$, we need to bound the error of $h_{k,k'}$. We start with the gradient bias when estimating $h_{k, k'}$. Define $\overline{\nabla}f_{k, k'}(h_{k, k'}):=\nabla F(\theta_k)h_{k, k'} - \EE[g_a(\xi_k, \omega_{k+1}^{i},\lambda_{k+1}^{i})]$, then it is easy to see that $\overline{\nabla}f_{k, k'}(h_{k, k'})$ is the unbiased gradient of the following problem
\[\frac{1}{2}h_{k, k'}^T\nabla F(\theta_k) h_{k, k'} - \EE[g_a(\xi_k, \omega_{k+1}^{i}, \lambda_{k+1}^{i})]^T h_{k, k'} .\]

Define the following notation for the ease of expression:
\begin{align*}
\widehat{\nabla}f_{k, k'}^{i}(h_{k, k'}) &:= \frac{1}{N_a}\sum_{n=1}^{N_a}\psi_{\theta_{k}^{i}}(s_n, a_n^i)\psi_{\theta_k}(s_n, a_n)^T h_{k, k'} - g_a^i(\xi_{k, k'}, \omega_{k+1}^{i}, \lambda_{k+1}^{i}) \\
\widehat{\nabla}f_{k, k'}(h_{k, k'}) &:= [\widehat{\nabla}f_{k, k'}^1(h_{k, k'}), \cdots, \widehat{\nabla}f_{k, k'}^N(h_{k, k'}) ]  \\
\widetilde{\nabla}f_{k, k'}^{i}(h_{k, k'}) &:= \frac{N}{N_a}\sum_{n=1}^{N_a}\psi_{\theta_{k}^{i}}(s_n, a_n^i)z_{n, K_z}^{i} - g_a^i(\xi_{k, k'}, \omega_{k+1}^{i}, \lambda_{k+1}^{i}) \\
\widetilde{\nabla}f_{k, k'}(h_{k, k'}) &:= [\widetilde{\nabla}f_{k, k'}^1(h_{k, k'}), \cdots, \widetilde{\nabla}f_{k, k'}^N(h_{k, k'})].
\end{align*}

We now analyze the error at outer-loop iteration $k$. For notational simplicity, we omit the subscript $k$ for the prementioned notations, e.g. we use $\widehat{\nabla}f_{k'}^{i}(h_{k'})$, $\widehat{\nabla}f_{k'}(h_{k'})$, $\widetilde{\nabla}f_{k'}^{i}(h_{k'})$, $\widetilde{\nabla}f_{k'}(h_{k'})$ to represent the above notations, respectively.

\begin{align*}
\| \overline{\nabla}f_{k'}(h_{k'}) - \widetilde{\nabla}f_{k'}(h_{k'})\|^2 \leq 2 \underbrace{\|\overline{\nabla}f_{k'}(h_{k'}) - \widehat{\nabla}f_{k'}(h_{k'}) \|^2}_{I_3} + 2 \underbrace{\|\widehat{\nabla}f_{k'}(h_{k'}) - \widetilde{\nabla}f_{k'}(h_{k'}) \|^2}_{I_4}.
\end{align*}

$I_3$ can be bounded as
\begin{align*}
I_3 &= \|\sum_{n=1}^{N_a}( \frac{1}{N_a}\psi_\theta(s_n, a_n)\psi_\theta(s_n, a_n)^T - F(\theta))h_{k'}\|^2\\
&\leq \|\sum_{n=1}^{N_a}( \frac{1}{N_a}\psi_\theta(s_n, a_n)\psi_\theta(s_n, a_n)^T - F(\theta))\|^2 C_h^2 \\
&\leq \frac{1}{N_a} C_{\psi}^4C_h^2. \numberthis \label{eq:nac-iid-thm-2}
\end{align*}

$I_4$ can be bounded as 
\begin{align*}
I_4 &= \sum_{i=1}^{N}\left\|\psi_{\theta^i}(s_n, a_n^i) \left(\frac{1	}{N_a}\sum_{n=1}^{N_a} N z_{n, K_z}^{i} - \psi_{\theta}(s_n, a_n)^T h_{k'} \right)\right\|^2 \\
&\leq \frac{1}{N_a} NC_{\psi}^2 \sum_{i=1}^{N} \sum_{n=1}^{N_a}\|z_{n, K_z}^{i} - \bar{z}_{n, K_z}\|^2 \\
&= \frac{NC_\psi^2}{N_a}\sum_{n=1}^{N_a}\|QW^{K_z}z_{n,0}\|^2 \\
&\leq \frac{NC_\psi^2}{N_a}\sum_{n=1}^{N_a} \nu^{K_z}\|z_{n, 0}\|^2 \leq N C_{\psi}^{4}C_h^2\nu^{K_z}. \numberthis \label{eq:nac-iid-thm-3}
\end{align*}
Let $K_z = \min \{c\in\NN^+ | \nu^c \leq \frac{4}{N_aN}\}$, then $K_z = \cO(\log\frac{1}{N_a})$. Combine (\ref{eq:nac-iid-thm-2}) and (\ref{eq:nac-iid-thm-3}) gives us
\[ \|\overline{\nabla}f_{k'}(h_{k'}) - \widetilde{\nabla}f_{k'}(h_{k'})\|^2 \leq \frac{4C_\psi^4C_h^2}{N_a}.\]

We now analyze the error of $h_{k, k'}$. Note that we omit the subscript $k$ here for simplifying notation. Define
\[h^*= \argmin_{h} \bar{f}_{\theta}(h) := h^T F(\theta)h := - \EE_{\xi \sim \mu_{\theta}}[g_a(\xi, \omega, \lambda)]^Th. \numberthis \label{eq:nac-direction-obj} \]

It is easy to see that the function on the RHS is strongly convex, since $ F(\theta)$ is positive definite w.r.t. $h$. We bound the optimal gap by
\begingroup
\allowdisplaybreaks
\begin{align*}
\EE\|h_{k'+1} - h^*\|^2 &= \EE\|h_{k'}- \varrho\widetilde{\nabla}f_{k'}(h_{k'}) - h^* \|^2 \\
&= \EE\|h_{k'}-h^*\|^2 - 2\varrho \EE\langle h_{k'}- h^*, \widetilde{\nabla}f_{k'}(h_{k'})\rangle + \varrho^2 \|\widetilde{\nabla}f_{k'}(h_{k'})\|^2\\
&\leq \EE\|h_{k'}-h^*\|^2 - 2\varrho \EE\langle h_{k'}- h^*, \overline{\nabla}f_{k'}(h_{k'})\rangle + 2\varrho \EE\langle h_{k'}- h^*, \overline{\nabla}f_{k'}(h_{k'}) - \widetilde{\nabla}f_{k'}(h_{k'}) \rangle \\
&\quad + 2 \varrho^2\|\overline{\nabla}f_{k'}(h_{k'})\|^2 + 2\varrho^2\|\widetilde{\nabla}f_{k'}(h_{k'}) - \overline{\nabla}f_{k'}(h_{k'})\|^2 \\
&\overset{(i)}{\leq} (1-\varrho\lambda_F)\EE\|h_{k'}-h^*\|^2 - 2\varrho(f_{k'}(h_{k'}) - \overline{f}^*) + 2\varrho \EE\langle h_{k'}- h^*, \overline{\nabla}f_{k'}(h_{k'}) - \widetilde{\nabla}f_{k'}(h_{k'}) \rangle \\
&\quad + 2 \varrho^2\|\overline{\nabla}f_{k'}(h_{k'})\|^2 + 2\varrho^2\|\widetilde{\nabla}f_{k'}(h_{k'}) - \overline{\nabla}f_{k'}(h_{k'})\|^2 \\
&\overset{(ii)}{\leq} (1-\varrho\lambda_F)\EE\|h_{k'}-h^*\|^2 - 2\varrho(1-2\varrho C_{\psi}^2)(f_{k'}(h_{k'}) - \overline{f}^*) \\
&\quad + 2\varrho \EE\langle h_{k'}- h^*, \overline{\nabla}f_{k'}(h_{k'}) - \widetilde{\nabla}f_{k'}(h_{k'}) \rangle + 2\varrho^2\|\widetilde{\nabla}f_{k'}(h_{k'}) - \overline{\nabla}f_{k'}(h_{k'})\|^2 \\
&\overset{(iii)}{\leq} (1-\varrho\lambda_F)\EE\|h_{k'}-h^*\|^2  + 2\varrho \EE\langle h_{k'}- h^*, \overline{\nabla}f_{k'}(h_{k'}) - \widetilde{\nabla}f_{k'}(h_{k'}) \rangle\\
&\quad + 2\varrho^2\|\widetilde{\nabla}f_{k'}(h_{k'}) - \overline{\nabla}f_{k'}(h_{k'})\|^2 \\
% 	&\overset{(iiii)}{\leq} (1-\varrho\lambda_F)\EE\|h_{k'}-h^*\|^2 + 2 \varrho^2\|\overline{\nabla}f_{k'}(h_{k'})\|^2 + 2\varrho^2\|\widetilde{\nabla}f_{k'}(h_{k'}) - \overline{\nabla}f_{k'}(h_{k'})\|^2 \\
&\overset{(iiii)}{\leq} (1-\frac{\varrho\lambda_F}{2}) \EE\|h_{k'}-h^*\|^2 + (\frac{2\varrho}{\lambda_F}+2\varrho^2) \|\widetilde{\nabla}f_{k'}(h_{k'}) - \overline{\nabla}f_{k'}(h_{k'})\|^2, 
\end{align*}
\endgroup
where $\overline{f}^*$ is the optimal value of $\overline{f}(h)$ defined in (\ref{eq:nac-direction-obj}), and the inequality follows the property of $\lambda_F$-strongly convex function: $\overline f(h_2) \geq \overline f(h_1) + \langle \nabla \overline{f}(h_1), h_2 - h_2 \rangle + \frac{\lambda_F}{2}\|h_1-h_2\|^2, \ \forall h_1, h_2$. $(ii)$ uses the PL condition implied by $\lambda_F$-strong convexity: $\overline{f}(h^*) - \overline f(h) \leq -\frac{1}{2\lambda_F}\|\nabla \overline f(h)\|^2, \ \forall h$. $(iii)$ is due to step size rule that $\varrho \leq \frac{1}{2C_\psi^2}$. $(iiii)$ applies Young's inequality.

Use the above induction, we have 
\begin{align*}
\EE\|h_{K_a} - h^*\|^2 &\leq (1-\frac{\varrho\lambda_F}{2})^{K_a}\|h_0 - h^*\|^2 + \sum_{t=0}^{K_a}(1-\frac{\varrho\lambda_F}{2})^t(\frac{2\varrho}{\lambda_F}+2\varrho^2)\|\overline{\nabla}f_{K_a-t}(h_{K_a}) - \widetilde{\nabla}f_{K_a}(h_{K_a})\|^2 \\
&\leq 4 C_h^2(1-\frac{\varrho\lambda_F}{2})^{K_a} + (\frac{4\varrho}{\varrho\lambda_F^2} + \frac{4\varrho}{\lambda_F})C_{\psi}^4C_h^2\frac{4}{N_a}.
\end{align*}
Let $K_a=\min\{c\in \NN^+| 4C_h^2(1-\frac{\varrho\lambda_F}{2})^c = (\frac{4\varrho}{\varrho\lambda_F^2} + \frac{4\varrho}{\lambda_F})C_{\psi}^4C_h^2\frac{1}{N_a} \}$, then $K_a=\cO(\log(\frac{1}{N_a}))$.
Define $C_{18}:= (\frac{16\varrho}{\varrho\lambda_F^2} + \frac{16\varrho}{\lambda_F})C_{\psi}^4C_h^2$, we have 
\[I_2 = \EE\|h_{K_a} - h^*\|^2 \leq \frac{2C_{18}}{N_a}. \numberthis \label{eq:nac-iid-thm-4} \]

Plug (\ref{eq:nac-iid-thm-1}) and (\ref{eq:nac-iid-thm-4}) back to (\ref{eq:nac-iid-thm-0}), we have
\begin{align*}
\EE[J(\theta_{k+1})] - J(\theta_k) &\geq \sum_{i=1}^{N}[\frac{\alpha_k}{4}C_{\psi}^{-2}\|\nabla_{\theta^i}J(\theta_k)\|^2 + \frac{\alpha_k}{2}\lambda_F\|F(\theta_k)^{-1}\EE[g_a^i(\xi_k, \omega_{k+1}^{i}, \lambda_{k+1}^{i})]\|^2 + \alpha_k C_\psi^2 \frac{2C_{18}}{N_a}\\
&\quad +8\lambda_F^{-1}(\varepsilon_{sp}+C_\psi^2\varepsilon_{app}) + 8\lambda_F^{-1} C_\psi^2 \| \omega^*(\theta_k) - \omega_{k+1}^i\|^2 + 4\lambda_F^{-1}C_\psi^2\|\lambda^*(\theta_k) - \lambda_{k+1}^{i}\|^2] 
\end{align*}

Consider the Lyapunov function
\[\VV^k = - J(\theta_k) + \lambda_F^{-1}(\|\omega_k-\omega^*(\theta_k)\|^2 + \|\lambda_k - \lambda^*(\theta_k)\|^2). \]
The difference of the Lyapunov function is 
\begin{align*}
\EE[\VV^{k+1}] - \EE[\VV^{k}] &= \EE[J(\theta_k)] - \EE[J(\theta_{k+1})] + \lambda_{F}^{-1}( \EE\|\omega_{k+1}-\omega^*(\theta_{k+1})\|^2 - \EE\|\omega_k-\omega^*(\theta_k)\|^2 \\
&\quad + \EE\|\lambda_{k+1} - \lambda^*(\theta_{k+1})\|^2 - \EE\|\lambda_k - \lambda^*(\theta_k)\|^2) \\
&\leq \sum_{i=1}^{N}\left[\frac{\alpha_k}{4}C_{\psi}^{-2}\EE\|\nabla_{\theta^i}J(\theta_k)\|^2 + \frac{\alpha_k}{2}\lambda_F\|F(\theta_k)^{-1}\EE[g_a^i(\xi_k, \omega_{k+1}^{i}, \lambda_{k+1}^{i})]\|^2 + \alpha_k C_\psi^2 \frac{2C_{18}}{N_a}\right]\\
&\quad + \lambda_F^{-1}\underbrace{\left[\sum_{i=1}^{N} 8 C_\psi^2\alpha_k \EE\| \omega^*(\theta_k) - \omega_{k+1}^i\|^2 + \EE\|\bar{\omega}_{k+1} - \omega^*(\theta_{k+1}) \|^2 - \EE\| \bar{\omega}_k - \omega^*(\theta_k) \|^2\right]}_{I_5} \\
&\quad + \lambda_F^{-1}\underbrace{\left[\sum_{i=1}^{N} 4 C_\psi^2\alpha_k\EE\|\lambda^*(\theta_k) - \lambda_{k+1}^{i}\|^2 + \EE\|\bar{\lambda}_{k+1} - \lambda^*(\theta_{k+1})\|^2 - \EE\|\bar{\lambda}_k - \lambda^*(\theta_k) \|^2\right]}_{I_6} \\
&\quad + 8N\lambda_F^{-1}(\varepsilon_{sp}+C_\psi^2\varepsilon_{app}). \numberthis \label{eq:nac-iid-thm-5}
\end{align*}
By following the similar procedures through (\ref{eq:markov-thm-9}) to (\ref{eq:markov-thm-15}), we can bound $I_5$ and $I_6$ as
\begin{align*}
I_5 \leq (1+C_{19}\alpha_k)C_{\delta}^2 \beta_k^2 + \frac{\alpha_k}{4}\lambda_F^{-1}\sum_{i=1}^{N}\EE\|F(\theta_k)^{-1}g_a^i(\xi_k, \omega_{k+1}^{i}, \lambda_{k+1}^{i})\|^2 + \alpha_k M_{k_1} + C_{20}\alpha_k^2 \numberthis \label{eq:nac-iid-thm-6} \\
I_6 \leq (1+C_{21}\alpha_k)C_{\lambda}^2 \eta_k^2 + \frac{\alpha_k}{4}\lambda_F^{-1}\sum_{i=1}^{N}\EE\|F(\theta_k)^{-1}g_a^i(\xi_k, \omega_{k+1}^{i}, \lambda_{k+1}^{i})\|^2 + \alpha_k M_{k_2} + C_{22}\alpha_k^2, \numberthis \label{eq:nac-iid-thm-7}
\end{align*}
where $C_{19}, C_{20}, C_{21}, C_{22}$ are some positive constants. Plug (\ref{eq:nac-iid-thm-6}) and (\ref{eq:nac-iid-thm-7}) back to (\ref{eq:nac-iid-thm-5}), we have
\begin{align*}
\EE[\VV^{k+1}] - \EE[\VV^{k}] &\leq \sum_{i=1}^{N}[\frac{\alpha_k}{4}C_{\psi}^{-2}\EE\|\nabla_{\theta^i}J(\theta_k)\|^2 + \alpha_k C_{\psi}^{2}\frac{2C_{18}}{N_a} + \cO(\alpha_k^2 + \beta_k^2 + \eta_k^2) \\
&\quad + (M_{k_1} + M_{k_2})\alpha_k + \cO(\varepsilon_{sp} + \varepsilon_{app})\alpha_k]. \numberthis \label{eq:nac-iid-thm-8}
\end{align*}
By telescoping (\ref{eq:nac-iid-thm-8}), we can get
\begin{align*}
\frac{1}{K}\sum_{k=0}^{K}\sum_{i=1}^{N}\EE\|\nabla_{\theta^i}J(\theta_k)\|^2 &\leq \frac{4C_{\psi}^{2}\VV_0}{K\alpha_k} + \cO(\varepsilon_{sp} + \varepsilon_{app}) + \frac{8C_\psi^2C_{18}}{N_a} + \cO(\alpha_k + \frac{\beta_k^2}{\alpha_k} + \frac{\eta_k^2}{\alpha_k}) \\
&\quad + 4C_{\psi}^2\frac{1}{K}\sum_{k=0}^{K}(M_{k_1} + M_{k_2})
\end{align*}

By (\ref{eq:markov-thm-18}), $M_{k_1} + M_{k_2} = \cO(\frac{1}{\sqrt{K}})$ when $K_c \leq \cO(K^{1/4})$. Therefore, let $C, \bar{\alpha}$ be some positive constants. Set $N_a=C\sqrt{K}$, $\alpha_k=\frac{\bar{\alpha}}{\sqrt{K}}$, $\beta_k=\frac{C_9}{2\lambda_{\phi}}\alpha_k$, $\eta_k=\frac{C_{10}}{2\lambda_{\varphi}}\alpha_k$, we obtain the desired result of (\ref{eq:nac-convergence-gradient}).

We now prove (\ref{eq:nac-convergence-optgap}). Let $\EE_{\theta^*}$ denote the expectation over $s \sim d_{\pi_{\theta^*}}, a \sim \pi_{\theta^*}(\cdot|s)$. By the smoothness of $\log\pi_{\theta}(a|s)$, we have
\begin{align*}
&\EE_{\theta^*}[\log \pi_{\theta_{k+1}}(a|s) - \log\pi_{\theta_k}(a|s) ]\\ &\geq \alpha_k \EE_{\theta^*}[\psi_{\theta_k}(s,a)^Th_{k}] - \frac{L_\psi\alpha_k^2}{2}C_h^2 \\
&\geq \alpha_k\EE_{\theta^*}[\psi_{\theta_k}(s,a)^T(h_k-h^*(\theta_k))] + \alpha_k\EE_{\theta^*}[\psi_{\theta_k}(s,a)^Th^*(\theta_k) - A_{\theta_k}(s,a)] \\
&\quad + \alpha_k \EE_{\theta^*}[A_{\theta_{k}}(s,a)] - \frac{L_\psi\alpha_k^2}{2}C_h^2 \\
&\geq -\alpha_k C_\psi\|h_k - h^*(\theta_k)\| - \alpha_k\sqrt{\varepsilon_{actor}} + \alpha_k(J(\theta^*) - J(\theta_k)) - \frac{L_\psi\alpha_k^2}{2}C_h^2.
\end{align*}

By telescoping the above inequality and rearranging terms, we have
\begin{align*}
\frac{1}{K}\sum_{k=1}^{K} (J(\theta^*) - J(\theta_k)) &\leq \frac{1}{K\alpha_k}\EE_{\theta^*}[\log \pi_{K}(a|s) - \log \pi_{0}(a|s)] + \sqrt{\varepsilon_{actor}} \\
&\quad + \frac{1}{K} \sum_{k=1}^{K}  C_\psi\|h_k-h^*(\theta_k)\| + \frac{1}{K}\sum_{k=1}^{K}\frac{L_\psi\alpha_k}{2}.
\end{align*}
The term $\|h_k - h^*(\theta_k)\| \leq \|h_k - F(\theta_k)^{-1}\EE[g_a(\xi_k, \omega_{k+1}, \lambda_{k+1}]\| + \|\EE[g_a(\xi_k, \omega_{k+1}, \lambda_{k+1}] - F^{-1}\nabla J(\theta_k)\|$. Since by the (\ref{eq:nac-iid-thm-4}) and (\ref{eq:markov-thm-gradbias-bound}), these two terms are of order $\cO(\frac{1}{N_{a}^{1/2}})$ and $\cO(\|\omega_k - \omega_{k+1}\| + \varepsilon_{app})$, respectively, we conclude that $\|h_k - h^*(\theta_k)\|$ is of order $\cO(\|\omega_k - \omega^*(\theta_k)\| + \varepsilon_{app})$. By following the step size rule as suggested by Theorem~\ref{thm:nac-convergence},
we obtain the desired result.

\end{document}